%% file: main.tex
\documentclass[runningheads]{llncs}

\usepackage{eccv}

\usepackage{eccvabbrv}

\usepackage{listings}
\usepackage{xcolor}
\usepackage{graphicx}
\usepackage{booktabs}

\usepackage[accsupp]{axessibility}  % Improves PDF readability for those with disabilities.

\usepackage{hyperref}

\usepackage{orcidlink}

\usepackage{booktabs}
\usepackage{subcaption}
\usepackage{graphicx}
\usepackage{multirow}
\usepackage{amsmath}

\renewcommand\thanks[1]{\footnotetext{#1}}

\begin{document}

% \title{NaLA: A Geometry-Aware Layout Agent for End-to-End 3D Scene Generation} 
\title{NaLA: A 3D Native LLM Layout Agent for High-quality 3D Scene Generation
\thanks{\dag \ Corresponding authors.}
\thanks{Project page: \url{https://adamcwan.github.io/NaLA/}.}
}

\titlerunning{NaLA: A 3D Native LLM Layout Agent}

%e.g. \orcidlink{0000-1111-2222-3333}
\author{Cheng Wan\inst{1,2} 
\and Yongsen Mao\inst{1}
\and Wenzheng Wu \inst{3} \and
Yuxuan Xie \inst{4} \and\\ 
Chucheng Xiang \inst{3}
\and Runze Wang \inst{3}
\and Xiang Zhang \inst{4}\and\\ 
Zhongyuan Liu \inst{4}\and
Rushi Dai \inst{5\dag}
\and Yuan Liu \inst{1\dag}
}

\authorrunning{C.~Wan et al.}

\institute{Hong Kong University of Science and Technology
\and Shenzhen Loop Area Institute
\and University of Science and Technology of China
\and Tencent IEG
\and Hong Kong University of Science and Technology (Guangzhou)
}

% \url{http://www.springer.com/gp/computer-science/lncs} 

\maketitle

\begin{abstract}
Recently, Large Language Models (LLMs) have emerged as promising layout agents for 3D scene generation. Existing layout agents still suffer from implausible layout generation because most of them convert 3D assets and 3D layouts into textual descriptions as inputs and outputs, which involves severe information loss due to the modality gap between texts and 3D assets and 3D layouts. We propose NaLA, a native 3D LLM layout Agent for high-quality 3D scene generation by placing 3D assets in the scene. For the inputs, NaLA encodes 3D scene boundaries and 3D assets directly into the LLM, preserving fine-grained geometry and enabling explicit reasoning over relationships like collisions, surface supporting, and containment. To accurately output the positions and orientations of assets, NaLA adopts a coarse-to-fine prediction mechanism that first predicts discrete poses in an autoregressive manner and then refines the discrete poses with a continuous regression. Trained on diverse layout datasets, NaLA attains strong geometric perception and layout coherence. Experiments demonstrate that NaLA outperforms prior layout agents in both generation quality and inference efficiency, with comprehensive ablation studies to verify each component's effectiveness.
  \keywords{3D Scene Generation \and Large Language Models \and 3D Layout Agent}
\end{abstract}

\begin{figure}[htbp]
    \centering
    \includegraphics[width=\linewidth]{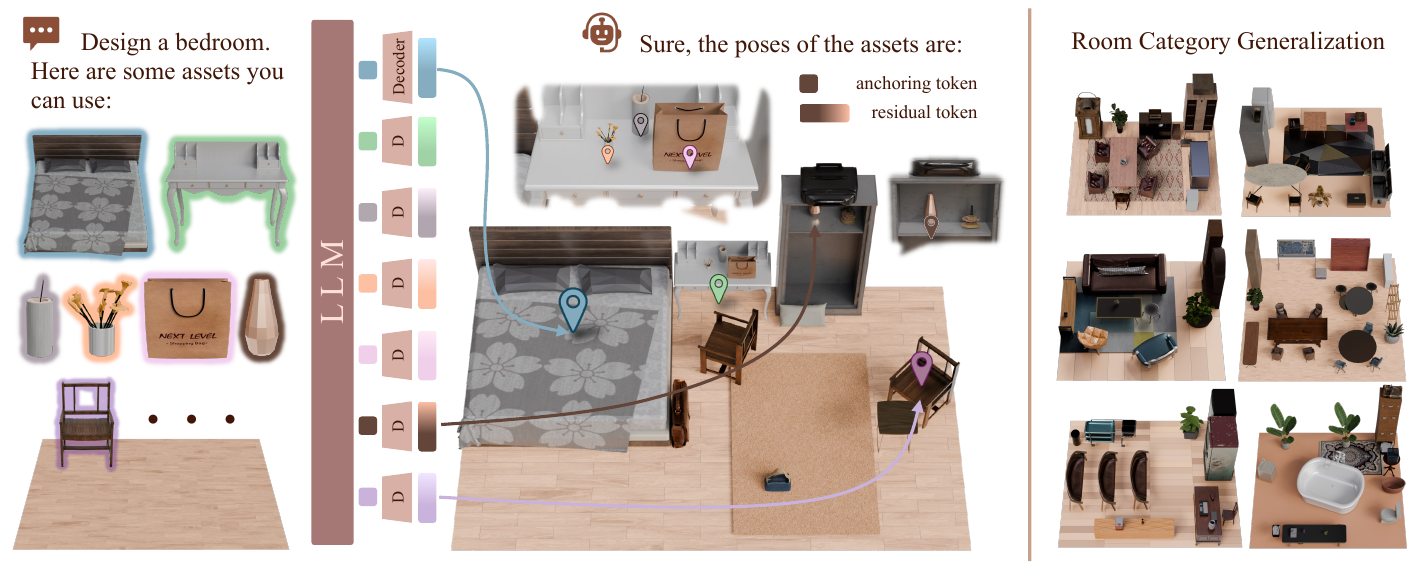}
    \caption{\textbf{Geometry-aware 3D layout generation with NaLA.} Given a design prompt and a set of 3D assets, NaLA arranges both large furniture and small objects into plausible layouts by reasoning over point-cloud geometry and predicting poses via a coarse-to-fine mechanism, generalizing across room categories (living room, dining room, lounge, bathroom).}
    \label{fig:intro demo}
\end{figure}

\input{main_text/introduction}

\input{main_text/related_work}

\input{main_text/problem_definition}

\input{main_text/method}

\input{main_text/experiment}

\input{main_text/conclusion}

\input{main_text/acknowledgement}

% \section*{Acknowledgements}
% Please insert your acknowledgments here.

\bibliographystyle{splncs04}
\bibliography{main}

\newpage
\appendix

\input{appendix/appendix}

\end{document}

%% file: main_text/introduction.tex
\section{Introduction}
\label{sec: intro}

3D Scene generation is widely used across various domains, including interior layout design \cite{makeithome2011}, game development \cite{scenecraft2023}, and embodied AI \cite{internscenes2025}. 
Traditionally, arranging 3D objects to construct a scene relied heavily on manually crafted rules \cite{rulebased2017} or complex optimization solvers \cite{MIQP2018}, which often fail to capture diverse human requirements and are difficult to scale. Recently, LLMs \cite{LLMfewshot2020} and Vision-Language Models (VLMs) \cite{radford2021learning} have transformed this landscape. Acting as central layout agents, these foundation models demonstrate remarkably high-level reasoning capabilities, seamlessly interpreting complex user instructions to generate coherent object arrangements \cite{SpatialLM}.

However, existing LLM-based layout agents fundamentally treat 3D assets as a linguistic abstraction rather than a native spatial signal, relying heavily on intermediate text or image translations. On the input side, these agents typically translate 3D scene boundaries and 3D assets into textual descriptions or rendered 2D images before feeding them into a general-purpose LLM \cite{LayoutGPT}. On the output side, they follow two main paradigms: either directly predicting object poses as sequences of textual coordinate digits (e.g., (0,1,0)) \cite{llplace}, or inferring pairwise spatial constraints (e.g., ``facing'', ``near'') to form a scene graph \cite{DIscene2024}, which is subsequently resolved into optimal poses by a downstream mathematical optimizer \cite{holodeck,sun2025layoutvlmdifferentiableoptimization3d}.
 
This indirect ``translation'' process creates a severe domain gap at both the input and output stages, fundamentally restricting the agent's capability to precisely perceive and manipulate 3D space.  On the input side, translating 3D assets and spatial layouts into text or flat images inevitably causes irreversible information loss. Fine-grained details, such as concave structures, precise supporting surfaces, or irregular room boundaries, cannot be faithfully preserved, leaving the LLM's understanding of spatial affordance incomplete. 
On the output side, the standard practice of representing continuous coordinates as sequences of discrete text tokens (e.g., decomposing ``1.23'' into ``1'', ``.'', ``2'', ``3'') destroys the inherent numerical integrity. This ``numerical blindness'' makes it difficult for LLMs to reason about distance and continuity \cite{zausinger2025regress}, frequently resulting in physically implausible placements, such as collisions, floating, or out-of-boundary objects. % or .%, and necessitating computationally expensive post-optimization. 

To bridge this gap between language and actual 3D space, we propose \textbf{NaLA}, a 3D \textbf{Na}tive \textbf{L}ayout \textbf{A}gent,
%an end-to-end autonomous layout agent 
that directly consumes 3D point clouds and natively regresses precise spatial placements. 
%Without relying on any predefined placement rules or external constraint solvers, 
Our approach eliminates the intermediate textual translation of geometric assets,
%and. Instead, it 
preserves absolute geometric fidelity, and allows the foundation model to directly perceive, reason over, and interact with true 3D spatial structures. This design significantly improves placement precision, robustness in handling irregular environments, and overall generation efficiency. \cref{fig:intro demo} presents diverse scenes generated by NaLA.
 
On the input side, NaLA avoids information loss by directly encoding raw scene and asset point clouds into compact input 3D tokens for the LLM. 
Specifically, we utilize point cloud encoders to extract features from the empty scene (including walls and floors) and the target objects, projecting them into the LLM's embedding space as prefix tokens. By perceiving actual input 3D tokens rather than textual approximations, NaLA inherently understands fine-grained object structure and irregular room shapes without requiring any hard-coded logic or boundary constraints.%, thereby improving its placement accuracy.
%, spatial capacities, and functional geometries. %This enables NaLA to capture complex, irregular room boundaries and arbitrary asset structures without requiring any hard-coded logic or boundary constraints.
 
On the output side, we implement a coarse-to-fine prediction mechanism that provides stable spatial anchoring while preserving precision. 
To avoid the destruction of continuous pose information caused by tokenization, a natural idea is to follow LISA \cite{lai2024lisa} by introducing a special token and decoding its corresponding feature to obtain continuous poses. However, we find that this strategy causes the model to confuse the poses of previously placed assets.
To address this issue, we propose a coarse-to-fine prediction mechanism. NaLA first partitions the space into multiple grids and outputs a sequence of pose anchoring tokens to determine the rough spatial location of an asset. It then emits a pose residual token, whose feature is decoded to obtain pose residuals that refine the final pose.
This coarse-to-fine prediction mechanism preserves the integrity of continuous pose information while ensuring stable and efficient generation, enabling NaLA to efficiently produce precise end-to-end asset pose sequences.
%First, to overcome the inherent fragmentation caused by tokenizing continuous coordinates into discrete digit strings, we introduce a specialized regression token placeholder, denoted as $\langle\texttt{POS\_TOKEN}\rangle$, whose hidden states are decoded into a continuous pose vector via a multi-layer perceptron (MLP). This design allows the model to predict object transformations as holistic geometric entities rather than fractured linguistic sequences, significantly compressing the output space while maintaining numerical fidelity. 
%Furthermore, to prevent the model from losing track of spatial context in long generation sequences, we introduce a set of discrete anchoring tokens. During generation, NaLA first emits four discrete tokens to anchor the asset within a coarse global grid, effectively ``grounding'' the layout history, before immediately generating the $\langle\texttt{POS\_TOKEN}\rangle$ for precise fine-grained adjustments. 
%This coarse to fine output mechanism enables NaLA to perform asset placement efficiently and with high precision.
%This hybrid mechanism ensures that the model retains both the macroscopic global context and the microscopic sub-grid precision required for realistic object interaction.

Extensive experiments demonstrate that NaLA achieves physical plausibility comparable to state-of-the-art baselines while significantly outperforming them in semantic coherence and aesthetic quality. Beyond standard metrics, NaLA exhibits superior robustness in handling irregular scene layouts and achieves drastically higher generation efficiency compared to existing methods. Furthermore, comprehensive ablation studies validate the necessity and effectiveness of each proposed component within our framework.

Our primary contributions are:
(1) We propose NaLA, an innovative layout agent with integrated 3D geometry perception, enabling the LLM to directly understand point clouds for accurate scene generation without intermediate textual translation.
(2) We introduce a specialized coarse-to-fine output mechanism that utilizes pose anchoring tokens with pose residual tokens, profoundly improving both the precision and efficiency of multi-asset pose prediction.
(3) By leveraging a comprehensive data augmentation strategy and a staged training curriculum on large-scale layout datasets, NaLA acquires robust geometric perception and advanced layout planning capabilities. Extensive experiments demonstrate that NaLA significantly outperforms previous baselines across various quantitative metrics and also generation speeds.

%% file: main_text/related_work.tex
\section{Related Work}

\subsection{3D Scene Generation}
As a central goal in computer vision, 3D scene generation aims to produce physically plausible and visually realistic environments \cite{wen20253d,zhang2026m3dlayoutmultisourcedataset3d}. 
% Recent advancements typically utilize 3D representations like Neural Radiance Fields (NeRF) \cite{mildenhall2021nerf} and 3D Gaussian Splatting \cite{kerbl20233dgaussian} trained on large-scale datasets \cite{NeuralField-LDMKim_2023_CVPR, li2024director3drealworldcameratrajectory}, or leverage strong generative priors to reconstruct 3D scenes from synthesized images or videos \cite{WonderJourney2024Yu, 4real2024Yu}. 
Recent advancements involve training a generative model to provide layout arrangement \cite{tang2024diffuscene}, utilizing 3D representations \cite{NeuralField-LDMKim_2023_CVPR, li2024director3drealworldcameratrajectory} like Neural Radiance Fields (NeRF) \cite{mildenhall2021nerf} and 3D Gaussian Splatting \cite{kerbl20233dgaussian}, or leveraging strong generative priors to reconstruct 3D scenes from synthesized images or videos \cite{WonderJourney2024Yu, 4real2024Yu}. 
However, despite their strong visual priors, these approaches often struggle to accurately follow complex user instructions and lack discrete spatial reasoning capabilities. To address this, recent pipelines have integrated LLMs or VLMs to handle high-level layout planning and asset composition.

\subsection{LLM for 3D Scene Generation}
\label{subsec: LLM 4 3D scene}
Existing LLM-based layout agents primarily focus on indoor environments and fall into two output paradigms. The first directly generates asset poses as textual coordinates (e.g., LayoutGPT \cite{LayoutGPT}, SceneTeller \cite{SceneTeller2024}, LLplace \cite{llplace}, and SceneX \cite{zhou2025scenex}). The second paradigm infers pairwise spatial constraints \cite{SceneGraphLI2024127052}, as seen in GraphDreamer \cite{gao2024graphdreamer}, Holodeck \cite{holodeck}, LayoutVLM \cite{sun2025layoutvlmdifferentiableoptimization3d}, and I-Design \cite{I-Design_elen_2025}, which rely on external optimization modules to subsequently resolve these constraints into specific poses. However, both paradigms represent 3D geometry using text or 2D images, lacking direct spatial perception \cite{LLaVA-3DZhu_2025_ICCV}. Furthermore, restricted by textual output, treating continuous 3D parameters as discrete tokens restricts both the precision and efficiency of pose generation \cite{zausinger2025regress}. Motivated by these limitations, we propose NaLA, a native 3D LLM layout Agent for 3D scene generation. %NaLA introduces a specifically designed 3D-aware LLM capable of direct spatial reasoning.

%% file: main_text/problem_definition.tex
\section{Problem Definition}
\label{sec: problem definition}
We formulate 3D layout generation as an autonomous spatial planning task. Given a text-based user requirement, information about the empty scene, and an asset library containing $n$ objects, the core objective of the layout LLM is to jointly process these input features and predict the 3D position $\boldsymbol P_i$, scaling factor $\boldsymbol S_i$, and orientation $O_i$ for every asset.
Without loss of generality, we assume a $y$-up coordinate system where objects rotate exclusively around the $y$-axis. Consequently, $\boldsymbol P_i$ and $\boldsymbol S_i$ are explicitly treated as 3D continuous vectors, and $O_i$ is simplified to a single rotation angle.
%While most existing methods encode scene and asset features via text descriptions or top-down images, they fail to capture fine-grained 3D contours. GeoAgent directly consumes point cloud geometric features. 

%% file: main_text/method.tex
\section{Method}

\begin{figure}[t] 
\centering \includegraphics[width=1\linewidth]{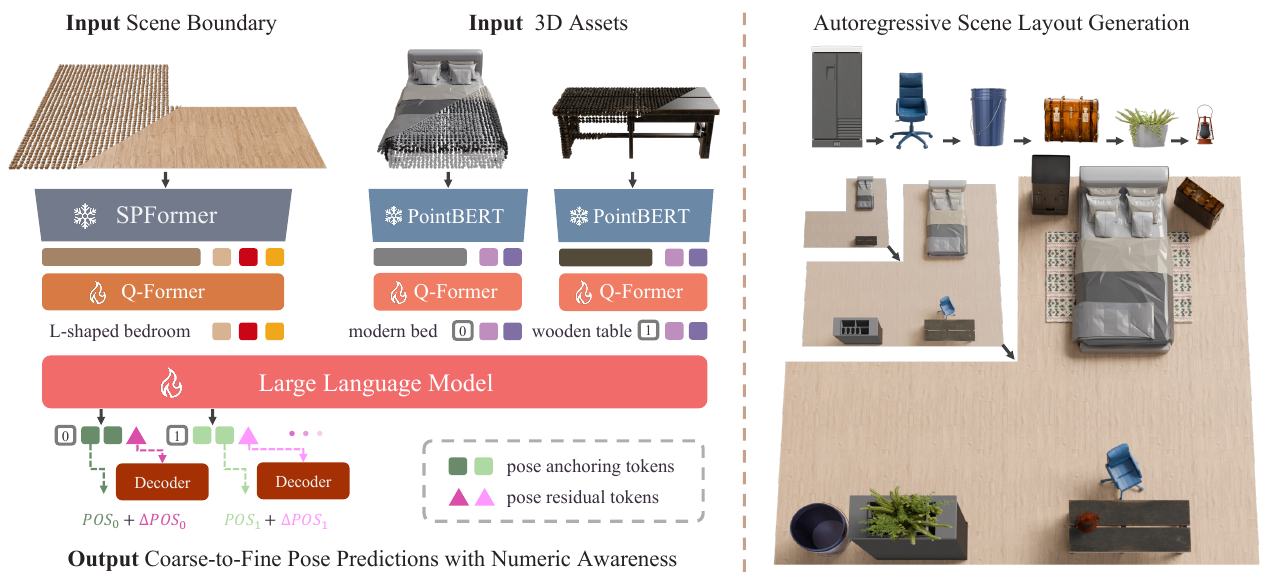} 
\caption{\textbf{Overview of the NaLA pipeline. }\textbf{3D token encoding:} room layout and asset point clouds are encoded by SPFormer and PointBERT, respectively, then mapped via Q-Formers into a unified token sequence for the LLM; pose anchoring tokens and pose residual tokens are decoded into coarse-to-fine poses (POS$_n$ + $\Delta$POS$_n$). \textbf{Autoregressive layout generation:} objects are selected from the asset pool and placed sequentially to form the complete scene layout.}
\label{fig: model pipeline} 
\end{figure} 

As an end-to-end layout generation framework, NaLA directly encodes point clouds of 3D assets and 3D scene boundaries and decodes precise object poses.% through a specialized coarse-to-fine prediction mechanism. 
As illustrated in \cref{fig: model pipeline}, NaLA's pipeline consists of three components: a point cloud input module for detailed geometry perception, a coarse-to-fine prediction mechanism for robust pose generation, and a tailored training strategy for layout rule acquisition.

\subsection{Input 3D Token Encoding}

\begin{figure*}[h]
  \centering
  \begin{subfigure}[b]{0.22\linewidth}
    \centering
    \includegraphics[width=\linewidth,trim=12cm 12cm 12cm 10cm,clip]{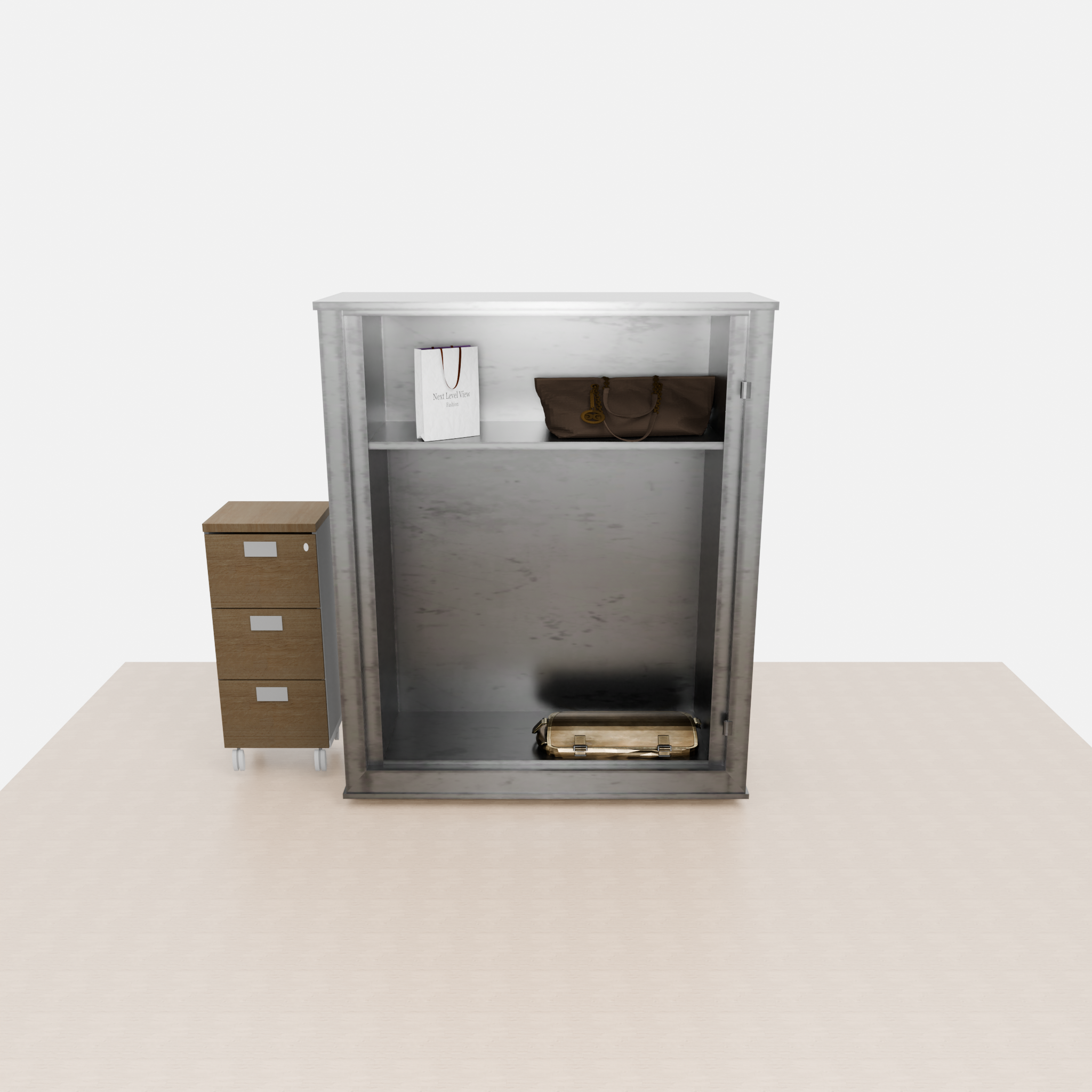} 
    \caption{NaLA}
    \label{fig:method_nala}
  \end{subfigure}
  \hfill 
  \begin{subfigure}[b]{0.22\linewidth}
    \centering
    \includegraphics[width=\linewidth,trim=12cm 12cm 12cm 10cm,clip]{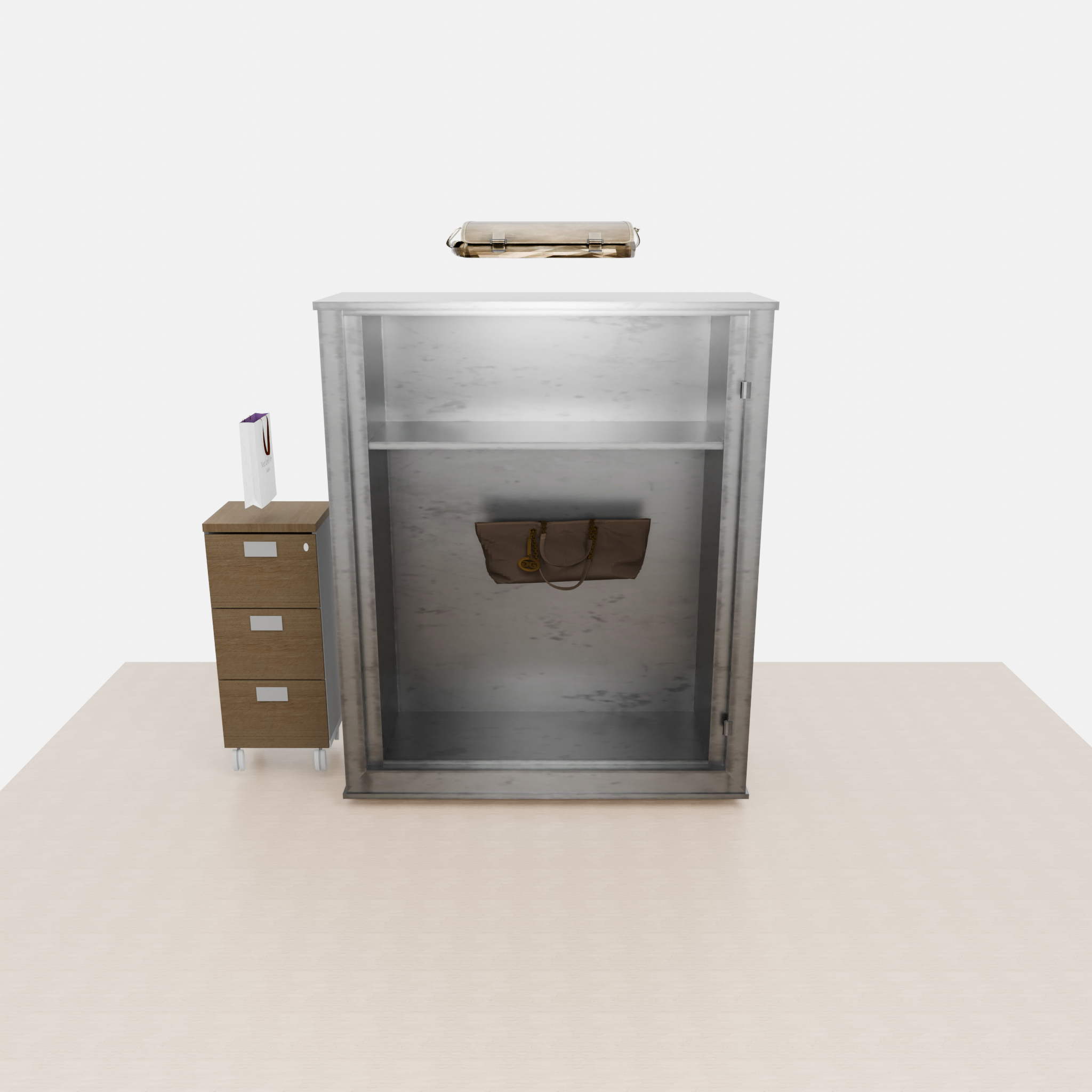}
    \caption{LayoutGPT}
    \label{fig:method_layoutgpt}
  \end{subfigure}
  \hfill
  \begin{subfigure}[b]{0.22\linewidth}
    \centering
    \includegraphics[width=\linewidth,trim=18cm 18cm 18cm 18cm,clip]{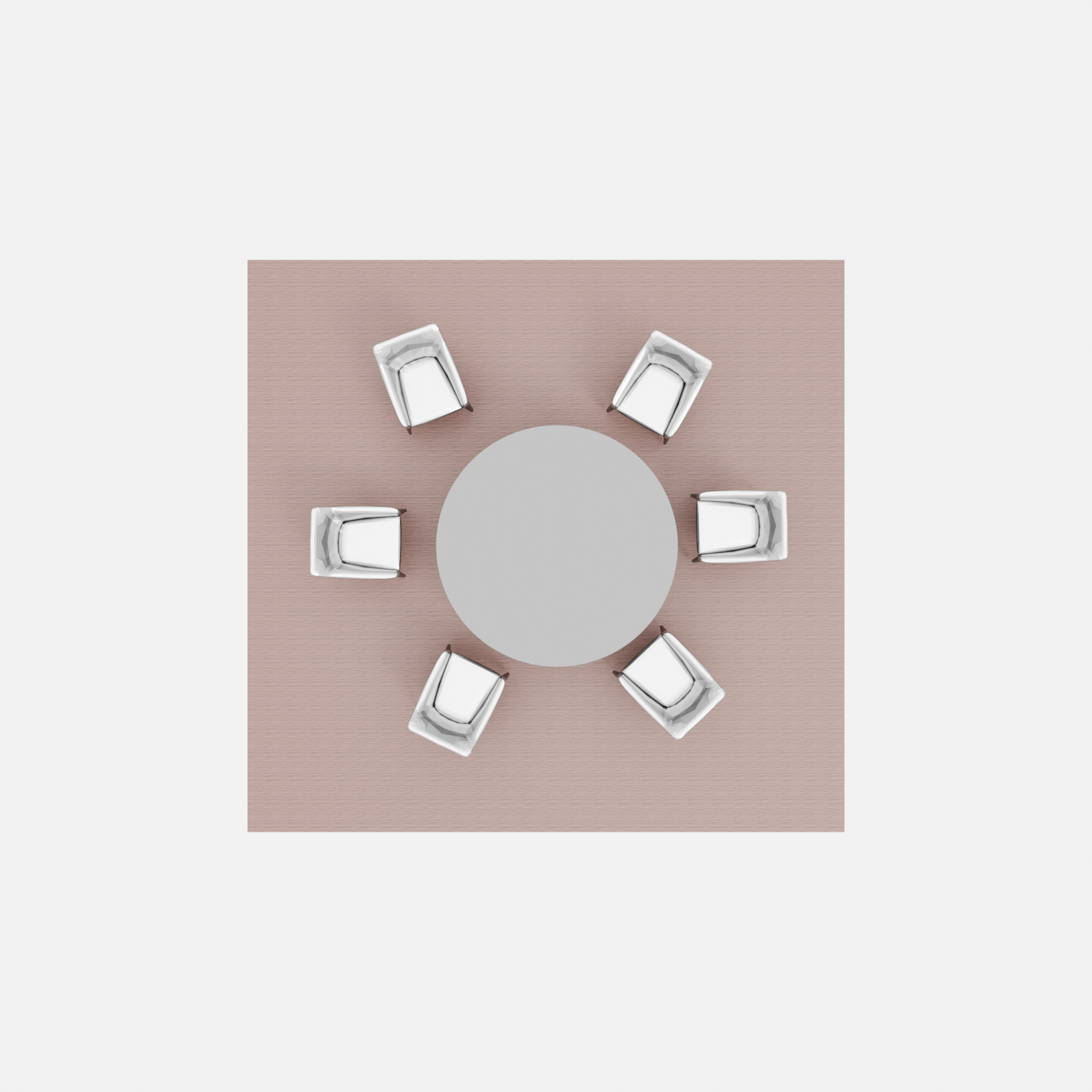}
    \caption{NaLA}
    \label{fig:method_nala2}
  \end{subfigure}
    \hfill
  \begin{subfigure}[b]{0.22\linewidth}
    \centering
    \includegraphics[width=\linewidth,trim=18cm 18cm 18cm 18cm,clip]{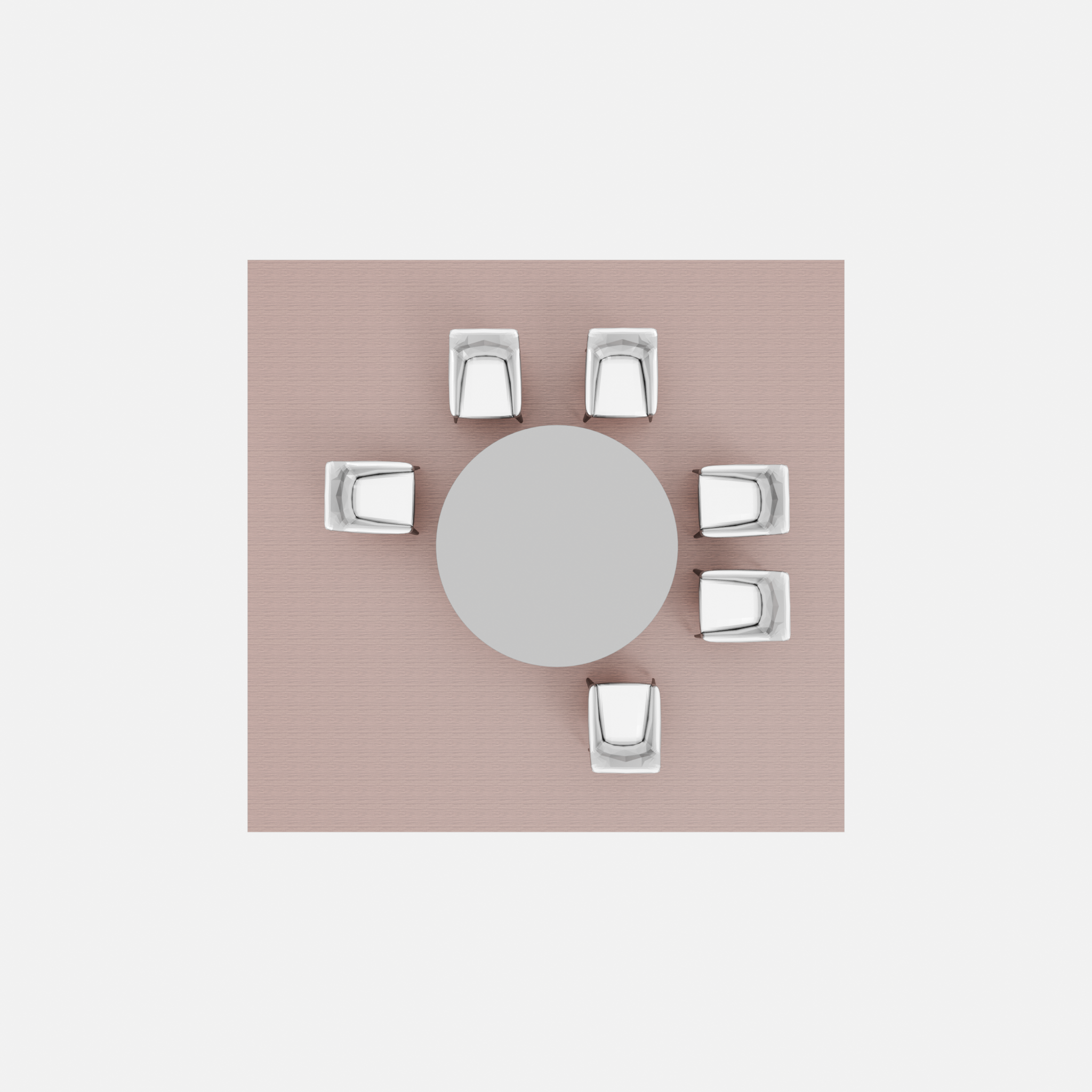}
    \caption{Holodeck}
    \label{fig:method_holodeck}
  \end{subfigure}
  \caption{Placement comparison between NaLA and baseline models. Baseline models cannot perceive fine-grained asset geometry (e.g., round tables, cabinets with shelves), failing to achieve the precise placements produced by NaLA.}
  \label{fig:four_comparisons}
\end{figure*}

Standard layout agents \cite{LayoutGPT,holodeck} typically represent assets using bounding boxes or text descriptions. This simplification often leads to physical inconsistencies, such as failing to nest objects (e.g., placing items inside a shelf) or aligning asymmetrical assets, as shown in \cref{fig:four_comparisons}. 
Instead, we propose an encoder that directly encodes all 3D information and feeds the encoded features into our LLM Agent to avoid information loss.
% To address this, NaLA explicitly incorporates geometric features.

Specifically, we encode both the 3D scene boundary information and 3D asset information as input 3D tokens.
First, to enable the model to perceive fine-grained asset details, we sample each asset into a point cloud and encode it using PointBERT \cite{yu2021pointbert}.
To preserve spatial geometry, the encoder outputs dense patch features explicitly concatenated with their absolute 3D coordinates.
On the other hand, since room shapes may be irregular, we also sample the empty room—including the floor, walls, and ceiling—into a point cloud and encode it using SPFormer \cite{spfprmer2022}. 
To bridge the modality gap, we employ two lightweight, trainable Q-Formers (Query Transformers \cite{li2022blip}) to encode the dense visual features of the assets and the scene into a compact set of query tokens, respectively. 
These tokens are projected via a trainable MLP to the LLM's embedding space and injected as prefixes, allowing the model to attend to fine-grained geometry.
%XXX
% We employ two specialized encoders to capture structural details. For the scene mesh (floors, walls, ceilings), we use a frozen Spformer \cite{spfprmer2022} to extract superpoint features. For assets, we use a frozen PointBERT \cite{yu2021pointbert} to extract patch features. To bridge the modality gap, we employ two lightweight, trainable Q-Formers (Query Transformers \cite{li2022blip}) for both branches. These modules aggregate the dense visual features into a compact set of query tokens, respectively. These tokens are projected via a trainable MLP to the LLM's embedding space and injected as prefixes, allowing the model to attend to fine-grained geometry.

\subsection{Coarse-to-Fine Output Layout Token Design}
\label{subsec: output}

\begin{figure*}[h]
  \centering
  \begin{subfigure}[b]{0.32\linewidth}
    \centering

    \includegraphics[width=0.8\linewidth]{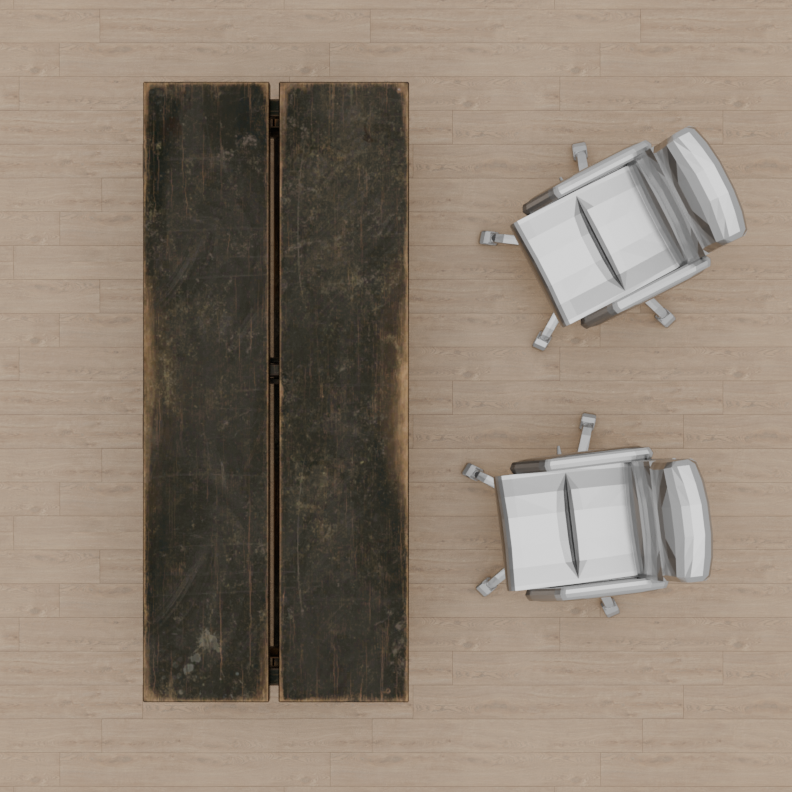} 
    \caption{Anchoring+Residual}
    \label{fig:method_standard}
  \end{subfigure}
  \hfill 
  \begin{subfigure}[b]{0.32\linewidth}
    \centering
    \includegraphics[width=0.8\linewidth]{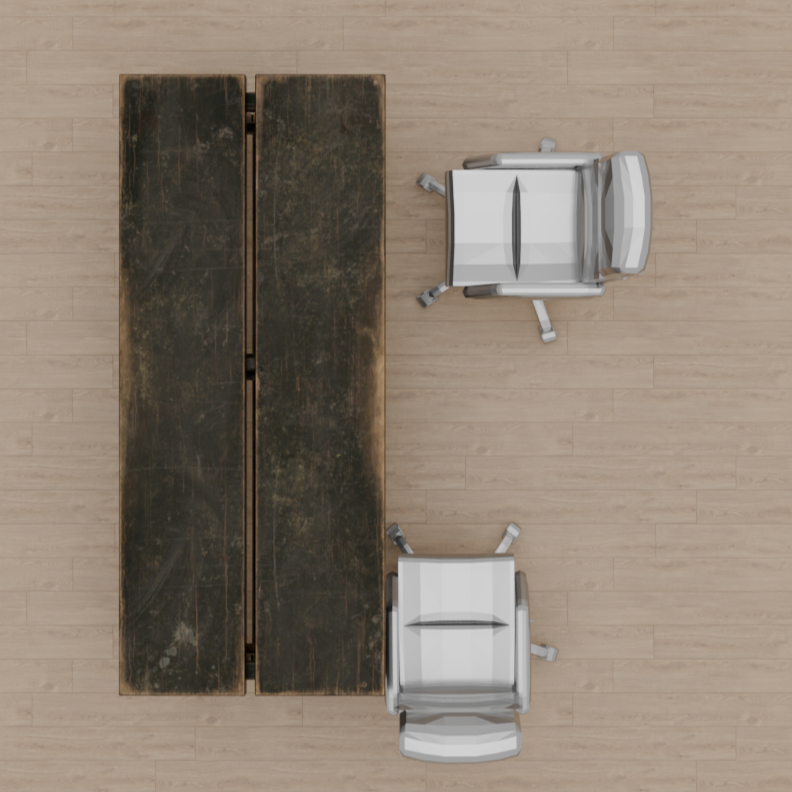}
    \caption{Anchoring}
    \label{fig:method_discrete}
  \end{subfigure}
  \hfill
  \begin{subfigure}[b]{0.32\linewidth}
    \centering
    \includegraphics[width=0.8\linewidth]{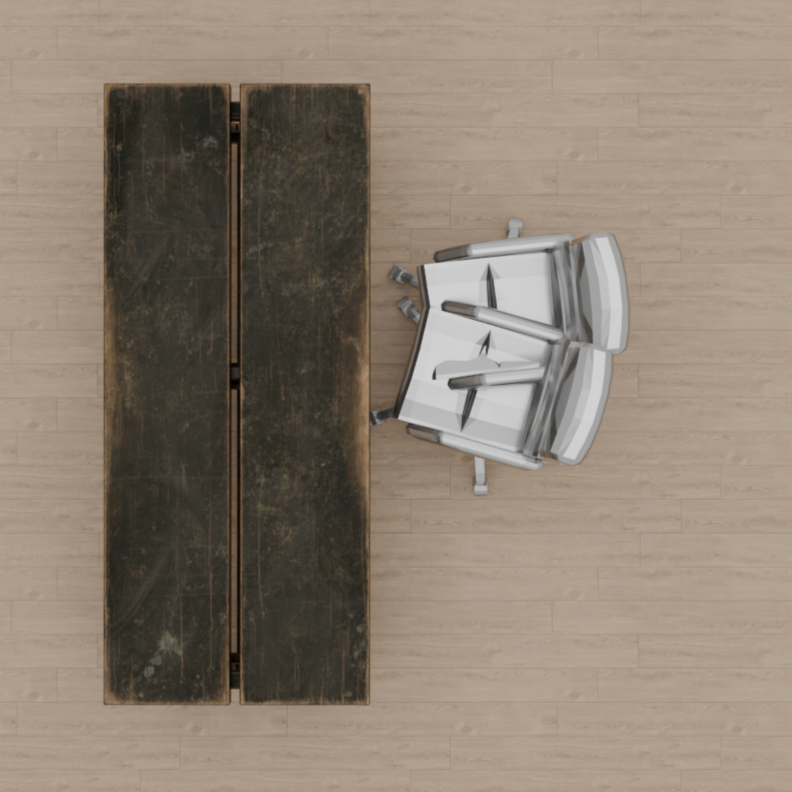}
    \caption{Residual}
    \label{fig:method_regression}
  \end{subfigure}
  \caption{\textbf{Comparison of different output designs in NaLA.}
Using only output anchoring tokens fails to achieve precise placement, while using only output residual tokens results in severe object overlap.}
  \label{fig:three_comparisons}
\end{figure*}

As noted in \cref{sec: intro}, conventional LLMs decompose object poses into discrete tokens, which breaks the integrity of pose information and leads to ``numerical blindness''. 
To avoid tokenizing pose values, a natural idea is to follow approaches such as LISA \cite{lai2024lisa}, introducing a single special token and decoding continuous poses directly from its feature. 
However, when multiple objects need to be placed, regressing all their poses from repeatedly reused special tokens causes the LLM to confuse them, leading to severe object collision (shown in \cref{fig:three_comparisons}). Therefore, we design a \textbf{Coarse-to-Fine Pose Prediction Mechanism} for NaLA. The model first localizes the rough spatial region of an asset using Pose Anchoring Tokens, followed by a Pose Residual Token that decodes the continuous pose refinement. We next introduce the detailed mechanism.

\textbf{Pose Anchoring Tokens:} 
To determine the asset's approximate spatial region,
%To better locate existing assets in the generation sequence, 
%we design a set of discrete Anchoring Tokens. W
we discretize the three scene dimensions into $B$ bins and register $B$ special tokens, $\langle\texttt{POS\_BIN\_1}\rangle, \ldots, \langle\texttt{POS\_BIN\_B}\rangle,$ to represent coarse spatial coordinates. Additionally, we register four special tokens, $\langle\texttt{ROT\_BIN\_1}\rangle, \ldots, \langle\texttt{ROT\_BIN\_4}\rangle,$ to represent coarse object orientations corresponding to the positive and negative directions of the $x$ and $z$ axes (assuming a $y$-up system). 
Before predicting the precise pose of an asset, 
NaLA first predicts an anchoring sequence to determine the coarse global position and orientation:
\begin{equation}
    \langle\texttt{POS\_BIN\_j}\rangle 
    \langle\texttt{POS\_BIN\_k}\rangle 
    \langle\texttt{POS\_BIN\_p}\rangle 
    \langle\texttt{ROT\_BIN\_q}\rangle,
    \label{equ: discrete tokens}
\end{equation}
where $1 \le j,k,p \le B$ and $1 \le q \le 4$.
These four tokens uniquely define a coarse orientation $\widehat{O}_i^{*}$, and coarse asset pose $\widehat{\boldsymbol{P}}_i^{*}$, which corresponds to the center of the selected voxel grid cell.

\textbf{Pose Residual Tokens:}
Although the pose anchoring tokens can localize the rough position of an asset, these four tokens alone are insufficient for fine-grained placement (e.g., placing an object precisely on the floor or against a wall).
To enable the LLM to output holistic continuous pose information, we introduce a specialized regression placeholder, denoted as $\langle\texttt{POS\_TOKEN}\rangle$. Unlike standard tokens that map to a fixed vocabulary, this token serves as a dynamic interface for continuous value prediction. Specifically, during generation, we extract the hidden states of $\langle\texttt{POS\_TOKEN}\rangle$ from the last $L$ layers of the LLM to capture rich semantic and spatial context. These aggregated features are then projected through a lightweight decoder head—consisting of a two-layer Transformer followed by an MLP—which directly regresses a 
%continuous pose vector (representing position $\boldsymbol P_i$, scaling factor $\boldsymbol S_i$, and orientation $O_i$) in a single forward pass, bypassing the precision loss associated with discrete text decoding.
fine-grained residual vector $(\Delta \widehat{\boldsymbol{P}}_i, \widehat{\boldsymbol S}_i, \Delta \widehat{O}_i)$. 
%Here, $\Delta \widehat{\boldsymbol{P}}_i$ and $\Delta \widehat{O}_i$ represent the residual offset and rotation relative to the coarse anchor, while $\widehat{\boldsymbol S}_i$ denotes the absolute scaling factor. 
By combining the pose anchoring tokens and pose residual token,
the final precise pose is computed as:
$$
(\widehat{\boldsymbol P}_i^{*} + \Delta \widehat{\boldsymbol P}_i,\; \widehat{\boldsymbol S}_i,\; \widehat{O}_i^{*} + \Delta \widehat{O}_i).
$$

%This procedure is illustrated in \cref{fig: model output}. 
Furthermore, \cref{fig:three_comparisons} illustrates the placements produced by NaLA under different output strategies. Without pose anchoring tokens, pose information must be heavily compressed into the hidden states of the single $\langle\texttt{POS\_TOKEN}\rangle$. Since this token is reused across all generation steps, the LLM struggles to differentiate between previously placed assets and the current target, creating ambiguity in the autoregressive chain. Consequently, combining pose anchoring tokens and pose residual tokens is essential for robust and accurate pose generation.

%Details of the discretization and reconstruction are provided in the appendix.

\begin{figure}[h] 
\centering 
\includegraphics[width=1\linewidth, trim=10cm 0cm 10cm 0cm, clip]{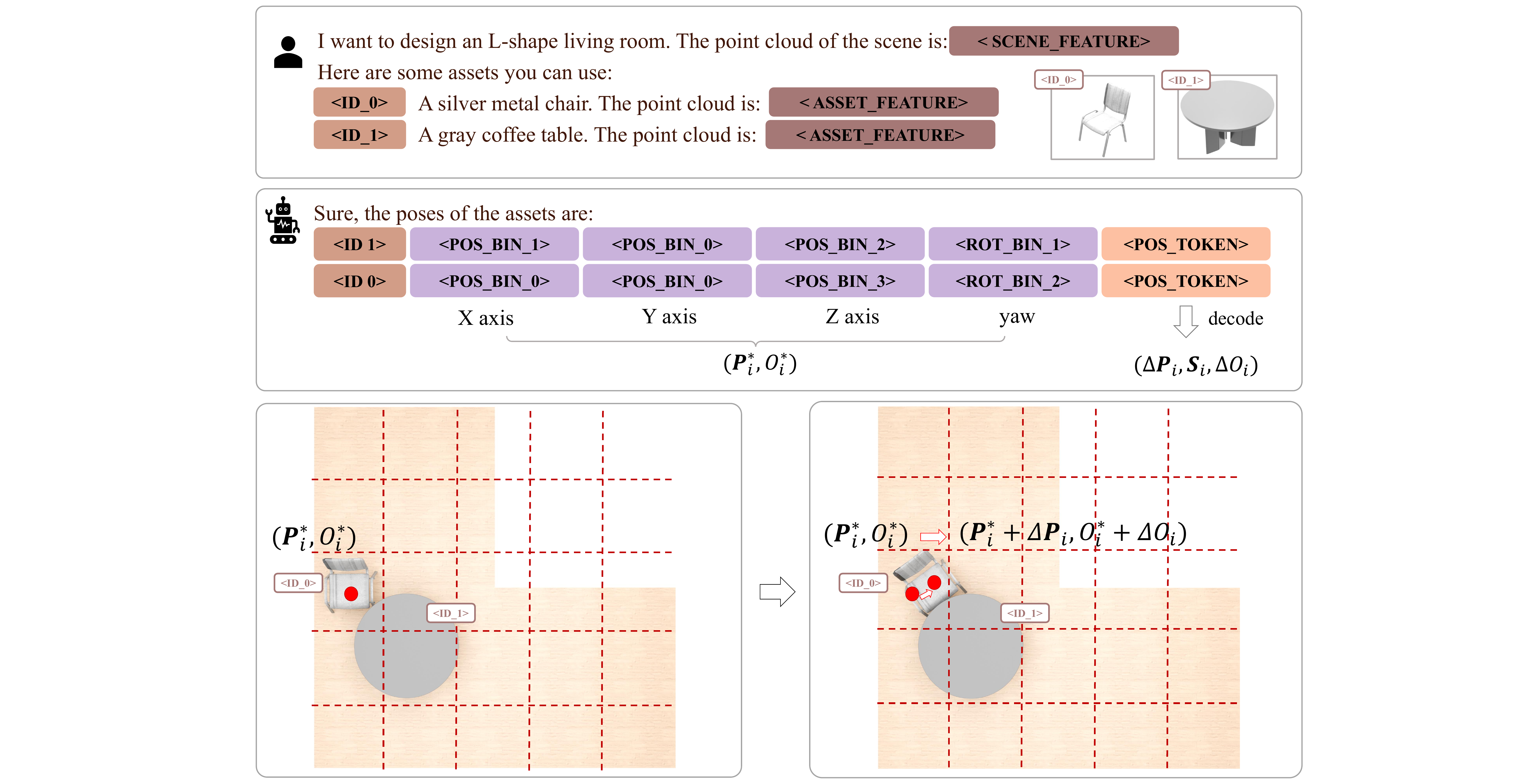} 
\caption{\textbf{Coarse-to-fine token design in NaLA.}
The first four tokens determine the coarse location and orientation of an asset, while the final regression tokens are decoded into fine-grained poses. Different assets are distinguished using different ID tokens. For clarity, we illustrate the mechanism in 2D.%\YS{Update}
} 
\label{fig: model output} 
\end{figure}

%\textbf{Coarse-to-Fine Pose Generation:} Leveraging both mechanisms, NaLA generates object poses through a rigorous coarse-to-fine procedure. For each asset $i$, the model first autoregressively predicts an anchoring sequence to determine the coarse global position and orientation:

%\textbf{Discussion:} One might question, if the regression token can precisely predict any coordinate, why are anchoring tokens necessary? Notably, we observe that relying solely on regression tokens leads to severe performance degradation: the model repeatedly places identical assets at the same position (see \cref{fig:three_comparisons}). We hypothesize that discrete tokens serve as explicit ``anchors'' for layout history. Without them, specific location information must be heavily compressed into the hidden states of the single $\langle\texttt{POS\_TOKEN}\rangle$. Since this token is reused across all generation steps, the LLM struggles to differentiate between previously placed assets and the current target, creating ambiguity in the autoregressive chain. This confirms that the combined strategy—using discrete tokens for global planning context and regression for local adjustment—is essential.

\textbf{Asset Identification:} Finally, as our model generates poses for all assets within a single output sequence, we introduce special ID tokens to distinguish which pose prediction corresponds to which asset. Specifically, assuming the model processes at most $K$ assets per scene, we register $K$ special ID tokens: $\langle\texttt{ID\_1}\rangle,\cdots,\langle\texttt{ID\_K}\rangle$. In the input prefix, each asset is assigned a unique ID token. 
%During autoregressive generation, 
Before generating the model's pose information, 
NaLA first emits the corresponding ID token. During decoding, this token is matched with the identical ID token in the prefix, thereby associating the predicted pose with the correct asset. \cref{fig: model output} reveals an example of NaLA's output structure.

%\begin{figure}[htbp] \centering \includegraphics[width=1\linewidth, page=5, trim=1cm 3cm 1cm 3cm, clip]{figure/figure.pdf} \caption{\textbf{Prompt Construction.} NaLA organizes all assets into a single prompt and distinguishes them using dedicated ID tokens. During generation, each predicted pose is associated with its corresponding asset through the emitted ID token. \YS{Merge to Fig 4}} \label{fig: prompt construction} \end{figure}

\subsection{Training Strategy}
\label{subsec: training strategy}

After designing the model architecture, we pretrain and fine-tune the model on high-quality layout datasets to endow it with spatial perception and reasoning capabilities. The model is trained in an end-to-end manner. As illustrated in \cref{fig: model output} (top), each empty scene together with its associated assets is organized into a prefix, while the poses of all assets are organized as the answer, forming a complete prompt–response pair.
Given the ground-truth scaling factor $\boldsymbol{S_i}$, position residual $\Delta \boldsymbol P_i$, and orientation residual  $\Delta O_i$, the overall layout loss is:
\begin{equation}
    \label{equ: loss}
    %\resizebox{\linewidth}{!}
    {
\begin{aligned}
\
\mathcal{L}_{\text{NaLA}}&=
\mathcal{L}_{\text{Cross-Entropy}}
+   %\frac{\lambda_1}{n} \sum_{i=1}^{n}
%\lambda_1 
\frac{\lambda_1}{n} \sum_{i=1}^{n}\left \lVert \Delta \widehat{\boldsymbol P}_i - \Delta \boldsymbol P_i \right\rVert_1 %
\\
 &+   %\frac{\lambda_2}{n} \sum_{i=1}^{n}
%\lambda_2
\frac{\lambda_2}{n} \sum_{i=1}^{n}\left\lVert \widehat{\boldsymbol S}_i - \boldsymbol S_i \right\rVert_1 
+  %\frac{\lambda_3}{n} \sum_{i=1}^{n}
%\lambda_3 
\frac{\lambda_3}{n} \sum_{i=1}^{n}\left(1-\cos\!\left(\Delta \widehat{O}_i - \Delta O_i\right)\right),
\end{aligned}
}
\end{equation}
where $\left \lVert\cdot\right\rVert_1$ denotes L1 loss, and $\lambda_1$, $\lambda_2$, and $\lambda_3$ are balancing coefficients. We next introduce our training strategy, including the processing and augmentation of prompt sequences.

\textbf{Processing Asset Prompt Sequence:} Our model autoregressively predicts asset poses one by one. This requires the model to learn a reasonable placement order during training. For example, predicting the pose of a table before predicting the pose of a vase placed on top of it is natural, whereas predicting the vase first would be unreasonable. To encourage such behavior, for each training sequence, we sort the assets in the answer by descending object volume, so that the model learns to place larger and more essential objects first.

\textbf{Data Augmentation:} High-quality layout datasets that are both physically and semantically plausible remain scarce. To mitigate this limitation, we apply aggressive data augmentation to each input sequence to ensure that the model effectively learns generalizable placement patterns. Specifically, we employ the following augmentation strategies: 
\textbf{(1) Input asset sequence shuffling.} The order of assets in the prefix is randomly permuted, while the asset order in the answer remains fixed. 
\textbf{(2) Scene rotation.} The ground-truth scene and all contained assets are jointly rotated by $0^\circ$, $90^\circ$, $180^\circ$, or $270^\circ$. 
\textbf{(3) Asset replacement.} 
%Assets in the ground-truth scene are replaced with other assets of the same category sampled from the asset library, encouraging the model to learn category-level placement rules rather than instance-specific patterns.
We pre-categorize assets and evaluate whether each asset in a scene is safely replaceable. 
During training, a replaceable asset is randomly swapped with another from the same class (updating its 3D/text features while keeping the exact pose). This prevents NaLA from overfitting to specific asset IDs.
\textbf{(4) Random omission of asset descriptions.} To force the model to rely on point cloud geometry rather than textual descriptions, 
% we randomly mask all textual descriptions of assets with a certain probability, retaining only the point clouds. 
for standard \textless 3D Input Tokens, Text\textgreater input, we randomly drop text descriptions to force the model to perceive 3D geometry (i.e., \textless 3D Input Token \textgreater), preventing it from relying on text-based shortcuts.

%% file: main_text/experiment.tex
\section{Experiments}

We conduct experiments to answer two key questions:
(1) whether our model outperforms commonly used baselines on layout generation tasks, and
(2) whether the proposed architectural designs and training strategies are effective.
We first introduce the experimental setup, baseline models, and evaluation metrics, followed by quantitative results and ablation studies.

\subsection{Model Settings and Training Strategy}
\label{subsec: model setting}

We implement NaLA using Qwen-2.5-7B-Instruct \cite{qwen2.5} as the LLM backbone. To perceive 3D geometry, we employ SPFormer and PointBERT as the scene-level and asset-level encoders, respectively. To maintain the LLM's general reasoning capabilities while adapting it to 3D tasks, we freeze the base LLM and both point cloud encoders; only the geometric projection adapters, the LoRA modules, and the pose decoder head are trainable. The training objective follows \cref{equ: loss}.

To address the scarcity of high-quality fine-grained 3D data, we adopt a two-stage coarse-to-fine training curriculum. First, we pretrain the model on the large-scale 3D-FRONT dataset \cite{fu20213d} to learn macro-level furniture arrangement rules (e.g., ensuring beds and tables are placed without collision). Second, we fine-tune the model on the Imaginarium dataset \cite{zhu2025imaginarium}—which contains significantly richer asset diversity and clutter details—to master micro-level object placement and stylistic refinement. We maintain an 80\%–20\% train-test split for asset libraries to prevent data leakage. Detailed hyperparameters and splitting strategies are provided in the appendix.

\subsection{Test Dataset and Baselines}
\label{subsec: baselines}

In test cases, each model is given the room type, the same empty-room description, and identical asset information. All assets are sourced from the test asset library of 3D-FRONT and Imaginarium. An independent AI judge selects suitable asset categories for each scene type, after which specific assets are retrieved from the asset test set; both empty rooms and candidate assets are further reviewed by human annotators. We evaluate 20 scene categories, generating three test cases per category, each containing on average 20 objects to be placed. Detailed evaluation procedures are provided in the appendix.

For baselines, we compare against three representative LLM/VLM-based layout generation methods: LayoutGPT \cite{LayoutGPT}, Holodeck \cite{holodeck}, and LayoutVLM \cite{sun2025layoutvlmdifferentiableoptimization3d}, which either leverage an LLM or VLM for layout generation. %In addition, we include ATISS, a classical Transformer-based model designed for learning indoor layout priors. %All baseline models are implemented using the configurations reported in their original papers.% 
All baselines are evaluated using their official configurations. Additional implementation details are provided in the appendix.

\subsection{Evaluation Metrics}
\label{subsec: metrics}
Following evaluation procedures in \cite{I-Design_elen_2025,ling2025scenethesis,sun2025layoutvlmdifferentiableoptimization3d}, we consider three metrics: 
%\YS{cite the metrics used in other paper}
\textbf{(1) Physical Plausibility}, whether the placed assets satisfy basic physical constraints, including absence of collisions, floating objects, or out-of-bound placements.
\textbf{(2) Semantic Plausibility}, whether the arrangement aligns with common human usage patterns and everyday conventions.
\textbf{(3) Aesthetics}, whether the layout appears visually appealing and natural.
Specifically, physical plausibility is quantitatively evaluated by computing voxel-level collision ratio, floating rate, and out-of-boundary (OOB) rate. For all three evaluation dimensions, we employ Gemini 3.0-Flash-Preview \cite{gemini_team2024gemini} as the AI judge and recruited twenty graduate students as human evaluators. Both AI and human judges score each scene on a 1–5 Likert scale \cite{likert1932technique} using a comprehensive, unified questionnaire. %We further conduct Kendall’s tau test \cite{kendall1938new} to measure the consistency between AI and human ratings, thereby assessing the reliability and validity of the evaluation. 
Details of the questionnaire and the computation of each metric are in the appendix.

\begin{figure*}[t]
\centering
\captionsetup[subfigure]{justification=centering}
% Row 1 (Scene type 1)
\begin{subfigure}[t]{0.24\textwidth}
  \centering
  \includegraphics[width=\linewidth,trim=2cm 14cm 2cm 1cm,clip]{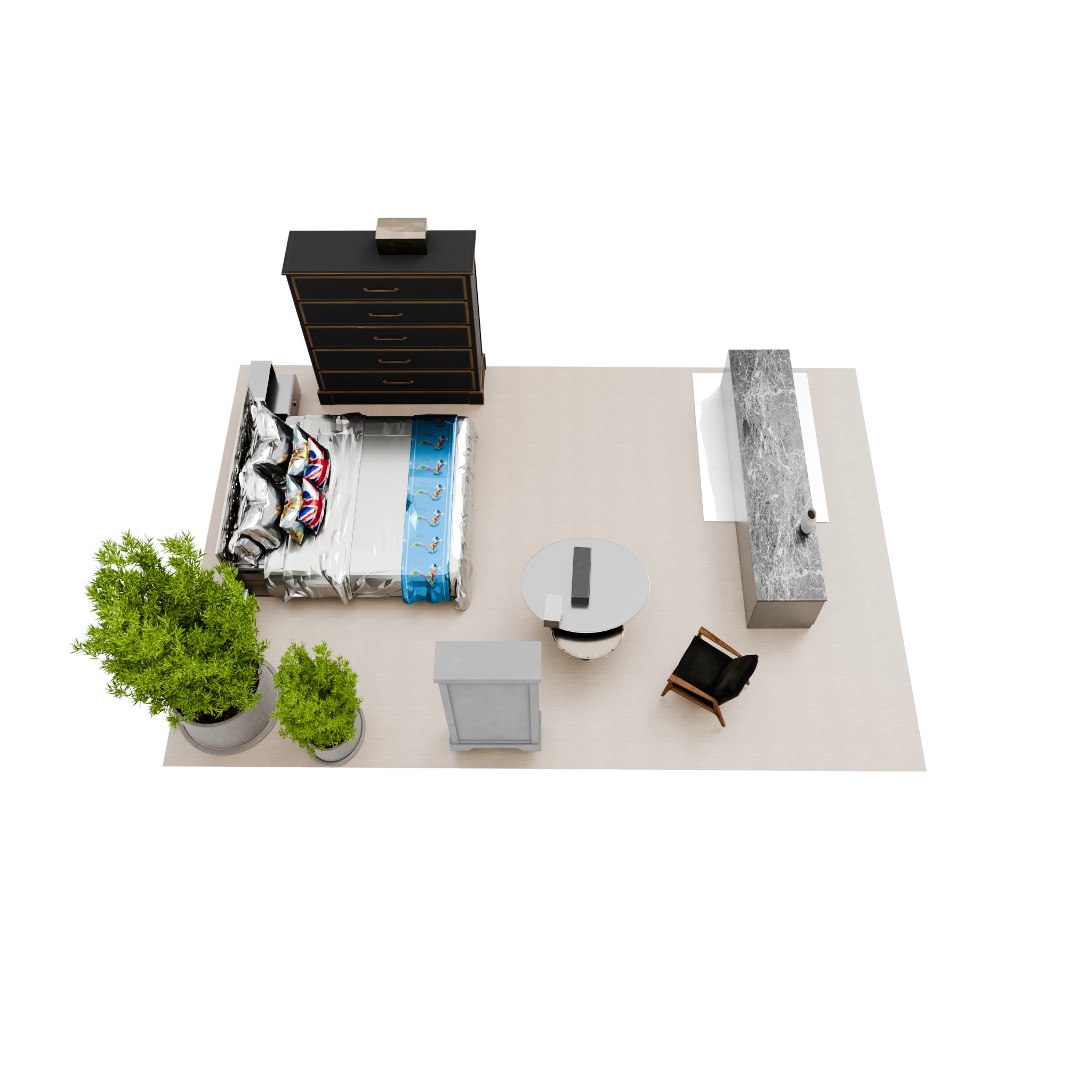}
\end{subfigure}\hfill
\begin{subfigure}[t]{0.24\textwidth}
  \centering
  \includegraphics[width=\linewidth,trim=2cm 14cm 2cm 1cm,clip]{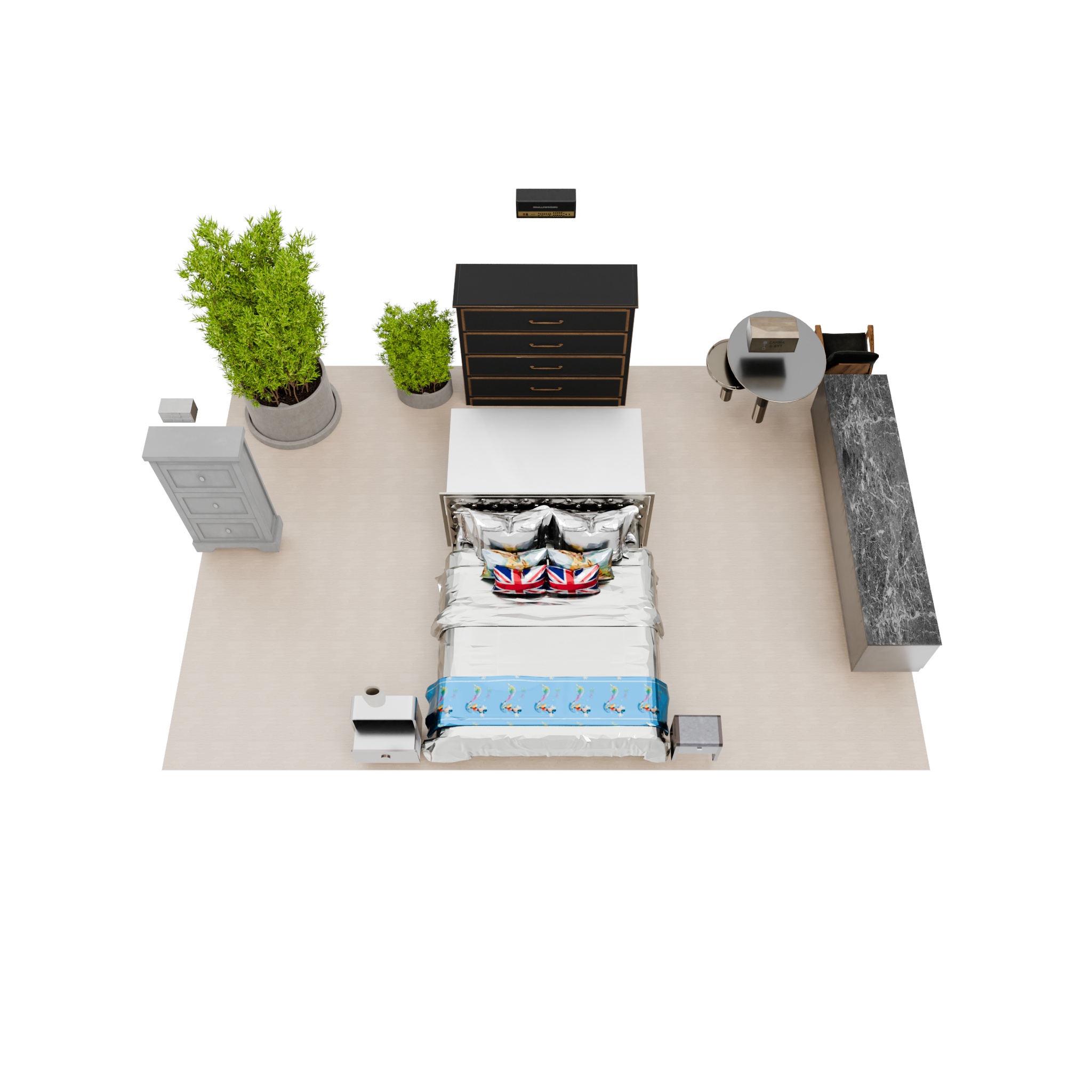}
\end{subfigure}\hfill
\begin{subfigure}[t]{0.24\textwidth}
  \centering
  \includegraphics[width=\linewidth,trim=2cm 14cm 2cm 1cm,clip]{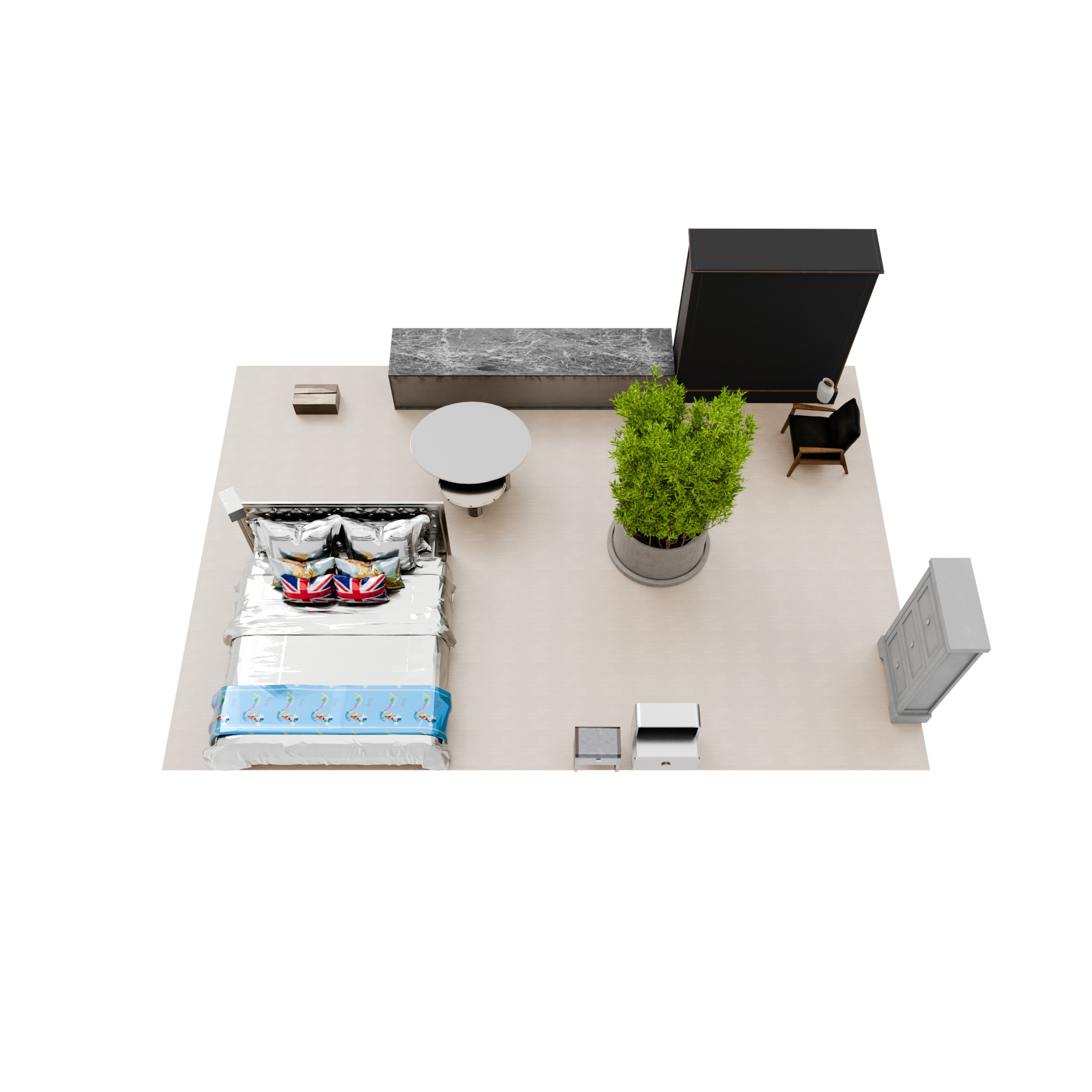}
\end{subfigure}\hfill
\begin{subfigure}[t]{0.24\textwidth}
  \centering
  \includegraphics[width=\linewidth,trim=2cm 14cm 2cm 1cm,clip]{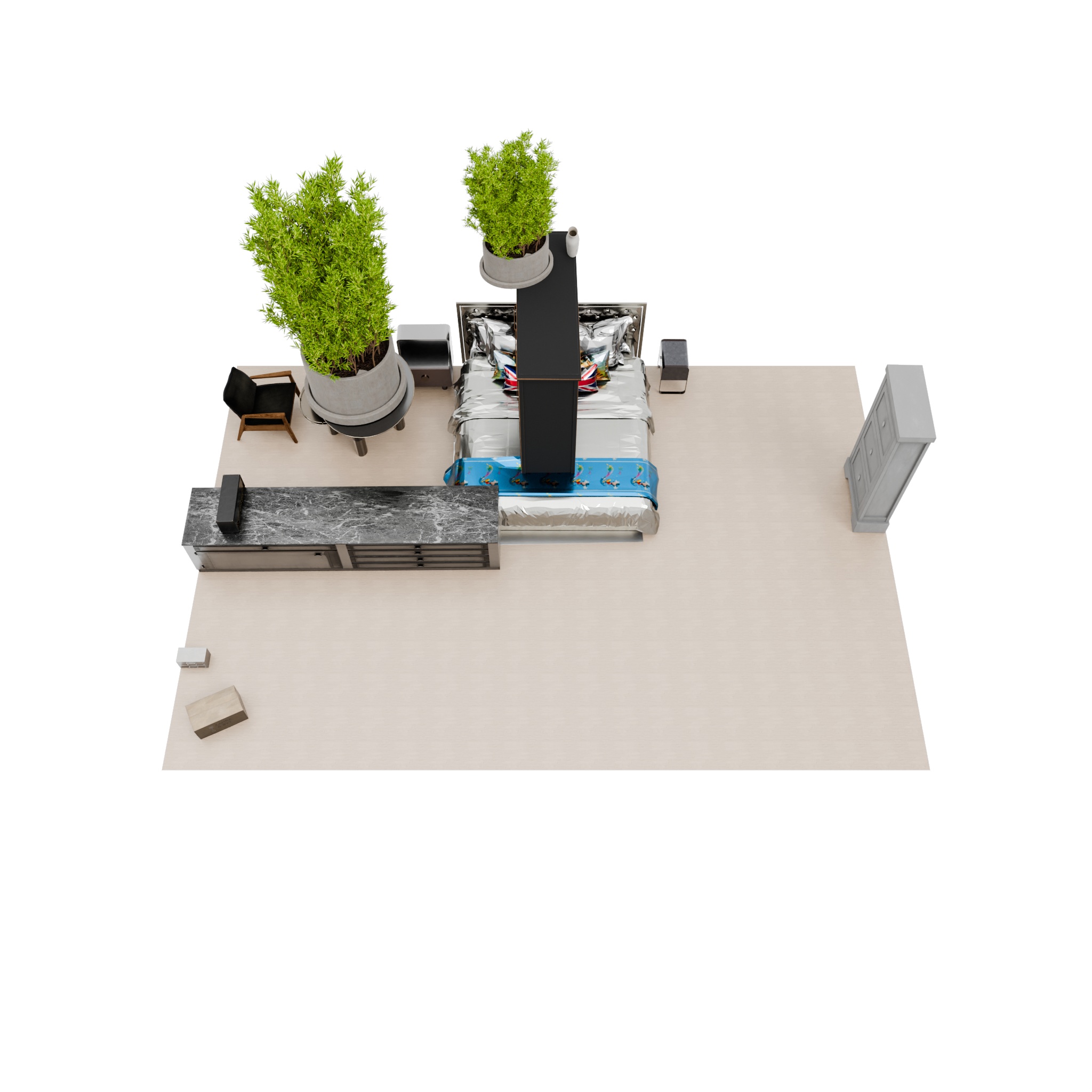}
\end{subfigure}
\vspace{-3mm}

% Row 2 (Scene type 2)
\begin{subfigure}[t]{0.24\textwidth}
  \centering
  \includegraphics[width=\linewidth,trim=2cm 14cm 2cm 1cm,clip]{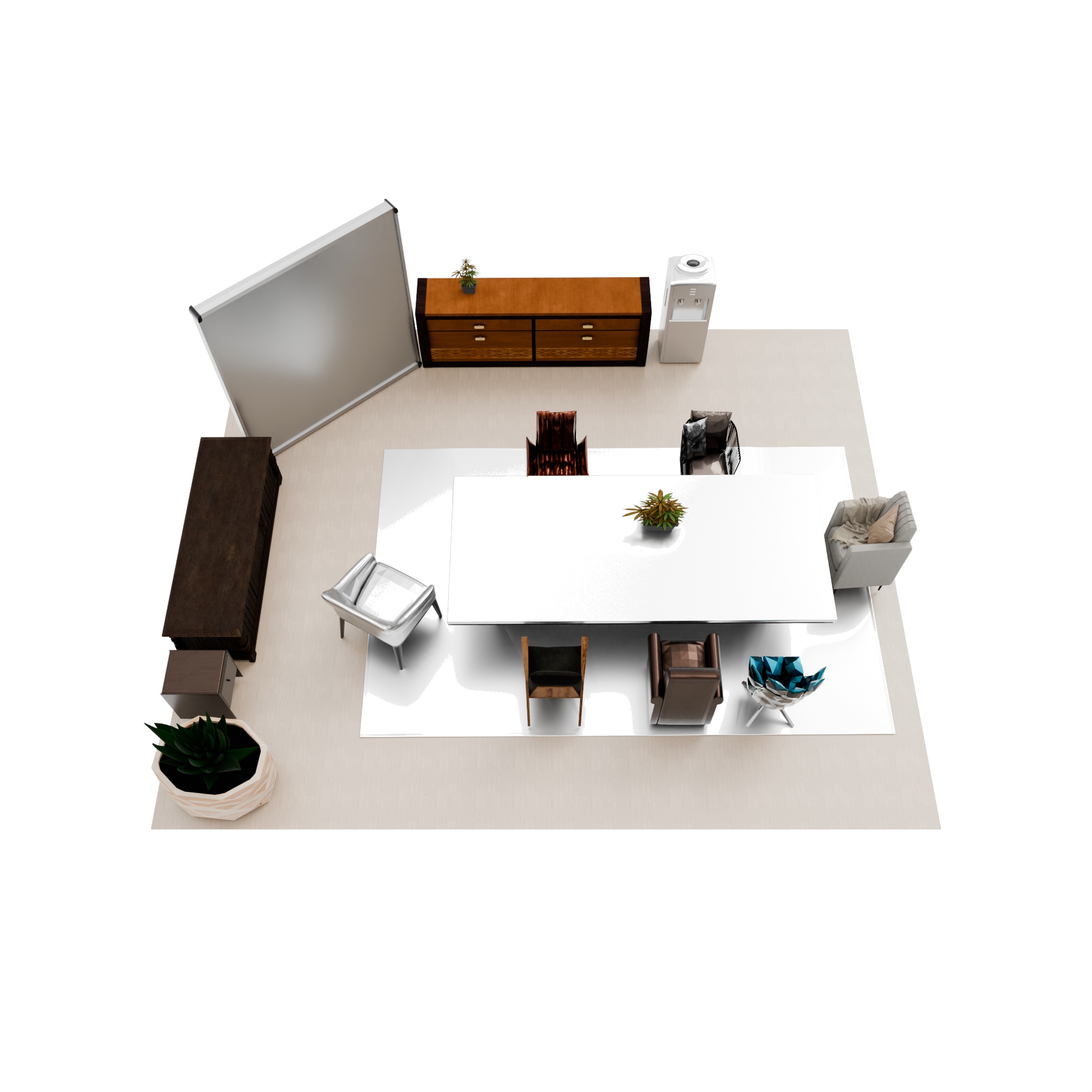}
\end{subfigure}\hfill
\begin{subfigure}[t]{0.24\textwidth}
  \centering
  \includegraphics[width=\linewidth,trim=2cm 14cm 2cm 1cm,clip]{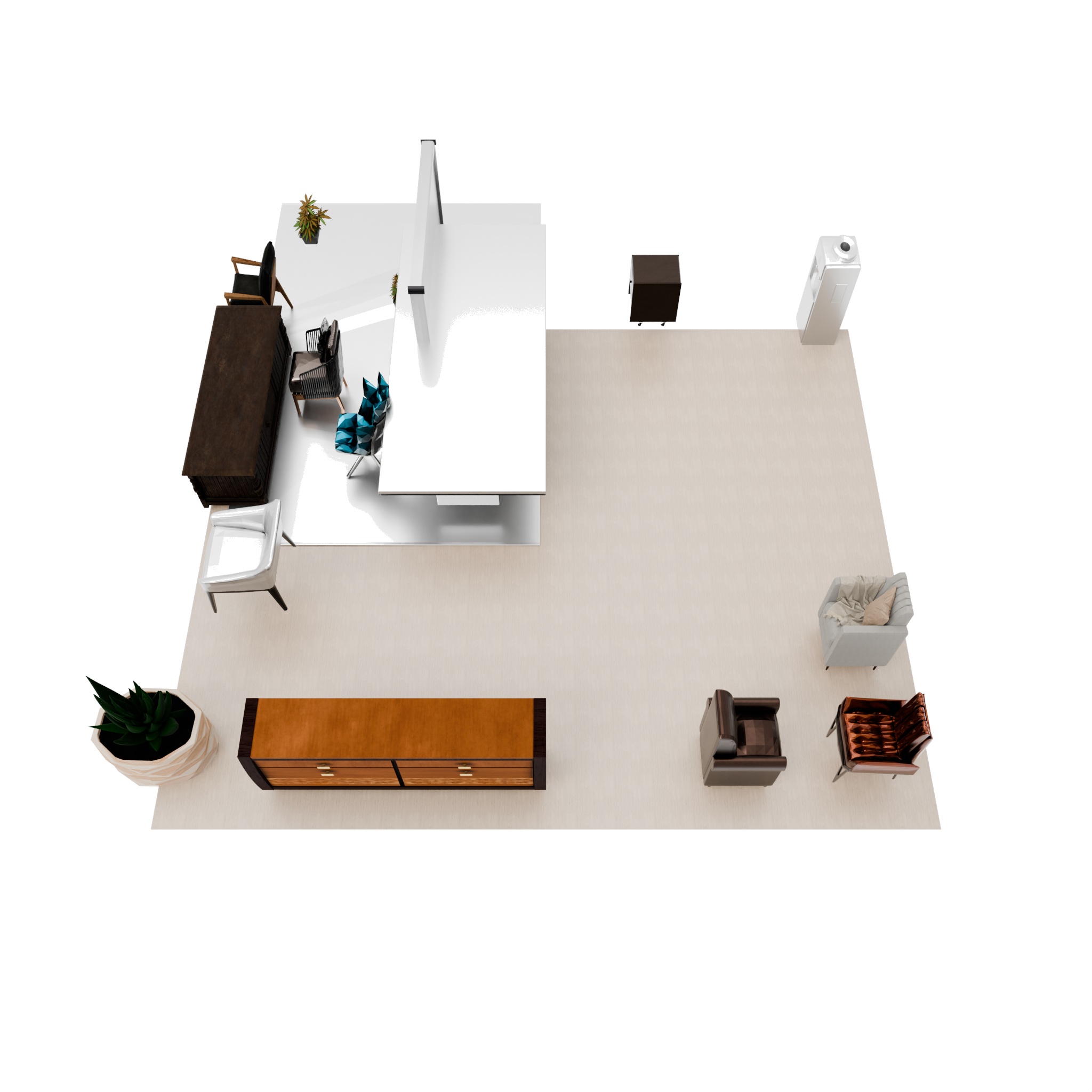}
\end{subfigure}\hfill
\begin{subfigure}[t]{0.24\textwidth}
  \centering
  \includegraphics[width=\linewidth,trim=2cm 14cm 2cm 1cm,clip]{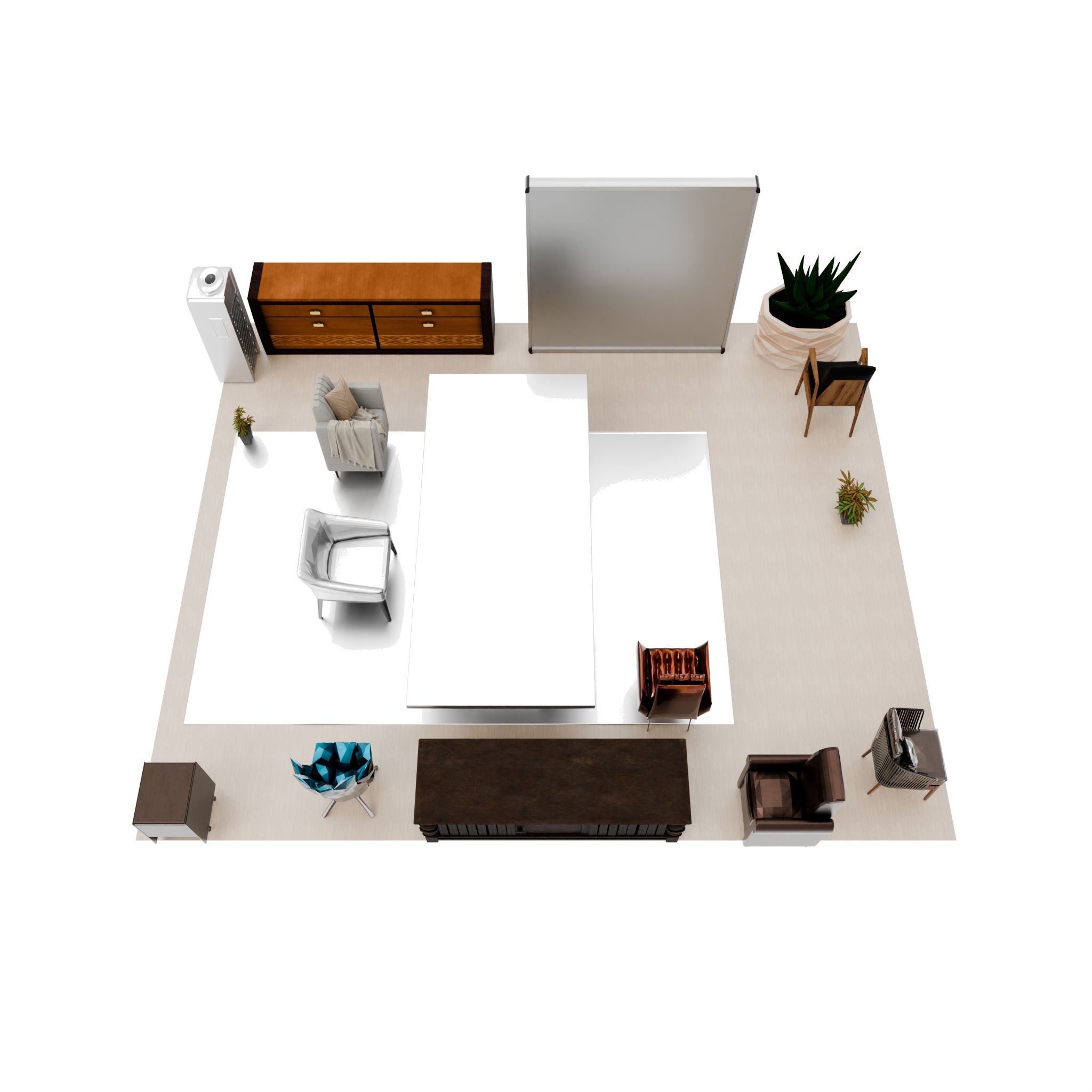}
\end{subfigure}\hfill
\begin{subfigure}[t]{0.24\textwidth}
  \centering
  \includegraphics[width=\linewidth,trim=2cm 14cm 2cm 1cm,clip]{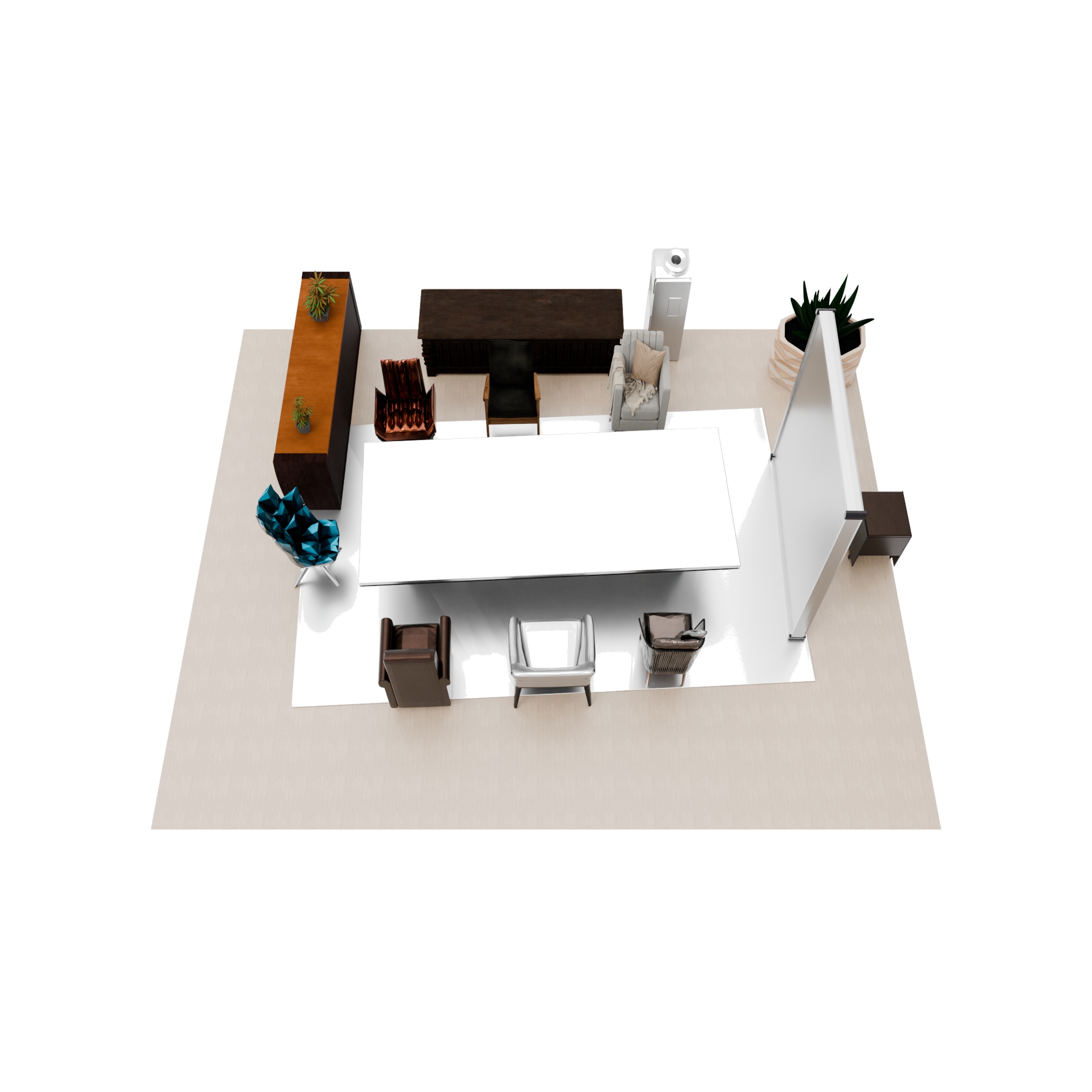}
\end{subfigure}

%\vspace{2mm}

% Row 3 (Scene type 3)
\begin{subfigure}[t]{0.24\textwidth}
  \centering
  \includegraphics[width=\linewidth,trim=2cm 14cm 2cm 1cm,clip]{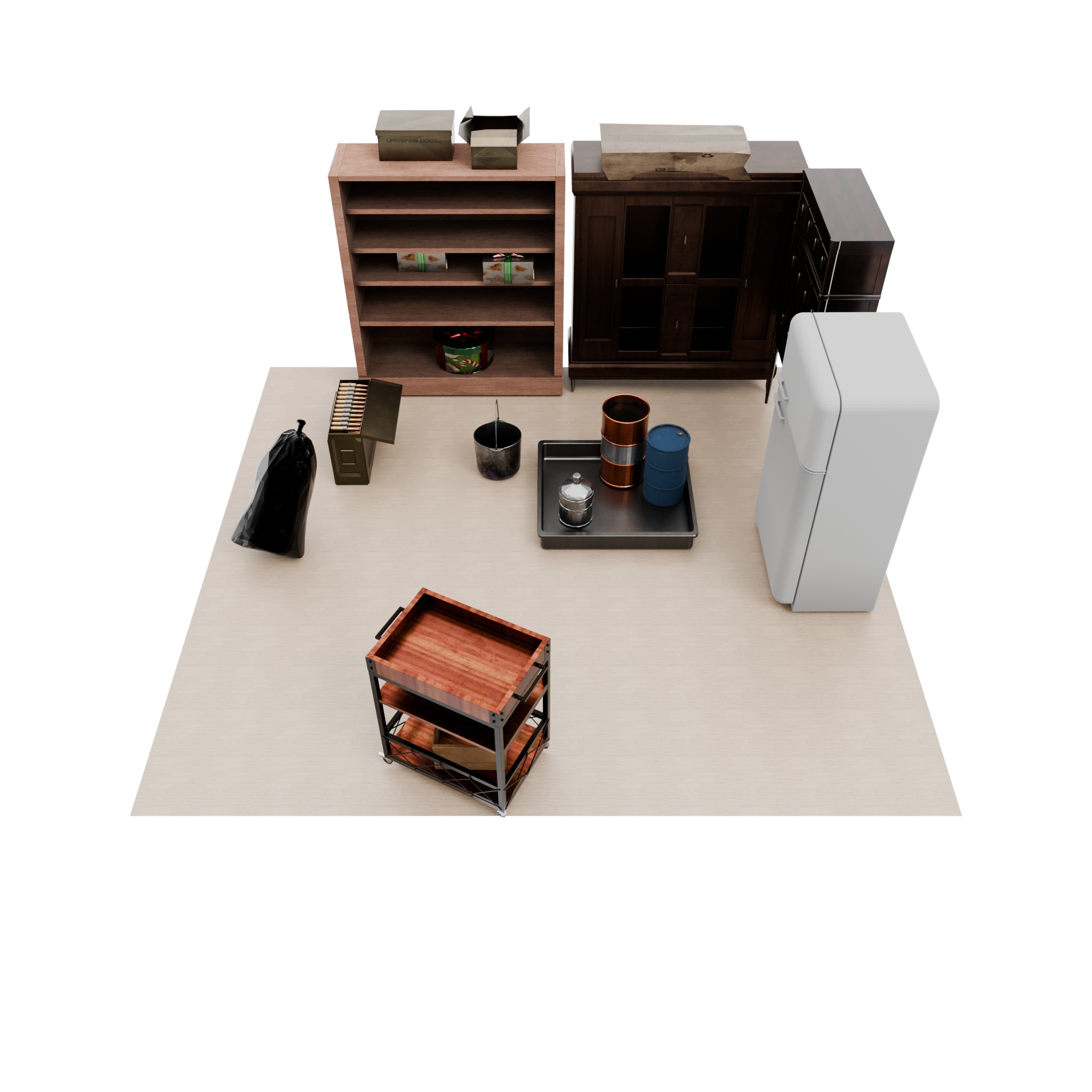}
\end{subfigure}\hfill
\begin{subfigure}[t]{0.24\textwidth}
  \centering
  \includegraphics[width=\linewidth,trim=2cm 14cm 2cm 1cm,clip]{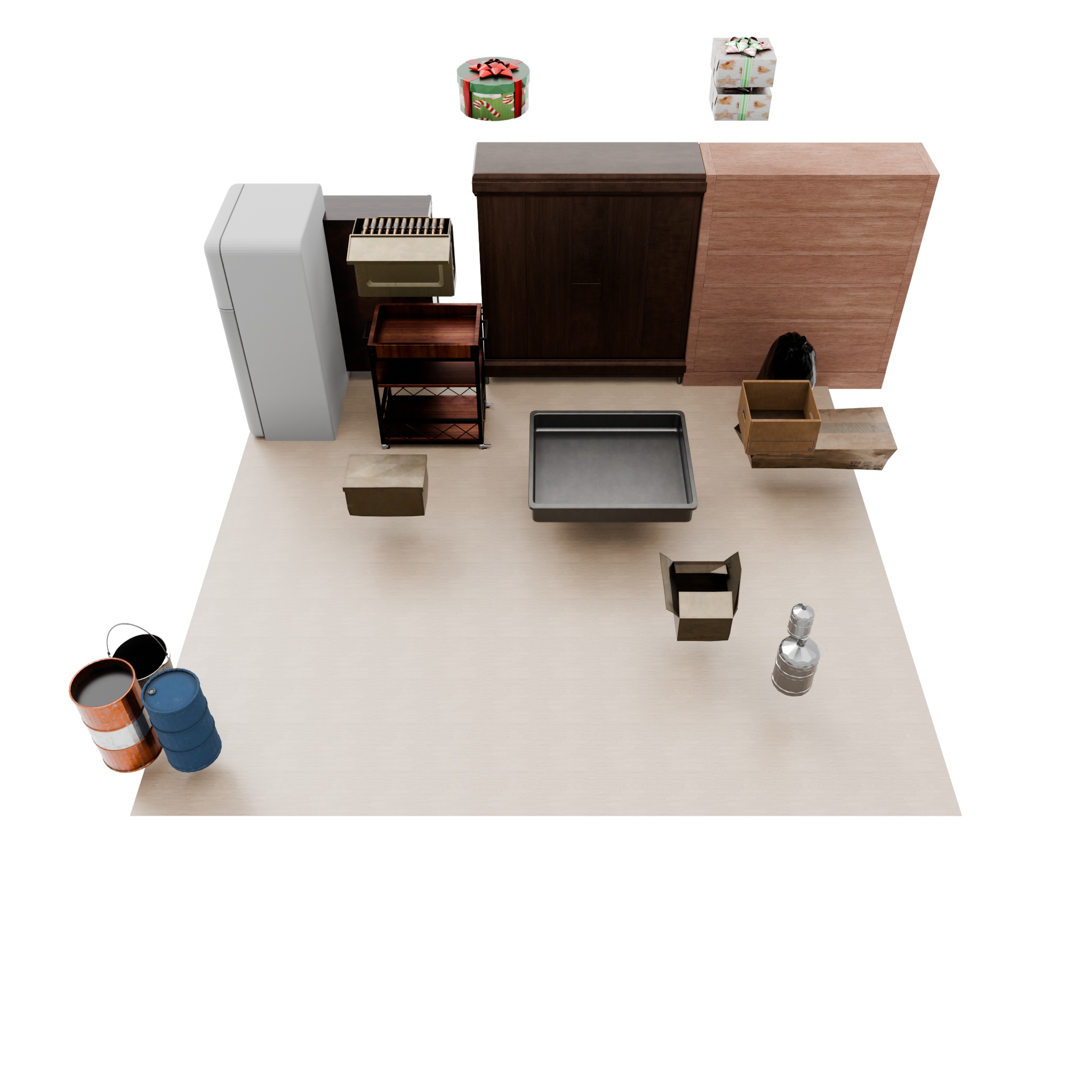}
\end{subfigure}\hfill
\begin{subfigure}[t]{0.24\textwidth}
  \centering
  \includegraphics[width=\linewidth,trim=2cm 14cm 2cm 1cm,clip]{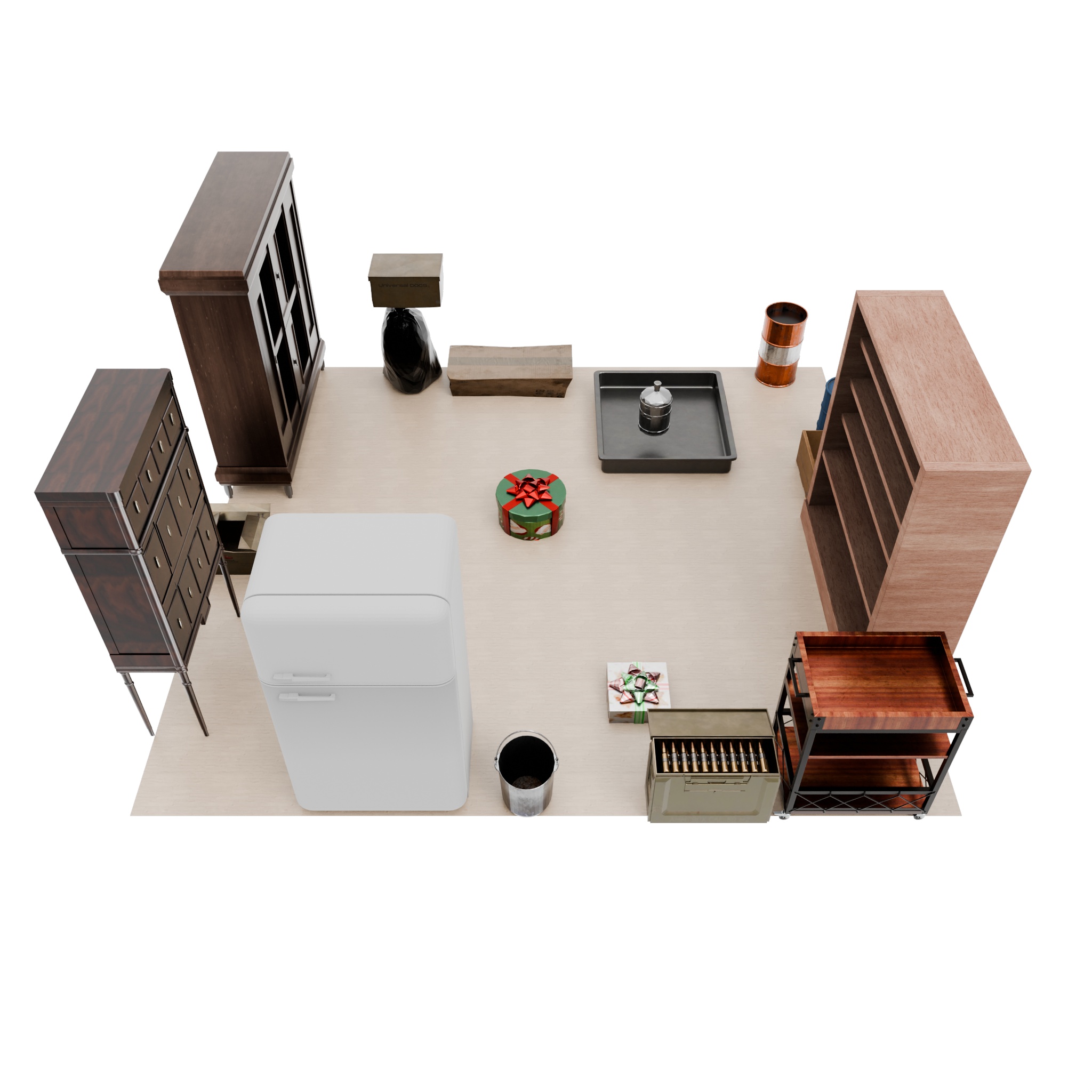}
\end{subfigure}\hfill
\begin{subfigure}[t]{0.24\textwidth}
  \centering
  \includegraphics[width=\linewidth,trim=2cm 14cm 2cm 1cm,clip]{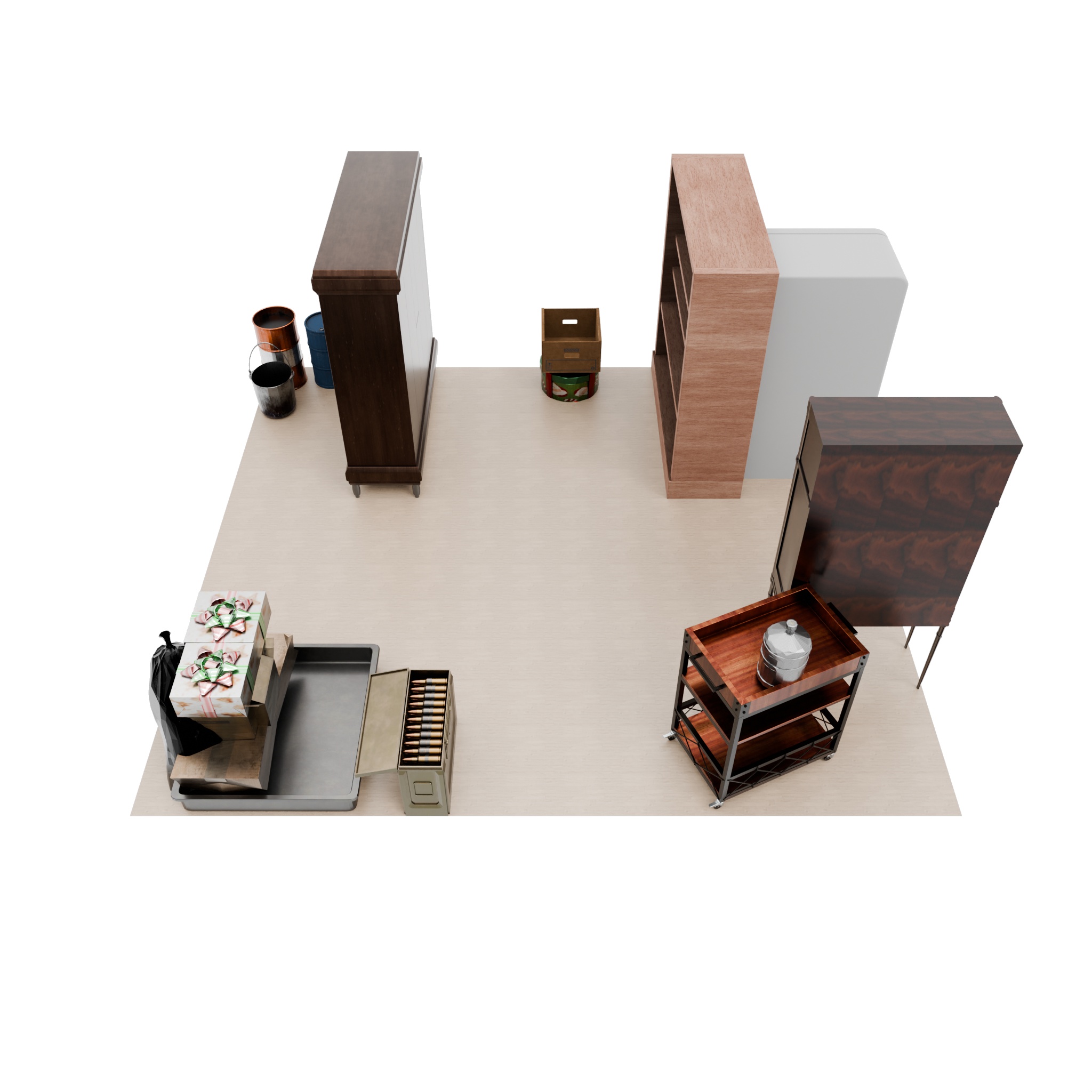}
\end{subfigure}

% \vspace{2mm}
% Method names (one row for columns)
\begin{minipage}[t]{0.24\textwidth}
  \centering NaLA
\end{minipage}\hfill
\begin{minipage}[t]{0.24\textwidth}
  \centering LayoutGPT
\end{minipage}\hfill
\begin{minipage}[t]{0.24\textwidth}
  \centering Holodeck
\end{minipage}\hfill
\begin{minipage}[t]{0.24\textwidth}
  \centering LayoutVLM
\end{minipage}

\caption{Qualitative comparison of layout generation. Each row shows top-down views for one scene type (bedroom, conference room, storage); each column shows results from a different method under identical room and asset conditions. NaLA produces physically plausible and semantically coherent arrangements.}
\label{fig:visual_grid}
\end{figure*}

\begin{table}[htbp]
\centering
\small
\caption{Quantitative comparison against baseline methods. Bold indicates the best result and underlined indicates the second best. Physical metrics (Collision, OOB, Floating) are lower-is-better ($\downarrow$); AI and human evaluation scores (1--5 scale) are higher-is-better ($\uparrow$). Standard errors are reported in parentheses, estimated via 1,000 bootstrap resamples \cite{efron1979bootstrap}. $^\dagger$Methods optimize over predefined plausibility constraints.}
\label{tab:main_comparison}
\begin{tabular}{ll@{\hskip 0.5em}rcccc}
\toprule
& \multicolumn{2}{l}{Metrics} & \multicolumn{1}{c}{NaLA} & \multicolumn{1}{c}{LayoutGPT} & \multicolumn{1}{c}{Holodeck$^\dagger$} & \multicolumn{1}{c}{LayoutVLM$^\dagger$} \\
\midrule
\multirow{5}{*}{\shortstack[l]{{Physical}\\ {Plausibility}}}
 & Collision (\%) & $\downarrow$ & \underline{0.86}$_{(0.11)}$ & 2.29$_{(0.34)}$ & \textbf{0.10}$_{(0.02)}$ & 1.18$_{(0.29)}$ \\
 & OOB (\%) & $\downarrow$ & \textbf{1.16}$_{(0.28)}$ & 18.19$_{(1.21)}$ & 10.14$_{(0.63)}$ & \underline{5.19}$_{(1.12)}$ \\
 & Floating (\%) & $\downarrow$ & 3.36$_{(0.78)}$ & 15.12$_{(1.51)}$ & \textbf{0.00}$_{(0.00)}$ & \underline{1.43}$_{(0.74)}$ \\
\cmidrule(lr){2-7}
 & AI & $\uparrow$ & \underline{4.24}$_{(0.13)}$ & 3.34$_{(0.13)}$ & \textbf{4.27}$_{(0.11)}$ & 4.17$_{(0.16)}$ \\
 & Human & $\uparrow$ & \textbf{4.17}$_{(0.18)}$ & 2.43$_{(0.26)}$ & 3.91$_{(0.19)}$ & \underline{3.93}$_{(0.25)}$ \\
\midrule
\multirow{2}{*}{\shortstack[l]{{Semantic}\\ {Plausibility}}}
 & AI & $\uparrow$ & \textbf{3.32}$_{(0.16)}$ & 2.62$_{(0.09)}$ & 2.68$_{(0.09)}$ & \underline{2.92}$_{(0.12)}$ \\
 & Human & $\uparrow$ & \textbf{3.48}$_{(0.19)}$ & 2.17$_{(0.19)}$ & 3.03$_{(0.22)}$ & \underline{3.37}$_{(0.23)}$ \\
\midrule
\multirow{2}{*}{\shortstack[l]{{Visual}\\ {Aesthetics}}}
 & AI & $\uparrow$ & \textbf{3.05}$_{(0.16)}$ & 2.19$_{(0.08)}$ & 2.44$_{(0.07)}$ & \underline{2.63}$_{(0.18)}$ \\
 & Human & $\uparrow$ & \textbf{3.42}$_{(0.19)}$ & 2.49$_{(0.17)}$ & 2.94$_{(0.23)}$ & \underline{3.10}$_{(0.21)}$ \\
\midrule
\multirow{2}{*}{\shortstack[l]{{Overall}\\ {(Mean)}}}
 & AI & $\uparrow$ & \textbf{3.53}$_{(0.13)}$ & 2.72$_{(0.09)}$ & 3.13$_{(0.08)}$ & \underline{3.24}$_{(0.13)}$ \\
 & Human & $\uparrow$ & \textbf{3.69}$_{(0.11)}$ & 2.36$_{(0.12)}$ & 3.30$_{(0.13)}$ & \underline{3.47}$_{(0.14)}$ \\
\bottomrule
\end{tabular}
\end{table}

\subsection{Results}

\cref{tab:main_comparison} reports quantitative results, and \cref{fig:visual_grid} presents qualitative comparisons. NaLA achieves the best overall performance, balancing physical precision with semantic richness.
In terms of physical plausibility, although Holodeck and LayoutVLM achieve slightly better or comparable scores to NaLA on collision, floating rate, and AI evaluation, this is largely because both methods explicitly optimize for these physical metrics through iterative refinement procedures.
In contrast, NaLA operates in a single pass without any post-optimization. Despite this, it maintains a small collision ratio and achieves a significantly lower OOB rate than baselines, demonstrating a precise grasp of scene boundaries via point cloud encoding.
Moreover, NaLA outperforms all baselines in semantic plausibility and visual aesthetics. By perceiving asset geometry directly, it handles complex spatial relations naturally—such as correctly orienting chairs to face a round table (Row 1, Col 1 in \cref{fig:visual_grid}) or placing boxes inside a cabinet (Row 3, Col 1 in \cref{fig:visual_grid}). This ability to capture implicit spatial patterns results in layouts that are not just physically valid but also functionally coherent, corroborated by the significantly higher human and AI preference scores. % (Kendall’s $\tau$=0.55, p<0.05).
Additionally, we further compare NaLA with more sophisticated and larger-scale layout agents, such as DirectLayout \cite{ran2026direct}. Additional experimental results and analyses can be found in the appendix.
% Adding
%In the appendix

\subsection{Irregular Scene Evaluation}
%To examine whether incorporating scene point cloud features improves model performance, w
We further evaluate asset placement in irregular scenes. When the room floor is replaced with an irregular prism, we compare NaLA with LayoutGPT in terms of OOB ratio (Holodeck and LayoutVLM are excluded from this test because their optimization pipelines fundamentally assume box-shaped room templates). As shown in \cref{fig:vis_geo,fig:vis_gpt,fig:chart_oob}, NaLA achieves a substantially lower OOB rate. This advantage stems from our point cloud encoder, which perceives the precise boundary details of the irregular room, whereas generic LLMs struggle to infer complex boundaries from text alone. Test details refer to the appendix.

\subsection{Efficiency Evaluation}
\begin{figure*}[h]
    \centering
    \begin{subfigure}[b]{0.24\textwidth}
        \centering
        \includegraphics[width=\linewidth, trim=8cm 16cm 8cm 16cm, clip]{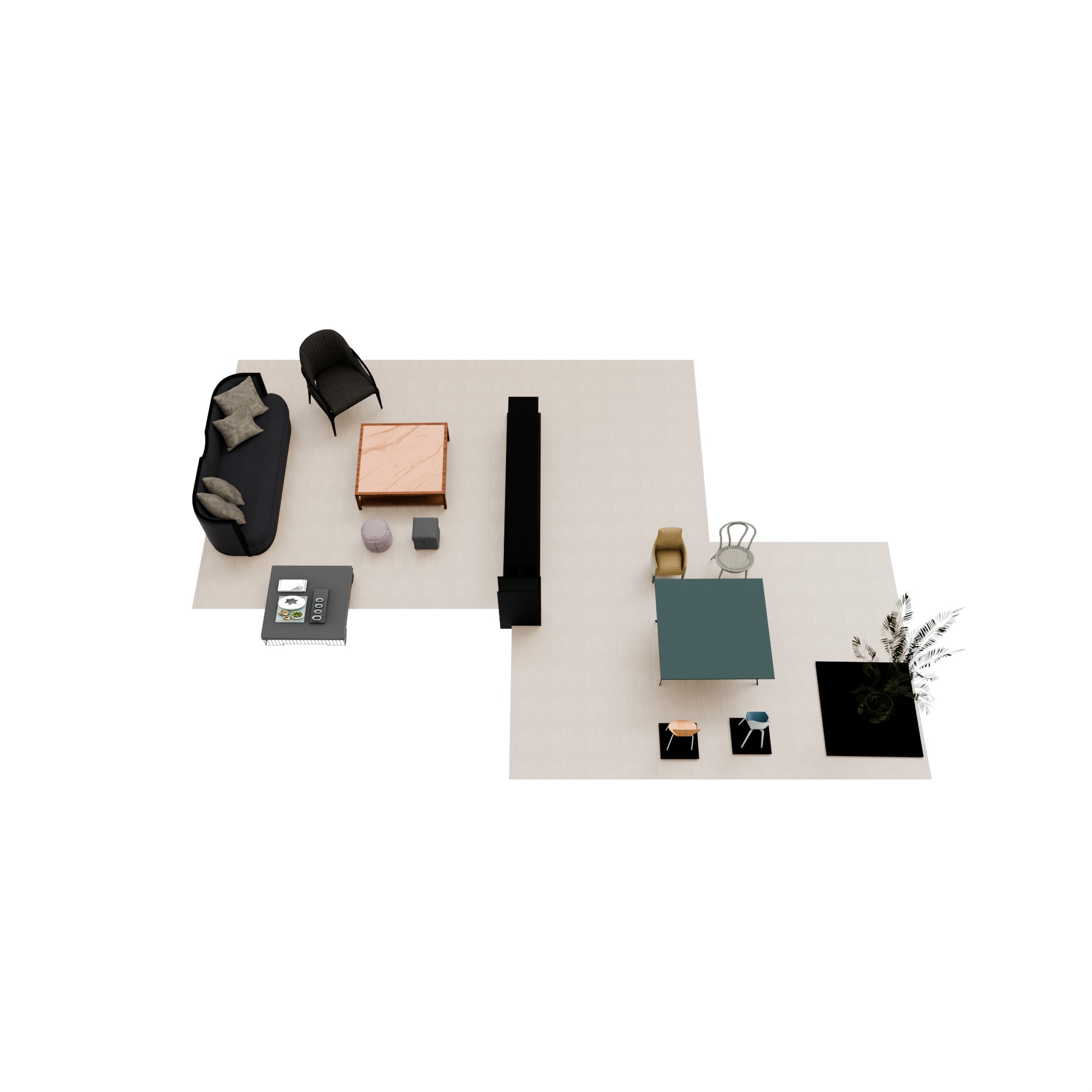}
        \caption{NaLA}
        \label{fig:vis_geo}
    \end{subfigure}
    \hfill
    \begin{subfigure}[b]{0.24\textwidth}
        \centering
        \includegraphics[width=\linewidth, trim=8cm 16cm 8cm 16cm, clip]{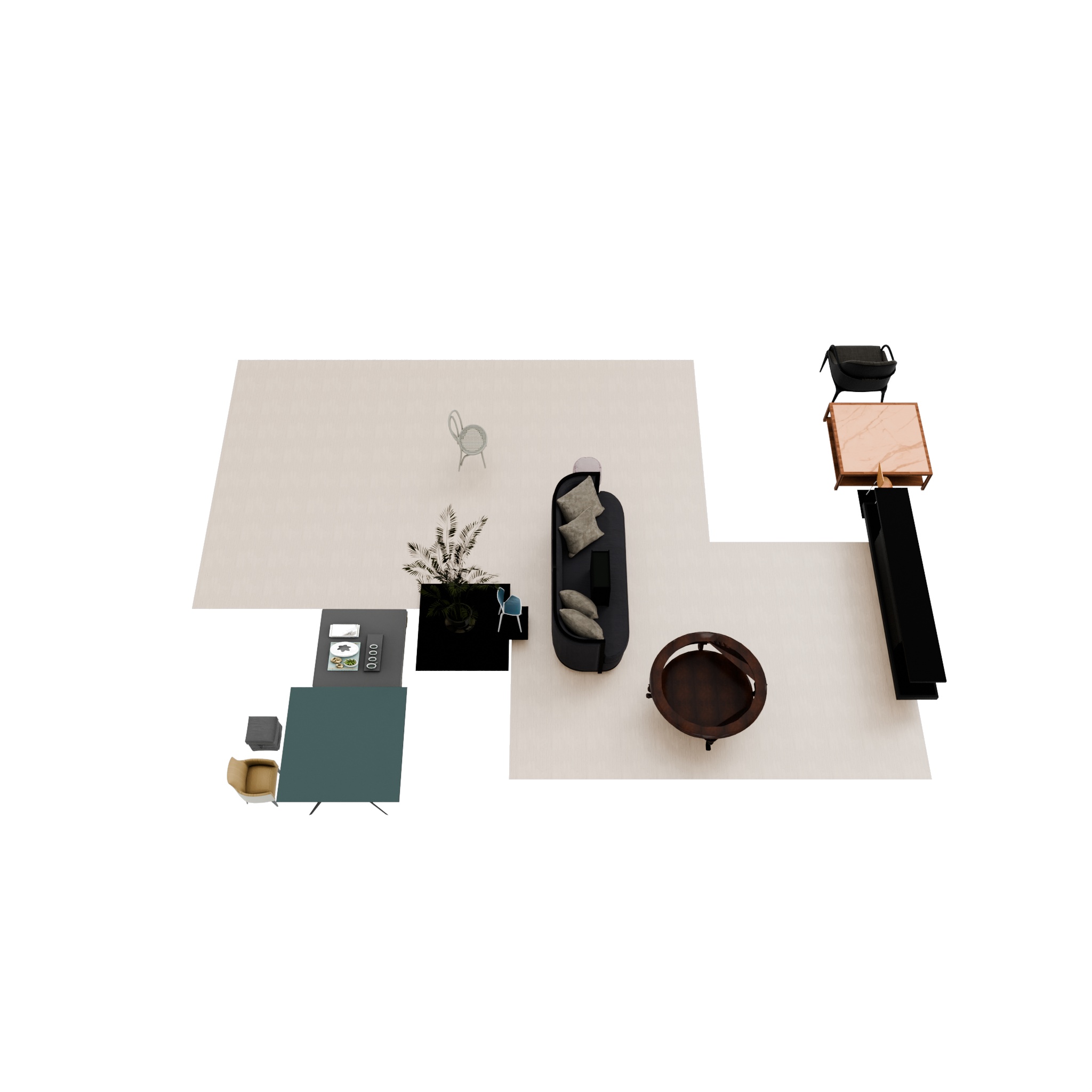}
        \caption{LayoutGPT}
        \label{fig:vis_gpt}
    \end{subfigure}
    \hfill
    \begin{subfigure}[b]{0.24\textwidth}
        \centering
        \includegraphics[width=\linewidth]{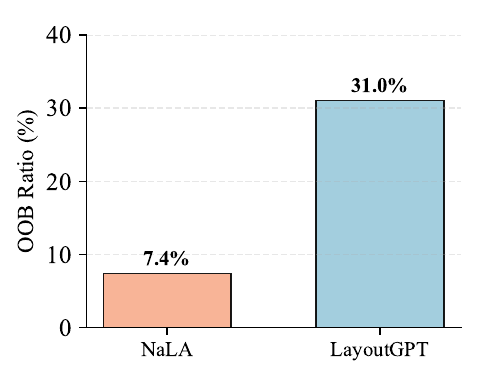}
        \caption{OOB Ratio}
        \label{fig:chart_oob}
    \end{subfigure}
    \hfill
    \begin{subfigure}[b]{0.24\textwidth}
        \centering
        \includegraphics[width=\linewidth]{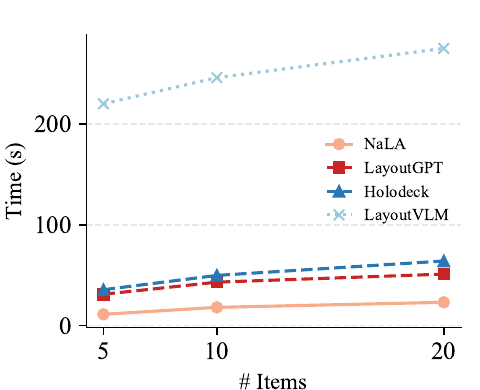}
        \caption{Inference Time}
        \label{fig:chart_time}
    \end{subfigure}
    
    \vspace{-2mm}
    \caption{Irregular-scene evaluation and inference efficiency. (a--b) Top-down views from NaLA and LayoutGPT under the same irregular room; NaLA keeps objects within the boundary while LayoutGPT exhibits out-of-bounds placements. (c) Out-of-bounds (OOB) ratio on irregular scenes. (d) Inference time vs.\ number of items across methods. NaLA achieves lower OOB and faster inference.}
    \vspace{-4mm}
\end{figure*}

Finally, we analyze inference efficiency across four models. %\cref{fig:chart_time} illustrates the runtime trends of different models as the number of assets increases. NaLA is significantly faster than all baselines, owing to its compact and efficient pose prediction framework. %Details of both tests refer to the appendix.
\cref{fig:chart_time} illustrates the runtime trends of different models as the number of assets increases.
NaLA significantly outperforms the other three baselines in terms of generation speed. Methods such as LayoutVLM rely on repeated optimization iterations, which are computationally expensive and time-consuming. In contrast, NaLA adopts an end-to-end generation strategy and requires only five tokens to represent the full pose of each asset. This compact and efficient representation substantially improves generation efficiency.

%\subsection{Human Evaluation}
%To verify whether the judgments produced by the GPT-based evaluator are consistent with human perception, we conduct a human evaluation study. We invite $X$ professional art and design students to score scenes generated by all baseline models. The evaluation questionnaire is provided in the appendix. The results of the human evaluation are reported in Table~X, along with Kendall’s $\tau$ correlation statistics between human ratings and GPT-based scores. The results show that human judgments consistently favor our model and exhibit strong agreement with GPT-based evaluations.

\subsection{Ablation Study}

In the ablation study, we answer three questions:
(1) whether incorporating point cloud inputs is effective,
(2) whether the proposed coarse-to-fine generation strategy improves performance, and
(3) whether the staged training strategy and data augmentation techniques are beneficial.
%Our standard model follows the configuration in \cref{subsec: model setting}. It takes both asset text descriptions and point cloud inputs, adopts the coarse-to-fine output strategy with $B=20$ spatial grids, and applies all training strategies described in \cref{subsec: training strategy}. During training, the textual description of each asset is randomly masked with a probability of 50\% to encourage reliance on geometric features.
In all experiments, we vary one factor and retrain the model in the same standard experiment setting. We then rerun the evaluation pipeline and examine the change in model performance.

\textbf{Point Cloud Input}: We compare the (I) full model with (II) a text-only model trained without point cloud inputs.
%; and (III) a full-modality model trained with both text and point clouds but without random text masking. 
At inference, we evaluate three settings: (a) standard input (text + point cloud); (b) text only; and (c) point cloud only.
%; and (d) text + point cloud with 30\% randomly missing point cloud data.
Average AI-judge scores are reported in \cref{tab:ablation_input}. 
Model (II) consistently underperforms across all scenarios, demonstrating the necessity of geometric input. 
\begin{table}[]
\centering
\small

\caption{Ablation study on Model Input. Average AI scores are reported. The full model consistently outperforms the text-only baseline across all inference settings.}
\label{tab:ablation_input}
\begin{tabular}{lccc}
\toprule
\multirow{2}{*}{{Training Setting}} & \multicolumn{3}{c}{{Test Condition}} \\
\cmidrule(lr){2-4}
 & {(a) standard} & {(b) text only} & {(c) point cloud only} \\
\midrule
(I) Full Model & 3.53$_{(0.13)}$ & 3.22$_{(0.10)}$ & 3.35$_{(0.15)}$\\
%2.82$_{(0.10)}$ & \textbf{3.64}$_{(0.14)}$ \\
(II) Text-only & 2.82$_{(0.10)}$ & 3.21$_{(0.14)}$ & 2.81$_{(0.11)}$ \\
%&\textbf{3.22}$_{(0.10)}$ & 
%(c) Point-cloud-only & \textbf{3.35}$_{(0.15)}$ & 2.81$_{(0.11)}$ & \underline{2.98}$_{(0.11)}$ \\
%(d) Partial Point-cloud & \textbf{3.29}$_{(0.13)}$ & 2.95$_{(0.11)}$ & \underline{3.08}$_{(0.13)}$ \\
\bottomrule
\end{tabular}
\end{table}
%Moreover, Model (III), trained without text masking, over-relies on textual descriptions and suffers a significant performance drop under point-cloud-only inference. In contrast, the standard model (I), trained with random text masking, maintains robust performance across all settings, including partial point cloud corruption.

\textbf{Coarse-to-Fine Pose Generation}:
Next, we evaluate the effectiveness of the coarse-to-fine output strategy. We examine two aspects. First, the sensitivity of discrete tokens to grid resolution. Compared to the standard setting (B = 20 grids), we test: (III) B = 5 and (IV) B = 100. Second, we analyze different output representations: 
(V) using the texts of positions as outputs.
% removing special tokens and directly predicting poses in text form; 
(VI) only using pose anchoring tokens without pose residual tokens;
% removing $\langle\texttt{POS\_TOKEN}\rangle$, using only four discrete tokens for localization as described in \cref{equ: discrete tokens}; 
(VII) only using pose residual tokens without pose anchoring tokens.
% removing discrete tokens, using only a single $\langle\texttt{POS\_TOKEN}\rangle$ to regress asset poses.

\begin{table}[htbp]
\centering
\small

\centering
\caption{Ablation on grid resolution settings and output representations. The proposed coarse-to-fine strategy significantly outperforms other approaches.}
\label{tab:ablation_coarse_fine}
\setlength{\tabcolsep}{3.5pt}
\begin{tabular}{lc cc c ccc}
\toprule
\multirow{2}{*}{Metric} & \multirow{2}{*}{Standard} & \multicolumn{2}{c}{Grid Sensitivity} & & \multicolumn{3}{c}{Output Method} \\
\cmidrule(lr){3-4} \cmidrule(lr){6-8}
 & & (III)& (IV)& & (V) & (VI) & (VII) \\
\midrule
Collision Ratio & \textbf{0.86}$_{(0.11)}$ & 1.26$_{(0.24)}$ & \underline{0.94}$_{(0.15)}$ & & 1.00$_{(0.25)}$ & 1.15$_{(0.16)}$ & 19.19$_{(0.61)}$ \\
OOB Ratio & \textbf{1.16}$_{(0.28)}$ & 5.37$_{(0.91)}$ & 3.93$_{(1.94)}$ & & 9.25$_{(0.98)}$ & \underline{3.49}$_{(0.84)}$ & 8.24$_{(1.27)}$ \\
Floating Ratio & \textbf{3.36}$_{(0.78)}$ & 8.52$_{(2.42)}$ & 7.66$_{(2.25)}$ & & 7.14$_{(1.25)}$ & \underline{3.75}$_{(0.88)}$ & 13.10$_{(1.55)}$ \\
AI Physics Score & \textbf{4.24}$_{(0.13)}$ & 3.17$_{(0.21)}$ & 3.72$_{(0.18)}$ & & \underline{4.10}$_{(0.16)}$ & 4.06$_{(0.16)}$ & 2.76$_{(0.13)}$ \\
%\midrule
{Avg. AI Score} & \textbf{3.53}$_{(0.13)}$ & 2.58$_{(0.15)}$ & 2.76$_{(0.14)}$ & & 3.12$_{(0.11)}$ & \underline{3.30}$_{(0.13)}$ & 1.95$_{(0.08)}$ \\
\bottomrule
\end{tabular}
\end{table}

Results in \cref{tab:ablation_coarse_fine} report physical plausibility and AI-judge average scores. First, grid resolution directly affects the regression range of fine-grained tokens; overly large or small residual ranges destabilize decoding and reduce placement accuracy. Moreover, compared to purely textual coordinate output (Model V), NaLA’s coarse-to-fine strategy achieves consistent improvements in overall evaluation scores. Furthermore, when the pose anchoring tokens are removed, and only pose residual tokens are used (Model VII), all performance metrics degrade significantly. This result validates our earlier analysis in \cref{subsec: output}: without explicit anchoring signals, the model cannot effectively track previously placed assets, and the output stability deteriorates noticeably.

%We hypothesize that without discrete tokens, pose information is compressed into the high-dimensional hidden states of the $\langle\texttt{POS\_TOKEN}\rangle$, making previously placed asset poses indistinguishable to the LLM—since all placed assets share the same $\langle\texttt{POS\_TOKEN}\rangle$ in the generation sequence. Discrete tokens explicitly encode historical pose information, while the $\langle\texttt{POS\_TOKEN}\rangle$ enables fine-grained adjustment. This confirms the necessity of the coarse-to-fine output strategy.

\textbf{Training Strategy}:
Finally, we evaluate the effectiveness of our data augmentation and training strategies. We test the following variants: (VIII) removing input asset sequence shuffling; (IX) removing asset replacement; (X) removing scene rotation; (XI) training only on the 3D-FRONT dataset; (XII) jointly training on 3D-FRONT and Imaginarium. \cref{tab:ablation_training} validates the effectiveness of all augmentation strategies. Moreover, the two-stage approach (Model I) outperforms joint mixed training (Model XII). Model (I) first learns stable, global layout patterns from the large-scale synthetic 3D-FRONT, and then fine-tunes on the diverse, high-quality Imaginarium data to master local details. This sequential transfer proves more effective than optimizing for both distributions simultaneously, which can lead to delayed convergence and inferior performance.

It is also worth noting that NaLA is a model-agnostic framework. Replacing the underlying foundation model still yields strong performance. We provide additional discussions in the appendix.

\begin{table}[htbp]
\centering
\small
\caption{Ablation study on training strategies. Both the proposed augmentation techniques and the two-stage transfer learning strategy contribute to optimal performance.}
\label{tab:ablation_training}
\setlength{\tabcolsep}{4pt}
\begin{tabular}{lc ccc c cc}
\toprule
\multirow{2}{*}{{Metric}} & \multirow{2}{*}{{(I) Standard}} & \multicolumn{3}{c}{{Augmentation}} & & \multicolumn{2}{c}{{Training}} \\
\cmidrule(lr){3-5} \cmidrule(lr){7-8}
 & & (VIII) & (IX) & (X) & & (XI) & (XII) \\
\midrule
Physics Score & \textbf{4.24}$_{(0.13)}$ & 3.22$_{(0.17)}$ & 3.50$_{(0.14)}$ & 3.72$_{(0.16)}$ & & \underline{4.17}$_{(0.09)}$ & 4.15$_{(0.16)}$ \\
Semantic Score & \textbf{3.32}$_{(0.16)}$ & 2.32$_{(0.11)}$ & 2.35$_{(0.15)}$ & 2.60$_{(0.14)}$ & & 2.49$_{(0.13)}$ & \underline{3.06}$_{(0.18)}$ \\
Aesthetic Score & \textbf{3.05}$_{(0.16)}$ & 1.85$_{(0.09)}$ & 1.95$_{(0.13)}$ & 2.20$_{(0.12)}$ & & 2.02$_{(0.08)}$ & \underline{2.72}$_{(0.17)}$ \\
%\midrule
{Avg. AI Score} & \textbf{3.53}$_{(0.13)}$ & 2.46$_{(0.11)}$ & 2.60$_{(0.12)}$ & 2.84$_{(0.12)}$ & & 2.90$_{(0.08)}$ & \underline{3.31}$_{(0.15)}$ \\
\bottomrule
\end{tabular}
\end{table}

%% file: main_text/conclusion.tex
\section{Conclusion and Limitation}

We propose NaLA, which injects both asset and scene point clouds into an LLM and introduces a coarse-to-fine output framework to improve placement precision while significantly accelerating inference. Through two-stage training on layout datasets, NaLA demonstrates stronger capability in perceiving asset geometry and planning object placement compared to previous layout agents. Extensive experiments validate its superiority over baseline models and confirm the effectiveness of each key design component.

Currently, NaLA is limited by the base model capability. In addition, due to the scarcity of high-quality layout datasets, the training scale of NaLA remains limited. In the future, as new datasets become available, we will further improve NaLA’s placement performance on more complex tasks.

%% file: main_text/acknowledgement.tex
\section*{Acknowledgment}
This work was completed while Cheng Wan was an intern at Tencent IEG.
Yuan Liu was supported by the HKUST ``30 for 30'' Talent Acquisition Campaign, with funding provided by Kerry Group Limited.
This work was supported in part by the Shenzhen Hetao Institute (SLAI) Focus Project ``Spatial Intelligence and Its Applications in Robot Simulation'' (FPF10120260001); 
and in part by the ``1+1+1'' Joint Funding Program for the project ``LAYRA-AGE: Behavior-Informed Generative AI System for Elderly-Centered Adaptive Living Environments''.

%% file: appendix/appendix.tex
\section{Supplementary Experiments}

\subsection{Additional Quantitative Results}
\begin{figure*}[htbp]
\centering
\captionsetup[subfigure]{justification=centering}
% Row 1 (Scene type 1)
\begin{subfigure}[t]{0.24\textwidth}
  \centering
  \includegraphics[width=\linewidth,trim=2cm 5cm 2cm 5cm,clip]{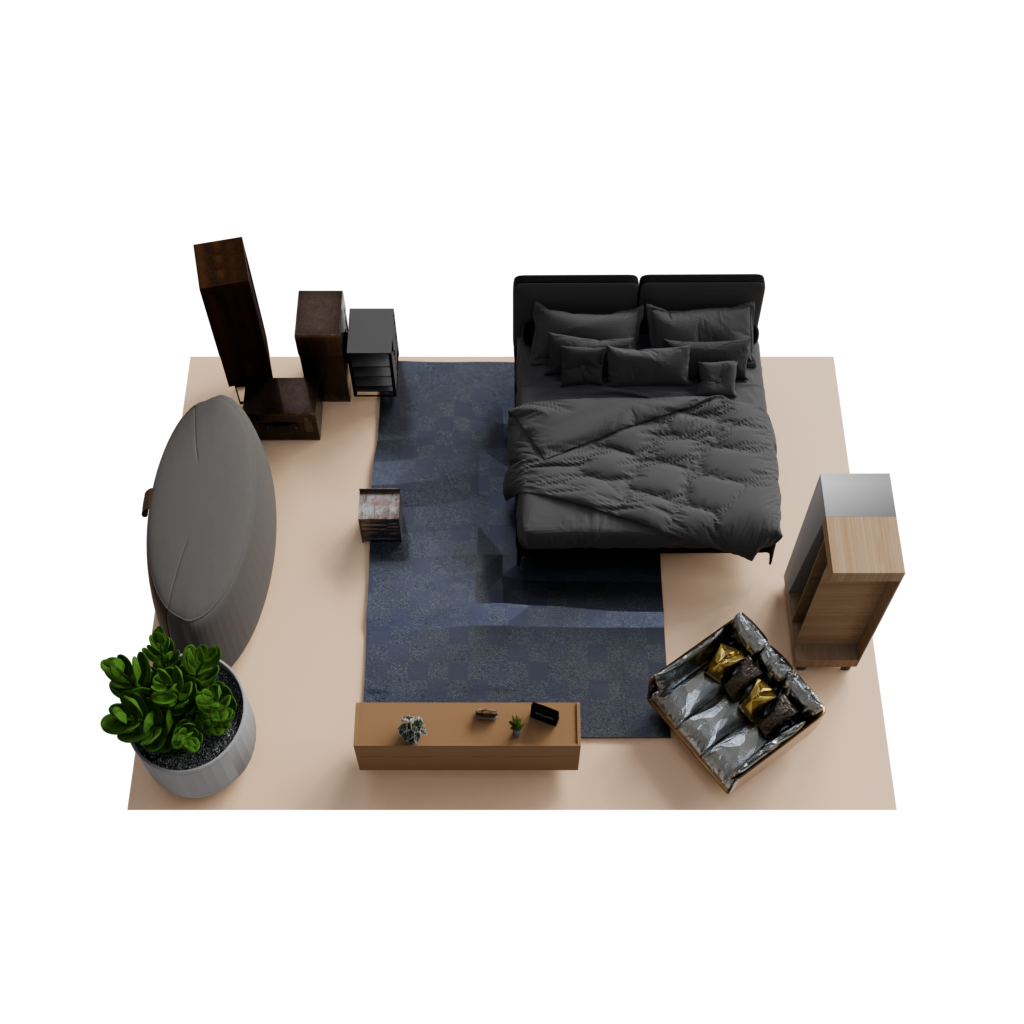}
\end{subfigure}\hfill
\begin{subfigure}[t]{0.24\textwidth}
  \centering
  \includegraphics[width=\linewidth,trim=2cm 5cm 2cm 5cm,clip]{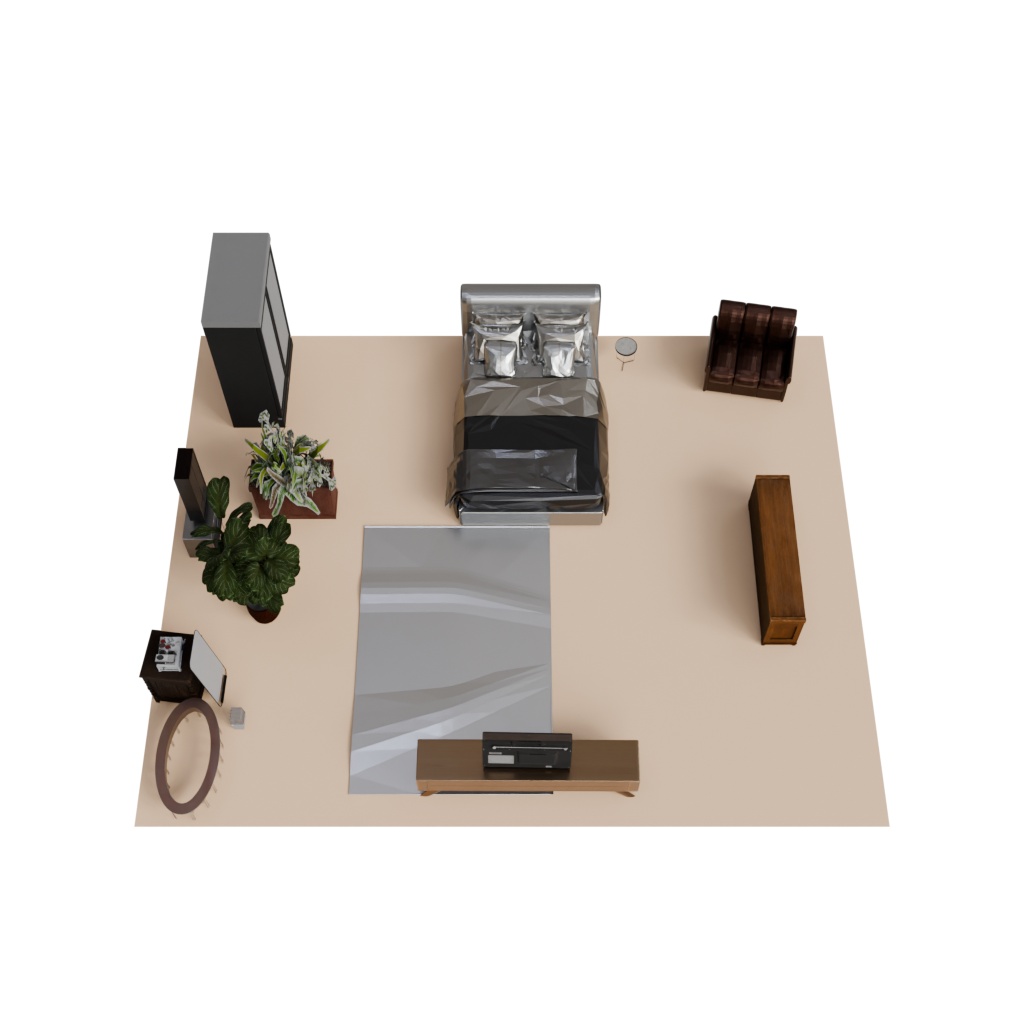}
\end{subfigure}\hfill
\begin{subfigure}[t]{0.24\textwidth}
  \centering
  \includegraphics[width=\linewidth,trim=2cm 4cm 2cm 5cm,clip]{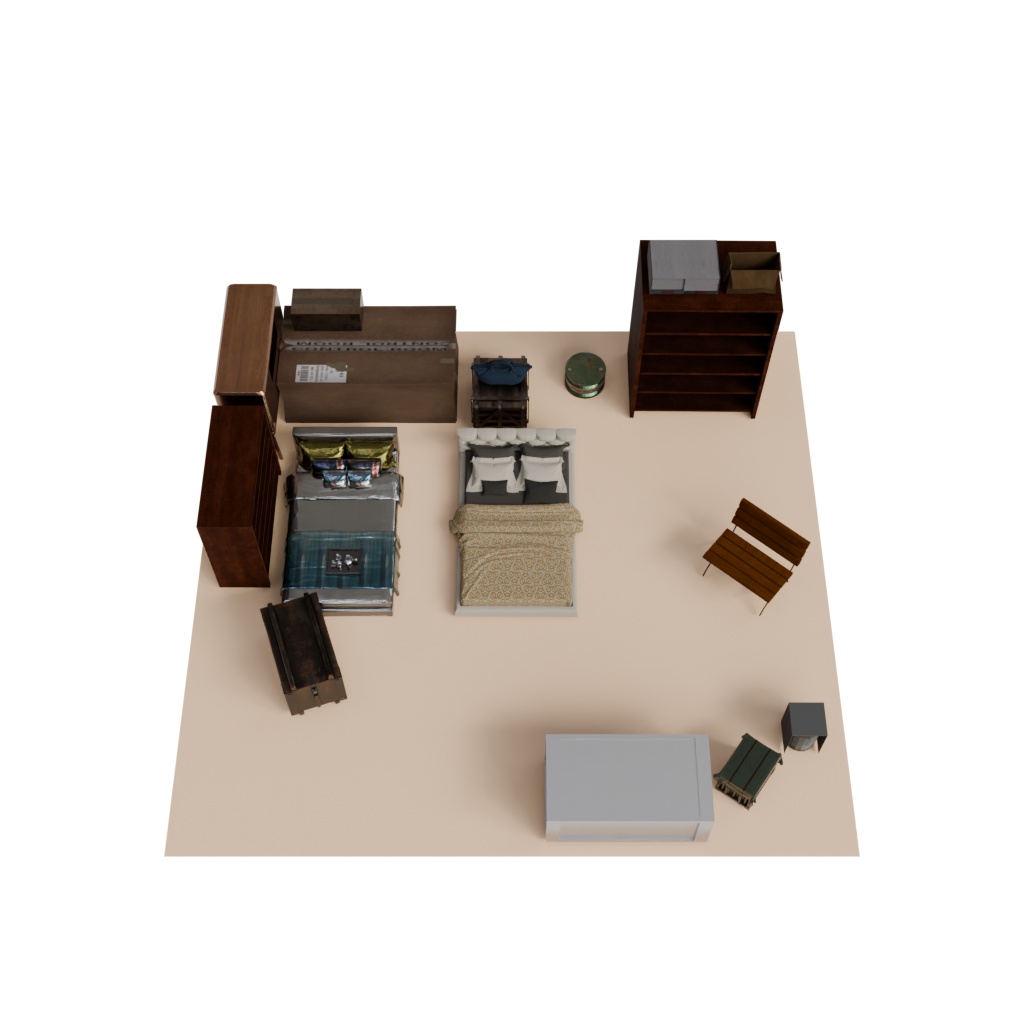}
\end{subfigure}\hfill
\begin{subfigure}[t]{0.24\textwidth}
  \centering
  \includegraphics[width=\linewidth,trim=0cm 7cm 0cm 7cm,clip]{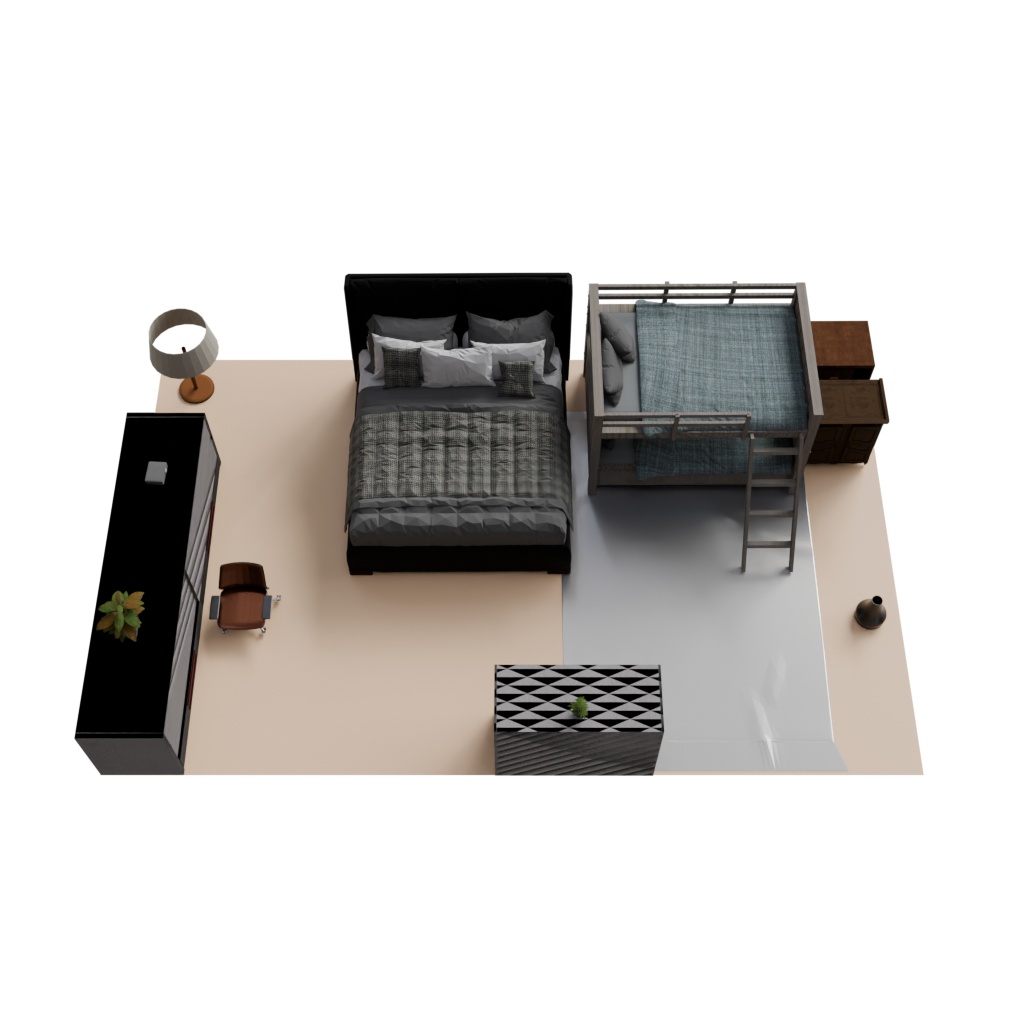}
\end{subfigure}

\centering
\captionsetup[subfigure]{justification=centering}
% Row 2 (Scene type 1)
\begin{subfigure}[t]{0.24\textwidth}
  \centering
  \includegraphics[width=\linewidth,trim=2cm 5cm 2cm 5cm,clip]{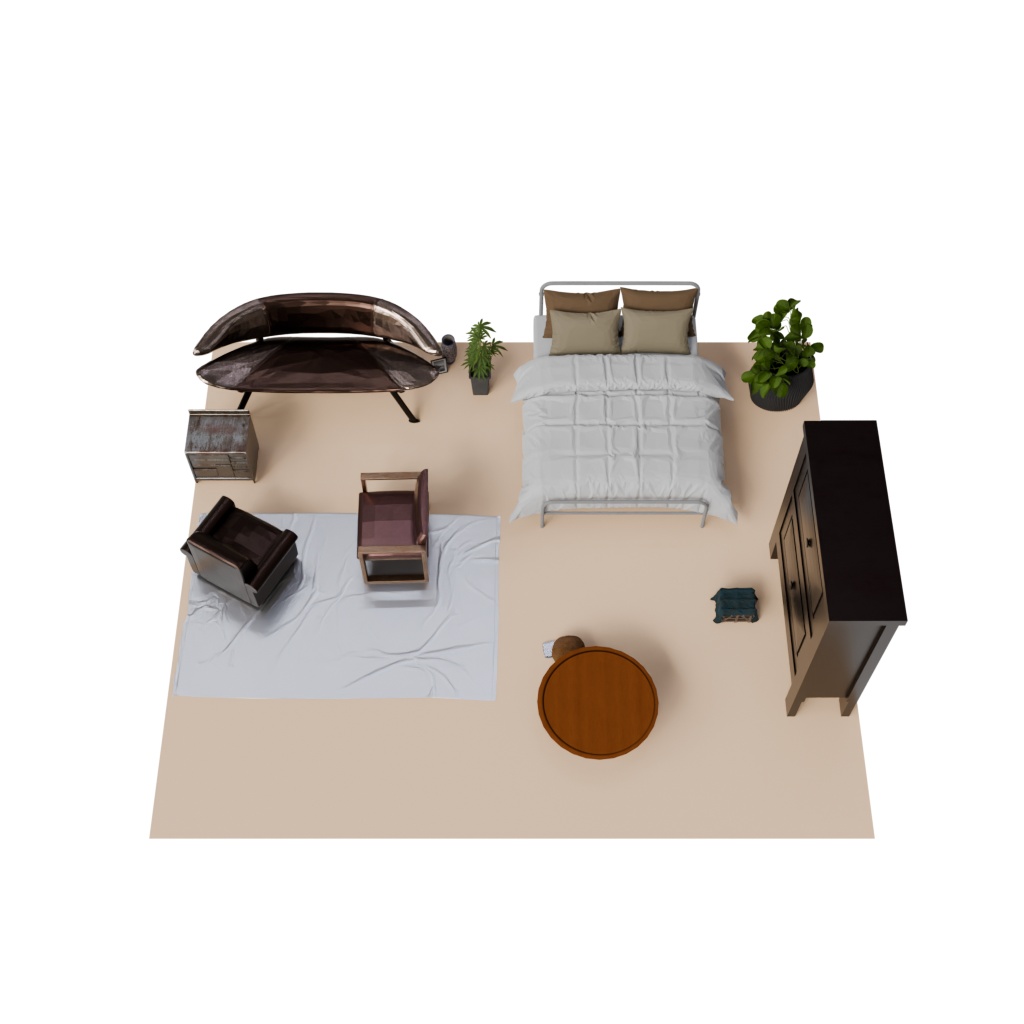}
\end{subfigure}\hfill
\begin{subfigure}[t]{0.24\textwidth}
  \centering
  \includegraphics[width=\linewidth,trim=2cm 5.7cm 2cm 5cm,clip]{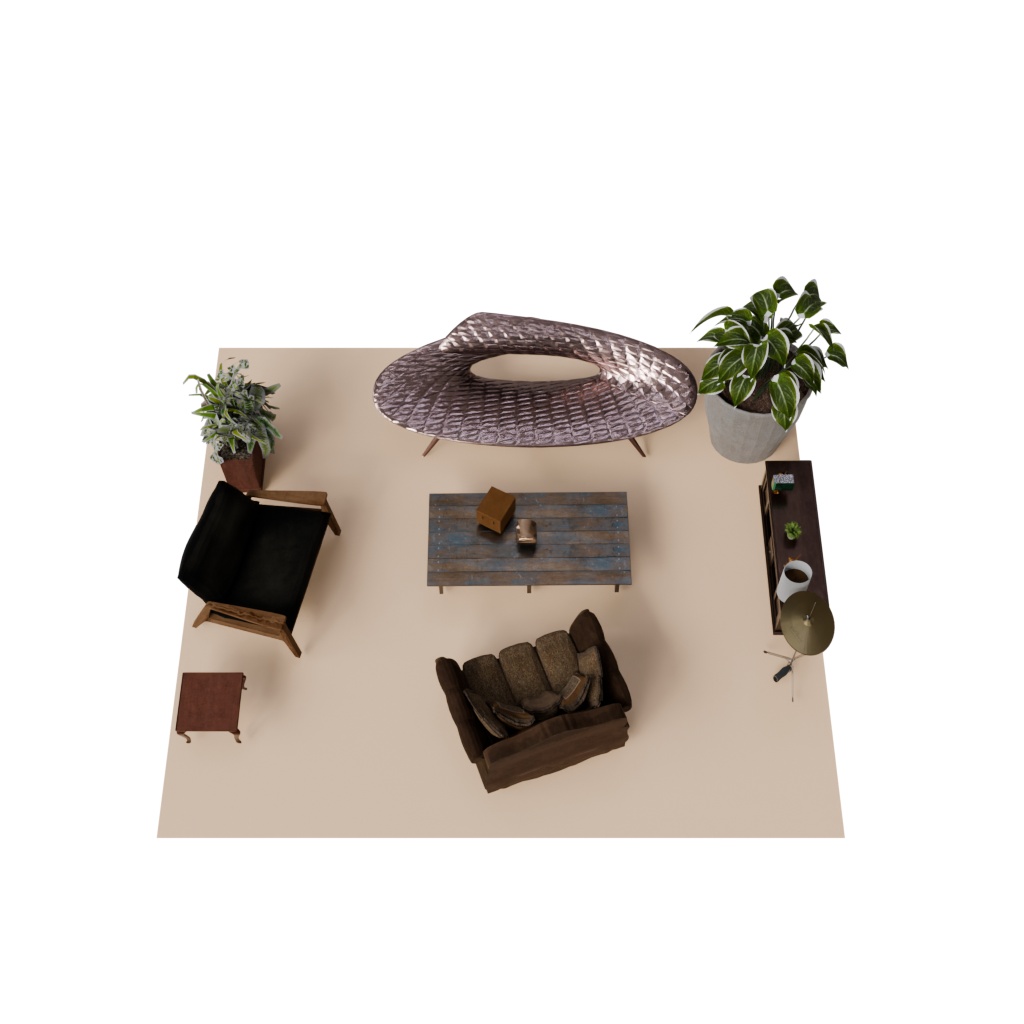}
\end{subfigure}\hfill
\begin{subfigure}[t]{0.24\textwidth}
  \centering
  \includegraphics[width=\linewidth,trim=2cm 5.8cm 2cm 5cm,clip]{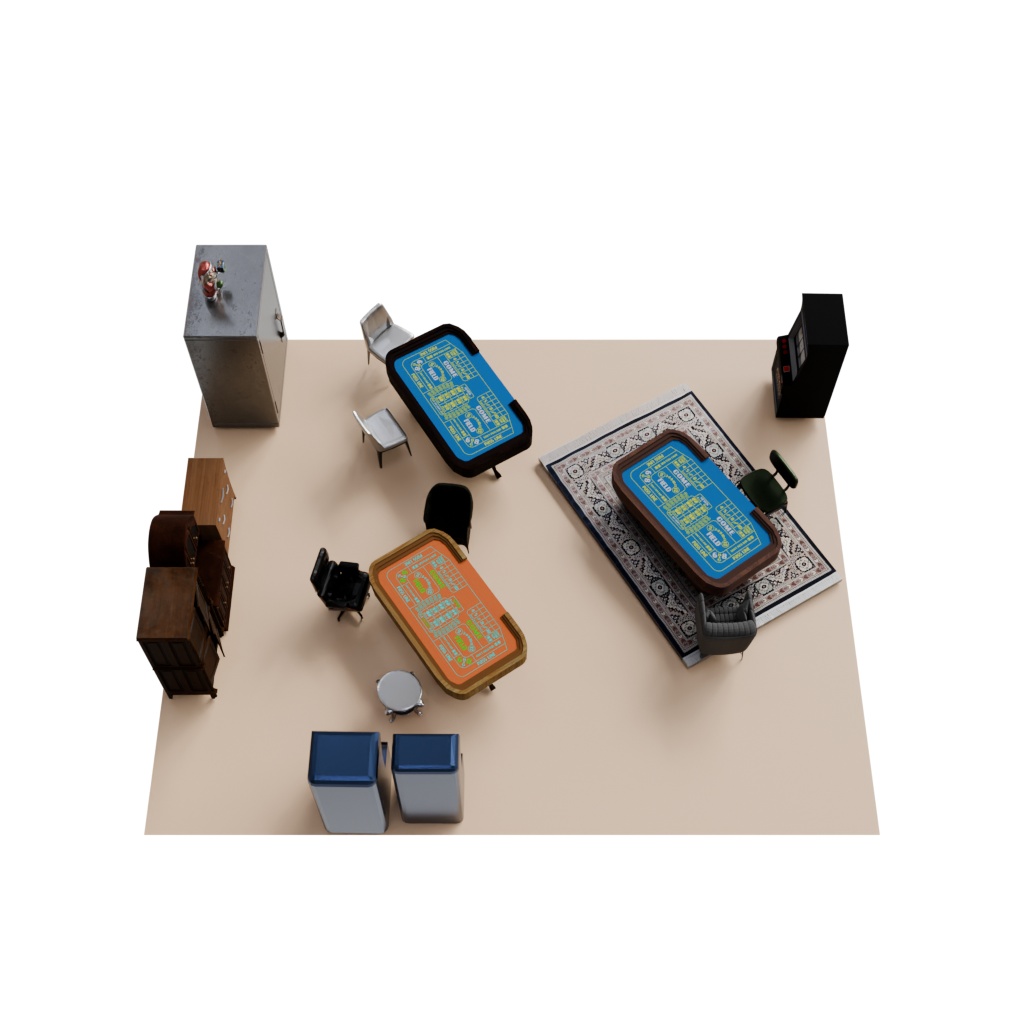}
\end{subfigure}\hfill
\begin{subfigure}[t]{0.24\textwidth}
  \centering
  \includegraphics[width=\linewidth,trim=2cm 5cm 2cm 5cm,clip]{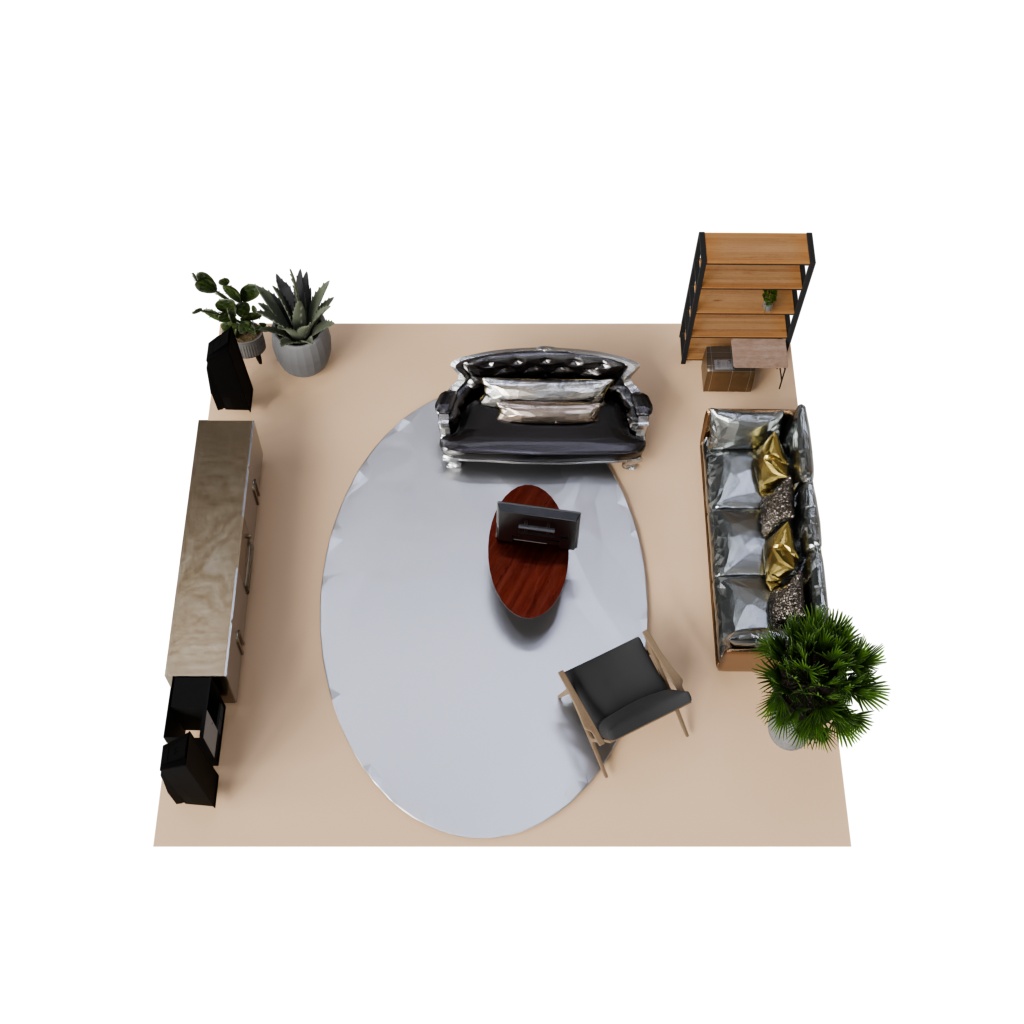}
\end{subfigure}

%row 3
\centering
\captionsetup[subfigure]{justification=centering}
% Row 1 (Scene type 1)
\begin{subfigure}[t]{0.24\textwidth}
  \centering
  \includegraphics[width=\linewidth,trim=2cm 5cm 2cm 5cm,clip]{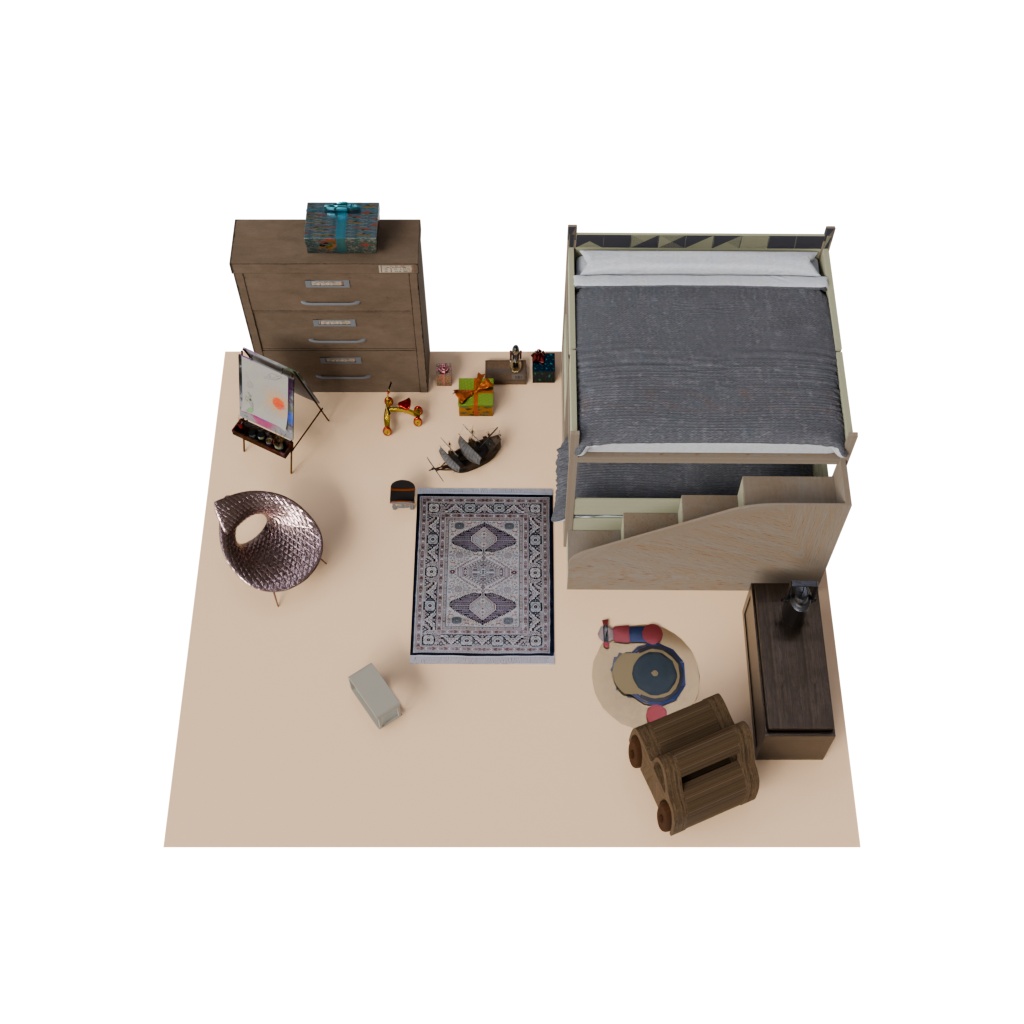}
\end{subfigure}\hfill
\begin{subfigure}[t]{0.24\textwidth}
  \centering
  \includegraphics[width=\linewidth,trim=2cm 5.7cm 2cm 5cm,clip]{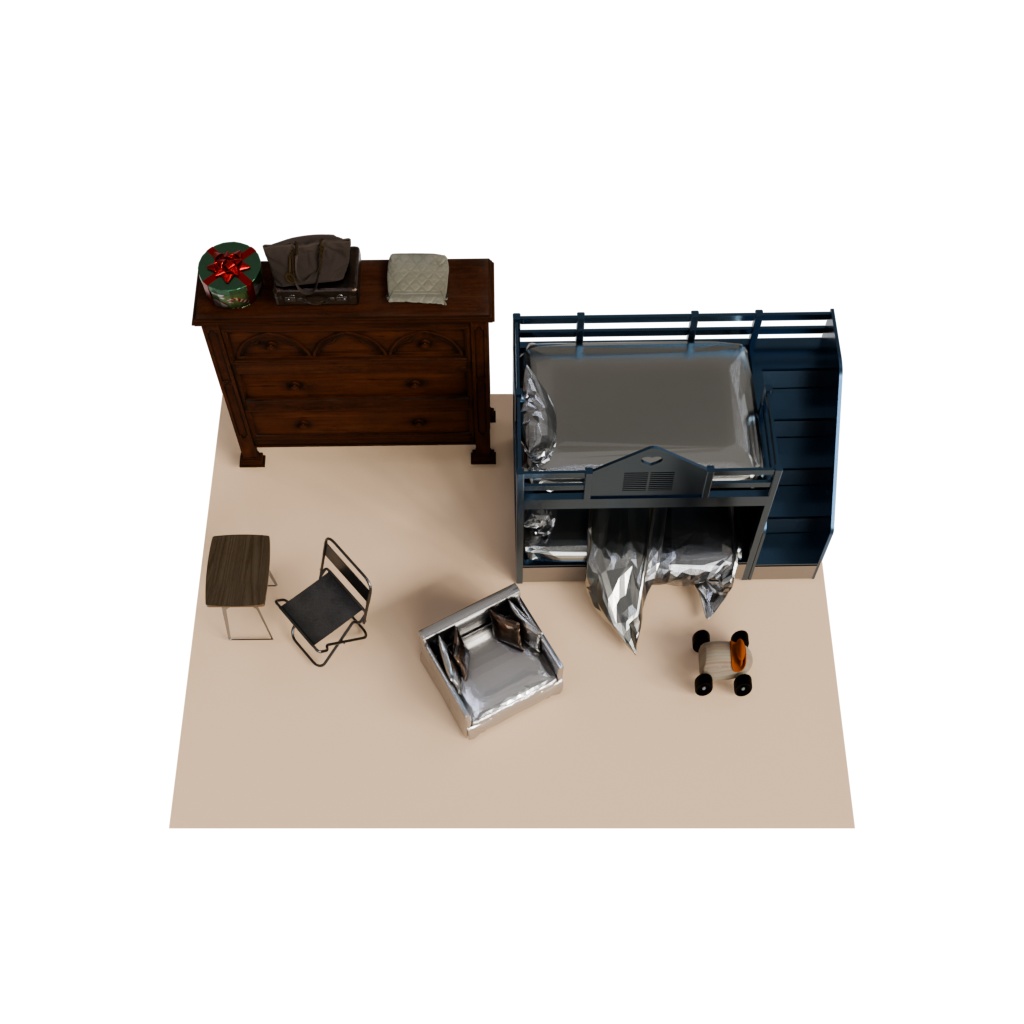}
\end{subfigure}\hfill
\begin{subfigure}[t]{0.24\textwidth}
  \centering
  \includegraphics[width=\linewidth,trim=2cm 5.7cm 2cm 5cm,clip]{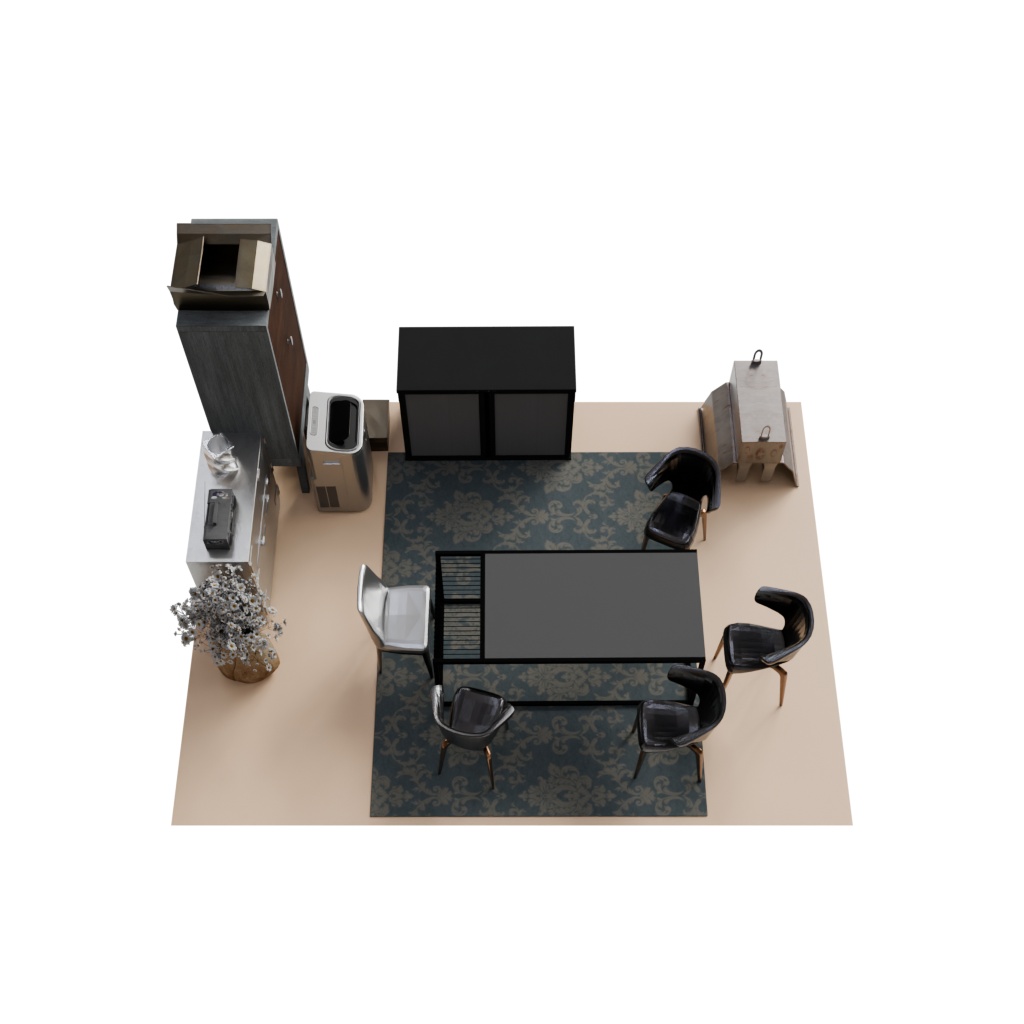}
\end{subfigure}\hfill
\begin{subfigure}[t]{0.24\textwidth}
  \centering
  \includegraphics[width=\linewidth,trim=2cm 4.5cm 2cm 5cm,clip]{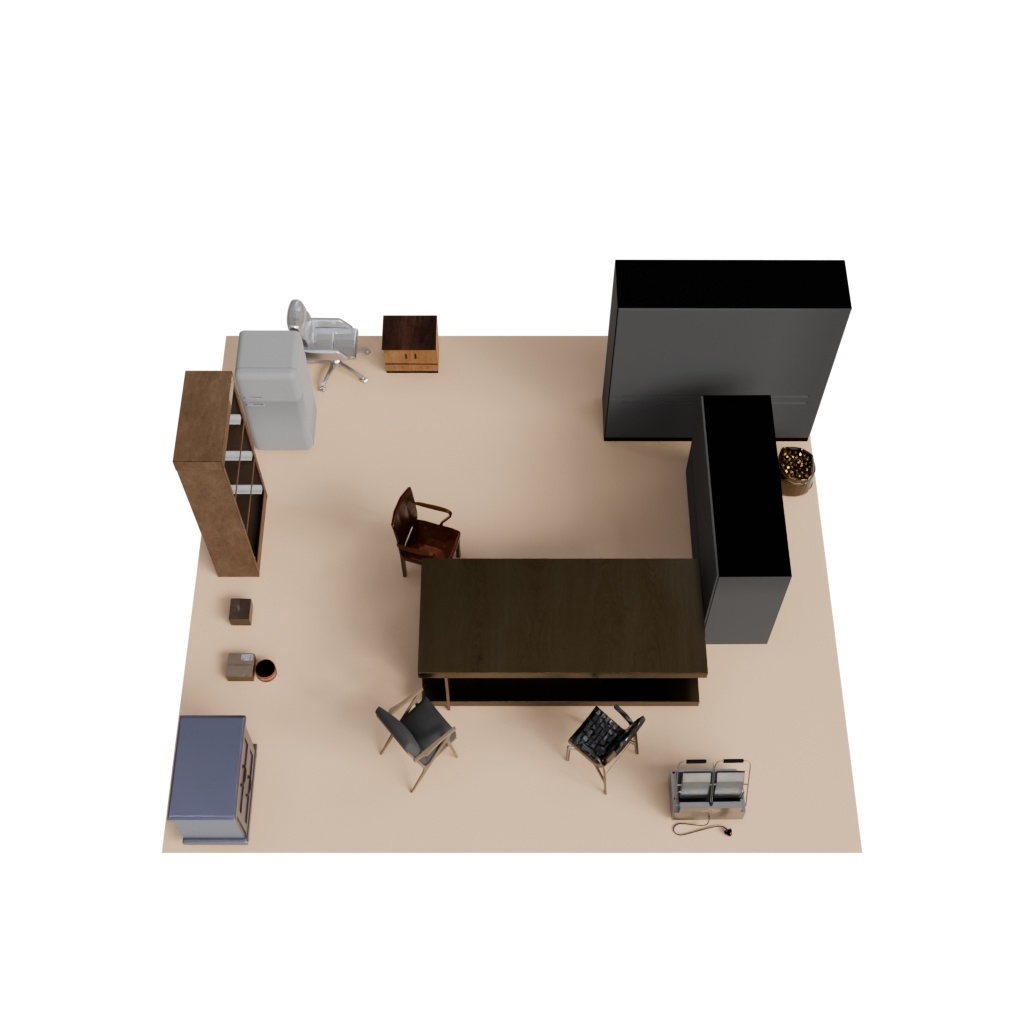}
\end{subfigure}

%row 4
\centering
\captionsetup[subfigure]{justification=centering}
% Row 1 (Scene type 1)
\begin{subfigure}[t]{0.24\textwidth}
  \centering
  \includegraphics[width=\linewidth,trim=2cm 5cm 2cm 5cm,clip]{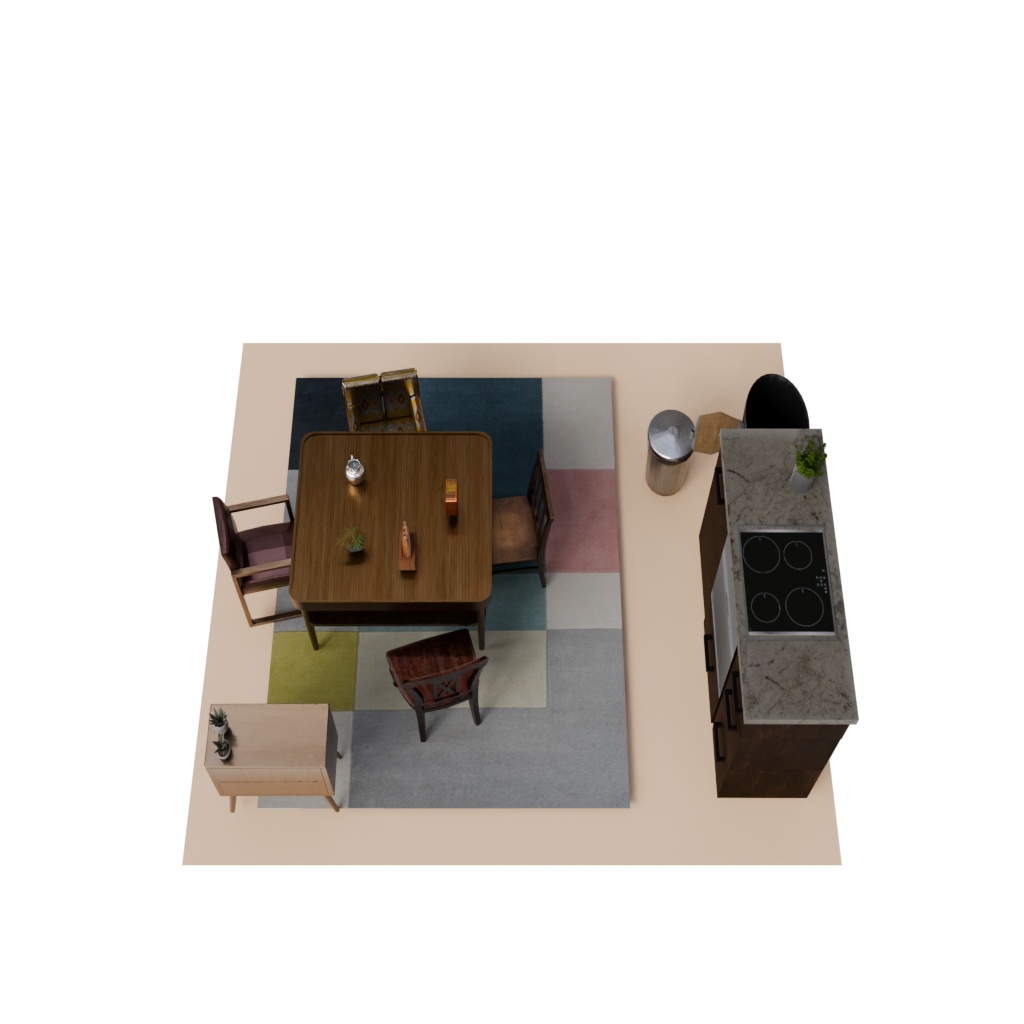}
\end{subfigure}\hfill
\begin{subfigure}[t]{0.24\textwidth}
  \centering
  \includegraphics[width=\linewidth,trim=2cm 5.7cm 2cm 5cm,clip]{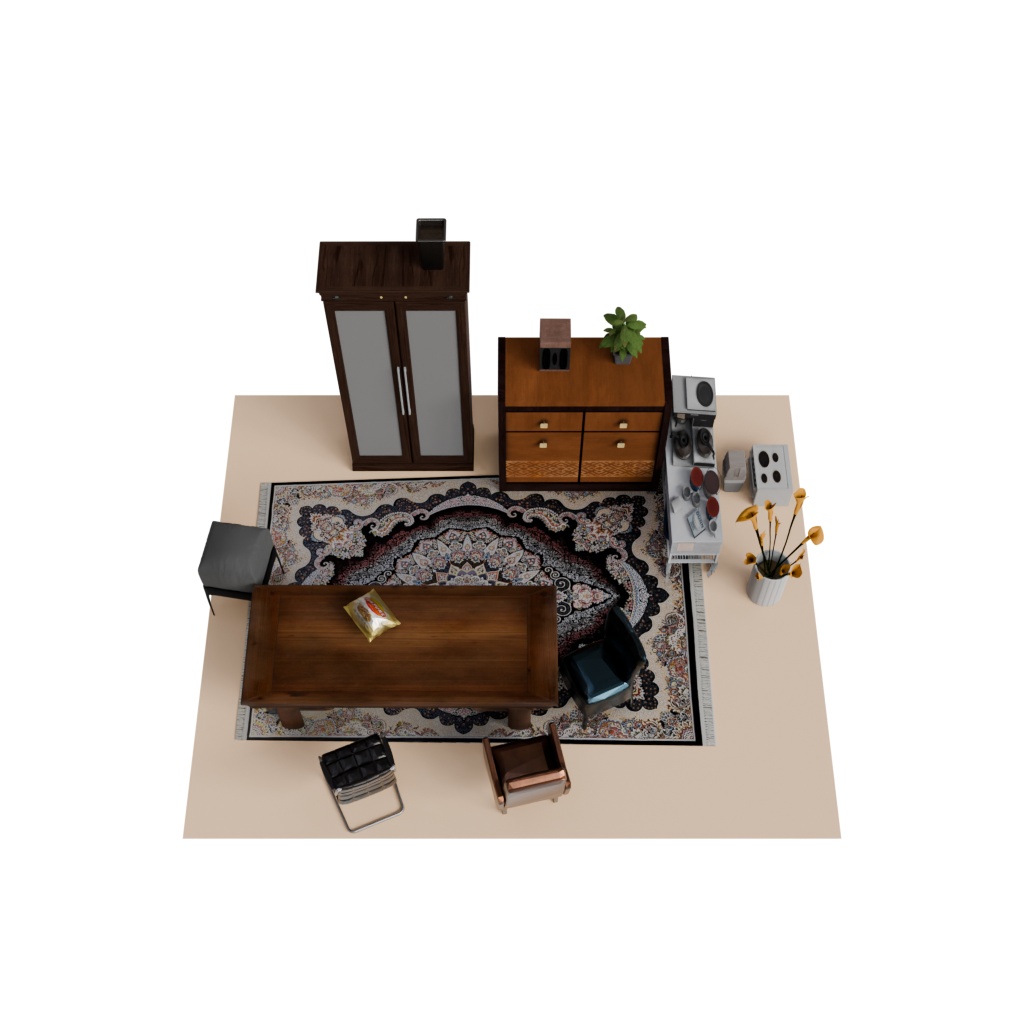}
\end{subfigure}\hfill
\begin{subfigure}[t]{0.24\textwidth}
  \centering
  \includegraphics[width=\linewidth,trim=2cm 5.7cm 2cm 5cm,clip]{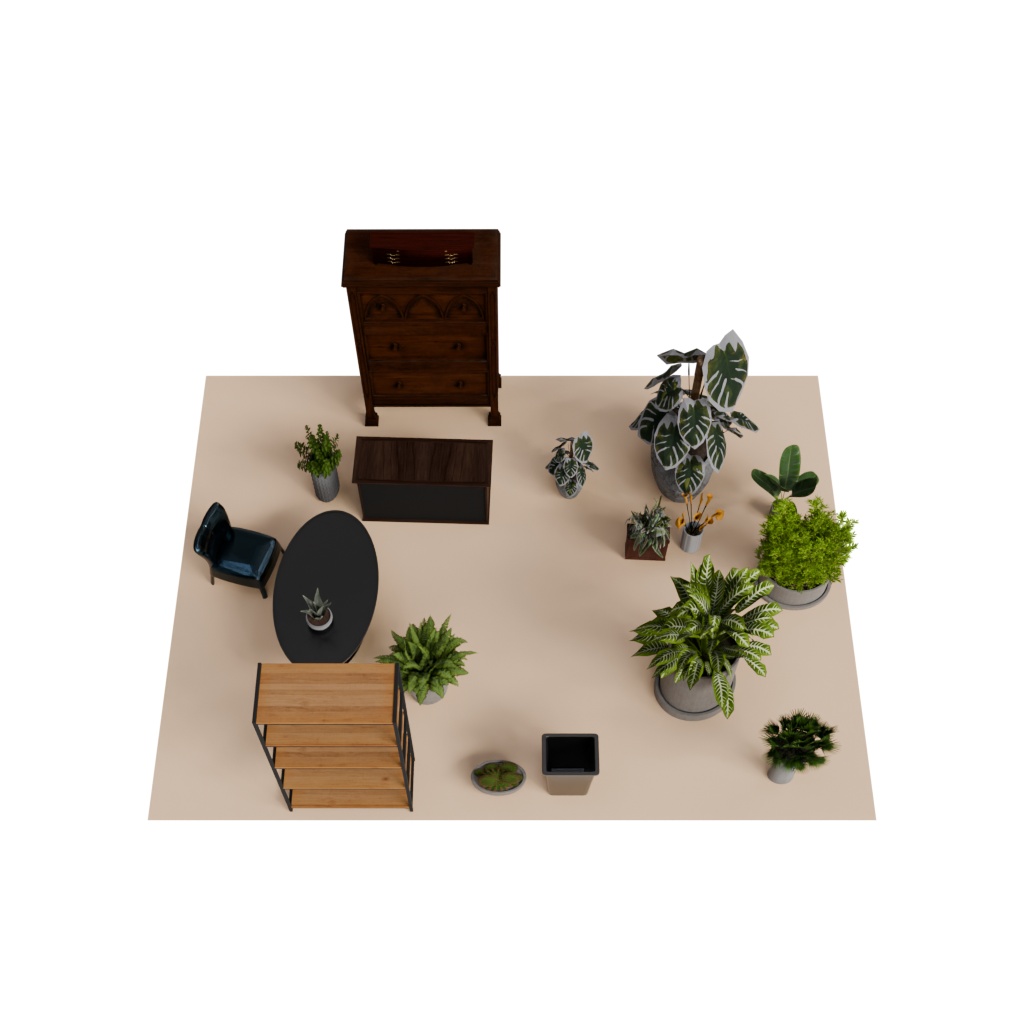}
\end{subfigure}\hfill
\begin{subfigure}[t]{0.24\textwidth}
  \centering
  \includegraphics[width=\linewidth,trim=2cm 4.5cm 2cm 5cm,clip]{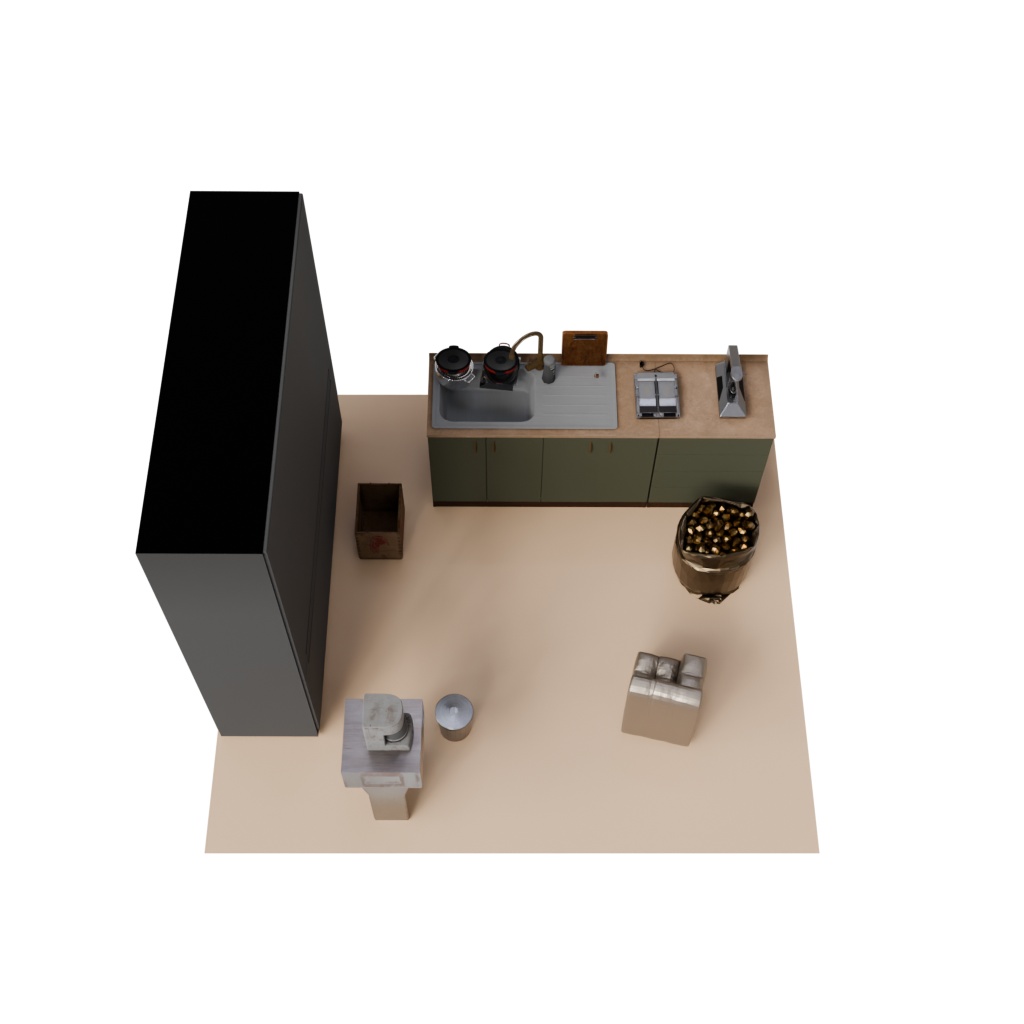}
\end{subfigure}

\caption{Qualitative results of NaLA's layout generation.}
\label{fig:more results}
\end{figure*}

We present additional placement results for NaLA in \cref{fig:more results}, including bedrooms, living rooms, kitchens, casinos, game rooms, and children's rooms. The model maintains strong placement performance across these scenarios, thereby validating our design choices: introducing 3D input tokens, output anchoring tokens, and output regression tokens, all of which contribute to NaLA's effectiveness.

\subsection{Additional Qualitative Results}

\textbf{Comparison with DirectLayout}:
We compare NaLA with DirectLayout (Ran et al.), a recent layout agent utilizing Chain of Thought and iterative refinement to improve layout quality.
The results are presented in \cref{tab:DirectLayout}.
Despite adopting a simpler end-to-end generation paradigm, NaLA demonstrates performance comparable to DirectLayout, further demonstrating its potential.

\begin{table}[h]
\centering
\small
\caption{Comparison with DirectLayout.}
\label{tab:DirectLayout}
\begin{tabular}{ccccccc}
\toprule
Methods & Collision & OOB & Float & Physics & Semantic & Aesthetic \\
\midrule
DirectLayout & \textbf{0.69} & 1.64 & 8.42 & \textbf{4.53} & \textbf{3.37} & 2.86 \\
NaLA & 0.86 & \textbf{1.16} & \textbf{3.36} & 4.24 & 3.32 & \textbf{3.05}  \\
\bottomrule
\end{tabular}
\end{table}

\textbf{More Ablation Experiments}: NaLA does not rely on a specific LLM backbone. We replace Qwen2.5-7B with a smaller Qwen3-4B and retrain the model.
Despite using a smaller backbone, the model still demonstrates acceptable performance,
%, which still achieves reasonable performance, 
as shown in \cref{tab:backbone}.
% , indicating that our approach is not tied to a fixed architecture.

\begin{table}[h]
\centering
\small
\caption{NaLA with different backbones.}
\label{tab:backbone}
\begin{tabular}{ccccccc}
\toprule
Backbones & Collision & OOB & Float & Physics & Semantic & Aesthetic \\
\midrule
Qwen3-4B & 1.10 & 4.85 & 8.12 & 4.13 & 3.09 & 2.51 \\
Qwen2.5-7B & 0.86 & 1.16 & 3.36 & 4.24 & 3.32 & 3.05  \\
\bottomrule
\end{tabular}
\end{table}

\section{Discrete Pose Tokenization and Reconstruction}

In this section, we describe how pose anchoring tokens are constructed and how final continuous poses are reconstructed.

\textbf{Scene Discretization.}
Given a scene with spatial extent
\[
[-\tfrac{X}{2}, \tfrac{X}{2}] \times
[-\tfrac{Y}{2}, \tfrac{Y}{2}] \times
[-\tfrac{Z}{2}, \tfrac{Z}{2}],
\]
we uniformly divide each spatial dimension into $B$ bins.

For an asset with ground-truth center position $(x,y,z)$ and orientation $\phi \in [0,2\pi)$, the coarse bin indices $(j,k,p)$ are computed as:

\begin{equation}
\begin{aligned}
j = \left\lfloor \frac{x + \frac{X}{2}}{X} B \right\rfloor, 
k = \left\lfloor \frac{y + \frac{Y}{2}}{Y} B \right\rfloor, 
p = \left\lfloor \frac{z + \frac{Z}{2}}{Z} B \right\rfloor .
\end{aligned}
\end{equation}

\textbf{Orientation Discretization.}
Object orientation is discretized into four canonical directions aligned with the positive and negative $x$ and $z$ axes. The orientation bin index $q$ is defined as:

\begin{equation}
q =
\begin{cases}
1, & \phi \in [\tfrac{7\pi}{4}, 2\pi) \cup [0, \tfrac{\pi}{4}) \\
2, & \phi \in [\tfrac{\pi}{4}, \tfrac{3\pi}{4}) \\
3, & \phi \in [\tfrac{3\pi}{4}, \tfrac{5\pi}{4}) \\
4, & \phi \in [\tfrac{5\pi}{4}, \tfrac{7\pi}{4})
\end{cases}
\end{equation}

\textbf{Coarse Pose Reconstruction.}
The spatial centers of the selected bins are defined as:

\begin{equation}
\begin{aligned}
x_j^{*} = -\frac{X}{2} + \left(j + \frac{1}{2}\right)\frac{X}{B},\ \  y_k^{*} =& -\frac{Y}{2} + \left(k + \frac{1}{2}\right)\frac{Y}{B}, \\
z_p^{*} = -\frac{Z}{2} + \left(p + \frac{1}{2}\right)\frac{Z}{B}&,\ \  \phi_q^{*} = \frac{\pi}{2} (q-1) .
\end{aligned}
\end{equation}

The coarse-grained pose is therefore:

\begin{equation}
\widehat{\boldsymbol P}_i^{*} = (x_j^{*}, y_k^{*}, z_p^{*}), \qquad
\widehat{O}_i^{*} = \phi_q^{*}.
\end{equation}

\textbf{Fine-Grained Pose Refinement.}
After predicting pose anchoring tokens, the model generates a $\langle\texttt{POS\_TOKEN}\rangle$ to regress continuous residuals:

\[
\Delta \widehat{\boldsymbol P}_i,\quad
\Delta \widehat{O}_i,\quad
\widehat{\boldsymbol S}_i.
\]

The final predicted asset pose is:

\begin{equation}
(\widehat{\boldsymbol P}_i^{*} + \Delta \widehat{\boldsymbol P}_i,\;
\widehat{\boldsymbol S}_i,\;
\widehat{O}_i^{*} + \Delta \widehat{O}_i).
\end{equation}

\section{Model Architecture and Training}
We next introduce the detailed architecture of NaLA used in our experiments.

\subsection{Encoder}
In the encoder, NaLA adopts the pretrained weights of SPFormer and PointBERT and keeps them frozen. 
For asset point clouds, PointBERT first extracts 512 patches. The extracted features are then projected to a 128-dimensional space through an MLP. Meanwhile, the 3D coordinates of the center point of each patch are also projected to 128 dimensions via another MLP. These coordinate embeddings are concatenated with the PointBERT features to preserve the absolute geometric coordinates within the representation. The resulting 512 features of dimension 256 are then processed by a two-layer Q-Former to extract 16 query tokens, which are subsequently projected to match the LLM hidden dimension and injected into the LLM input.

For scenes, we adopt a similar encoding strategy. We first sample 4096 points from the empty scene, which are then processed by SPFormer to obtain 2048 superpoint features with a feature dimension of 32. To preserve absolute scene coordinates and scene scale information, the superpoint coordinates and scene-level features are projected to 32 dimensions through MLPs and concatenated with the SPFormer features, resulting in 2048 features of dimension 96. These features are then processed by an independent Q-Former to extract 64 query tokens.

\subsection{Decoder}
During the output stage, the decoder used for the pose residual token operates as follows. We extract the hidden states from the last $L=4$ layers of the LLM backbone. These four feature representations are first projected to a lower dimension and then processed by a lightweight two-layer Transformer. The final hidden state after the Transformer is passed through an MLP to decode the asset pose vector. Notably, instead of directly mapping the last hidden state through an MLP, our decoder aggregates high-level representations from multiple upper layers of the LLM that are closely related to high-level semantic information. This design helps capture richer semantic cues from the LLM for pose prediction.

\subsection{Model Training}
During training, we split the datasets into training and test sets. The asset libraries of the 3D-FRONT and Imaginarium datasets contain approximately 17,000 assets across about 300 categories. For each category, assets are split into training and test sets with an 80\%–20\% ratio. Assets in the training split are used for asset-replacement data augmentation, while assets in the test split are used to construct evaluation scenes.

For each asset in the database, we use Gemini-3-Flash-preview to generate a one-sentence description based on its rendered image, which serves as the textual label of the asset during training. During preprocessing, all assets are rotated along the $x$ and $z$ axes in advance so that the model only needs to learn the rotation around the $y$ axis.

During training, the LLM backbone and the point cloud encoders remain frozen. Only the LoRA modules, the pose decoder, and encoder-related adapters and Q-Former modules are trainable. We register 50 ID tokens, 20 pose-bin tokens, and 4 rotation-bin tokens in the LLM vocabulary. For the input asset and scene features, we use special tokens such as \texttt{<item\_start>}, \texttt{<item\_end>}, \texttt{<scene\_start>}, and \texttt{<scene\_end>} to indicate the positions of the 3D features in the prefix. Additionally, in the loss term, we set hyperparameters $\lambda_1=\lambda_2=\lambda_3=1$.

The first-stage training is conducted on 3D-FRONT with a batch size of 24 and a learning rate of $3\times10^{-4}$ for 200 epochs. The second-stage training is performed on the Imaginarium dataset, while mixing in 50\% of 3D-FRONT data to prevent catastrophic forgetting. In this stage, the batch size is set to 16 and the learning rate to $1\times10^{-4}$ for 50 epochs. In both stages, weight decay is set to 0.01, and a cosine learning rate schedule is used. We adopt the AdamW optimizer and train the model on 8 NVIDIA H20 GPUs.

\section{Evaluation}

\subsection{Test Case Construction}
We next describe the testing procedure and experimental details. During evaluation, we use the following prompt and employ Gemini-3-Flash-Preview to select 20 objects for each scene. Specifically, we provide the AI judge with the asset categories from the test splits of the 3D-FRONT and Imaginarium asset libraries and inform it of the room type. The prompt is as follows:

\begin{lstlisting}
"""You are a professional interior designer helper. You output strictly valid JSON lists. I want to design a {style_description} {target_scene_type}. Please select 20 asset categories from the allowed list below to place in this room. You can select duplicate categories if reasonable (e.g., multiple "Dining_chair" or "Plate").

The allowed category list is: {all_categories}

Output Requirement:
1. Select items that naturally belong in a {target_scene_type}.
2. The total count must be 20.
3. Output strictly a JSON list of strings [ "str1", "str2" ... ].

Your JSON List:
"""
\end{lstlisting}

Here, \textit{\{target\_scene\_type\}} specifies the room type, and \textit{\{style\_description\}} is filled with a randomly sampled room style (out of ``minimalist'', ``cozy and warm'', ``messy and cluttered'', ``vintage and retro'', ``industrial'', ``luxurious and expensive'', ``Modern and clean'') to increase the diversity among the three test cases generated for each room. After the AI judge selects the asset categories, we randomly sample one asset for each category from the test splits of the 3D-FRONT and Imaginarium datasets, forming the set of assets to be placed in the scene. For each asset, we use Gemini-3-Flash-Preview to generate a one-sentence description based on its rendered image, which serves as its textual feature.

Next, we randomly generate an empty rectangular room, where the height is randomly sampled from 2.8–4.5 meters, and the width and length are randomly sampled from 4–8 meters. We then double-check the validity of the generated test cases and regenerate any unreasonable combinations of assets and rooms. Finally, we feed the point cloud features, textual features, or image information of these rooms and assets into NaLA or the baseline models for layout generation and evaluation.

\subsection{Baseline Models}
For the baseline models (LayoutGPT, Holodeck, and LayoutVLM), we reproduce their results using the official code released in their repositories. It is worth noting that Holodeck is originally a complete room generation pipeline, which includes not only the object placement module but also modules for floor, wall, and ceiling generation, as well as asset retrieval. In our reproduction, we retain only the object placement module of Holodeck, allowing it to generate placement configurations for the given assets. Similarly, when evaluating generation efficiency, we disable the other modules of Holodeck to ensure a fair comparison.

\subsection{Evaluation Metrics}
We employ three standard physical metrics to evaluate the geometric quality of the generated layouts. The calculation details are as follows:

\paragraph{Collision Ratio.}
We evaluate collisions by measuring the volumetric intersection between objects. To account for object size differences (e.g., a collision between two large wardrobes is more severe than between two small cups), we use a volume-weighted formulation. We first voxelize all meshes with a fixed pitch to estimate their volumes and intersection ratios. For a scene with $N$ objects, let $V_i$ be the volume of object $i$, and $\text{IoU}(i, j)$ be the intersection-over-union of the voxelized representations of objects $i$ and $j$. The weighted collision score $\mathcal{S}_{\text{col}}$ is defined as:
\begin{equation}
    \mathcal{S}_{\text{col}} = \frac{\sum_{i<j} (V_i + V_j) \cdot \mathbb{I}(\text{collision}_{i,j})}{\sum_{i<j} (V_i + V_j)},
\end{equation}
where $\mathbb{I}(\text{collision}_{i,j})$ is the binary indicator of whether the voxelized intersection volume is non-zero. A lower score indicates fewer physical violations tailored to object significance.

\paragraph{Out-of-Bounds (OOB) Ratio.}
This metric measures the proportion of object volume placed outside the valid room boundary. Let $\mathcal{P}_{\text{room}}$ be the 2D polygon of the room floor and $\mathcal{P}_{i}$ be the projected 2D footprint of object $i$. The individual OOB ratio $r_i$ for object $i$ is calculated as the area of its footprint falling outside the room polygon relative to its total area:
\begin{equation}
r_i = \frac{\text{Area}(\mathcal{P}_{i} \setminus \mathcal{P}_{\text{room}})}{\text{Area}(\mathcal{P}_{i})}.
\end{equation}
The scene-level OOB score is the volume-weighted average of individual ratios: $\mathcal{S}_{\text{oob}} = (\sum_{i} V_i \cdot r_i) / (\sum_{i} V_i)$. %To ensure robustness against coordinate system shifts, we compute the OOB score using both the raw coordinates and a centralized room boundary, reporting the minimum of the two.

\paragraph{Floating Ratio.}
The floating ratio measures the percentage of objects that are not physically supported. 
Before computing the floating rate, we first filtered out all objects that are attached to walls or ceilings in the scene (for example, all assets with the category “chandelier” were removed). Afterward, an object $i$ is considered supported if either: (1) its bottom face is within a tolerance threshold of the floor level, or (2) it rests on top of another already supported object $j$. We construct a support graph where edges represent vertical contact (distance $<\tau$) and horizontal overlap. The floating ratio is the fraction of unsupported objects:
\begin{equation}
    \mathcal{S}_{\text{float}} = \frac{1}{N} \sum_{i=1}^{N} \mathbb{I}(i \text{ is unsupported}).
\end{equation}

\subsection{Qualitative Evaluation Protocol}
\label{appendix:qualitative_prompt}

For qualitative evaluation, we employ a specific system instruction and prompt template to guide both the AI judge (Gemini-3-Flash-Preview) and human evaluators. The prompt enforces a strict scoring rubric based on three dimensions: Physical Plausibility, Semantic Plausibility, and Visual Aesthetics. 
The system instruction is:
\begin{lstlisting}
"""You are an expert architect and interior design critic with a specialization in spatial planning evaluation. Your task is to strictly evaluate a generated 3D room layout based on provided rendered images (Top-down view + Perspective views).

You will evaluate the layout on 3 main dimensions:
1. Physical Plausibility
2. Semantic Plausibility
3. Visual Aesthetics

**SCORING METHODOLOGY (CRITICAL):**
For each main dimension, you must evaluate specific **Sub-criteria**.
1. Assign a score from 1 to 5 (Integer) for EACH sub-criterion.
2. Calculate the average of these sub-scores to get the Final Dimension Score.
3. Round the Final Dimension Score to the nearest integer for the JSON output.

Scoring Rubric:
- 1: Very Poor (Critical failure, completely unusable)
- 2: Poor (Major flaws, breaks immersion)
- 3: Fair (Acceptable logic, but unrefined)
- 4: Good (Functional and pleasing, minor issues)
- 5: Excellent (Professional quality, flawless)

Output strictly in JSON format."""
\end{lstlisting}

Additionally, the AI judge receives the rendered scene images, including a top-down view and four 45-degree oblique views from the north, south, east, and west directions, along with the room type. The corresponding user prompt is shown below. 
\begin{lstlisting}
"""Here are the rendered images of a generated room layout.
**Target Room Type:** "{room_type}"

Please evaluate the layout based on the following strict rubric. You must score every sub-criterion to derive the final score.

---

### Dimension 1: Physical Plausibility
**Does the layout obey basic physics?**
*(Sub-criteria)*
1. **Gravity & Support:** Are objects floating in the air? Are heavy objects naturally supported by the floor or other surfaces?
2. **Collision:** Do objects intersect or clip into each other or walls significantly? (Ignore very minor mesh overlaps).
3. **Stability:** Are objects placed in a way that implies they would fall over in real life?

---

### Dimension 2: Semantic Plausibility
**Is the layout practical and functional for human use?**
*(Sub-criteria)*
1. **Scene Identity (Layout-based):** Given the fixed set of assets, does this specific **arrangement** successfully convey the function of a "{room_type}"? (e.g., A bathroom layout should look like a bathroom, not a bedroom, based on how items are grouped).
2. **Accessibility / Flow:** Can a human physically walk through the space? Are pathways clear? Are doors, drawers, or critical zones blocked by other objects?
3. **Usability Logic:** **Definition:** The strict functional relationship between interacting objects.
    *   *Examples:* A sofa must face the TV; A toilet must have legroom; A desk chair must face the desk. Is the primary function of the furniture enabled by its orientation and position?
4. **Everyday Habits:** **Definition:** The "soft" constraints of human behavior and comfort, distinct from strict logic.
    *   *Examples:* Is the nightstand practically placed within reach of the bedhead? Is the coffee table at a comfortable reach distance from the sofa (not too far, not too close)? Does the layout feel "natural" to live in?

---

### Dimension 3: Visual Aesthetics
**Is the layout visually pleasing and well-composed?**
*(Sub-criteria)*
1. **Composition & Spatial Balance:**
    *   *Explanation:* Evaluate the distribution of "visual weight". Does the room feel lopsided? Is there an appropriate use of negative space (empty floor), or is it overcrowded/too sparse?
2. **Alignment & Grid Logic:**
    *   *Explanation:* Evaluate the geometric order. Are objects aligned to implied architectural lines (walls, rugs)? Are rotations clean or chaotically random without purpose? Do edges align pleasingly?
3. **Arrangement Harmony:**
    *   *Explanation:* Do the objects feel like they belong together in this specific cluster? Is the grouping aesthetically coherent, or does it look like a random pile of assets dumped on the floor?

---

### Output Format
Output ONLY valid JSON.
**Important:** Inside the "reasoning" text, you must explicitly list the scores for each sub-criterion (e.g., "Gravity: 5, Collision: 3... Average: 4").

{
  "physical_plausibility": {
    "reasoning": "Sub-scores: [Gravity: X, Collision: X, Stability: X]. Explanation...",
    "score": <int, Rounded Average>
  },
  "semantic_plausibility": {
    "reasoning": "Sub-scores: [Scene Identity: X, Accessibility: X, Usability: X, Habits: X]. Explanation...",
    "score": <int, Rounded Average>
  },
  "visual_aesthetics": {
    "reasoning": "Sub-scores: [Composition: X, Alignment: X, Harmony: X]. Explanation...",
    "score": <int, Rounded Average>
  }
}
"""
\end{lstlisting}

\subsection{Irregular Scene Test and Efficiency Test}
For the irregular scene and efficiency evaluations, we first design a set of irregular room layouts. The asset generation process remains consistent with the main experiments, while the irregular floor shapes are additionally generated using the AI judge (Gemini-3-Flash-Preview). Specifically, the AI outputs a list of vertices representing the floor polygon, where consecutive vertices form the boundary of an irregular floor plan. For example, the vertex list \([(0,0), (1,0), (1,0.5), (0.5,0.5), (0.5,1), (0,1)]\) forms an L-shaped room. We use the following prompt to generate the irregular scene:
\begin{lstlisting}
    Act as a 2D floor plan layout generator. Please generate exactly 2 distinct coordinate lists for irregular, orthogonal rooms (one with 1 cutout, the other with 2 cutouts).

Strict Rules:
1. Orthogonal Only: All corners must be right angles. Adjacent points in the list must share either the exact same X or the same Y coordinate. No diagonal lines.
2. Form: Treat the room as a main rectangular Bounding Box with 1 or 2 rectangular "cutouts" missing from its edges or corners. 
   - 1 cutout requires exactly 6 points (e.g., L-shape).
   - 2 cutouts require exactly 8 points (e.g., Z-shape, T-shape).
3. Area Limit: Let the area of the main Bounding Box be $S$. The area of each individual cutout must be $\le 0.25 \times S$.
4. Tracing: The coordinates must trace the perimeter of the room in a continuous, non-intersecting closed loop (clockwise or counter-clockwise), starting at [0.0, 0.0]. Scale is in meters.

Output Format:
For each room, briefly state the Bounding Box dimensions, cutout dimensions, and prove the area constraint ($Cutout Area \le 0.25 \times S$). Then output the final JSON array: [[x1, y1], [x2, y2], ...].
\end{lstlisting}
After generating the scenes, we further verify the validity of each scene’s floor layout and regenerate any unreasonable cases. For the asset sets to be placed in the test scenes, the generation process is the same as in the main experiments. In total, we generate 20 irregular-scene test cases.
LayoutGPT and NaLA then perform object placement in the same scenes, and we compute the resulting OOB rate for comparison.

For the efficiency experiments, all tests are conducted on the same NVIDIA H20 GPU. NaLA performs inference locally, while the other three baseline models obtain prompts via API calls and perform optimization locally. Each scene generation process is repeated five times, and the average runtime is reported.

\section{Failure Case Study}

\begin{figure*}[htbp]
    \centering
    \begin{subfigure}[b]{0.45\textwidth}
        \centering
        \includegraphics[width=0.6\linewidth, trim=2cm 5cm 2cm 5cm, clip]{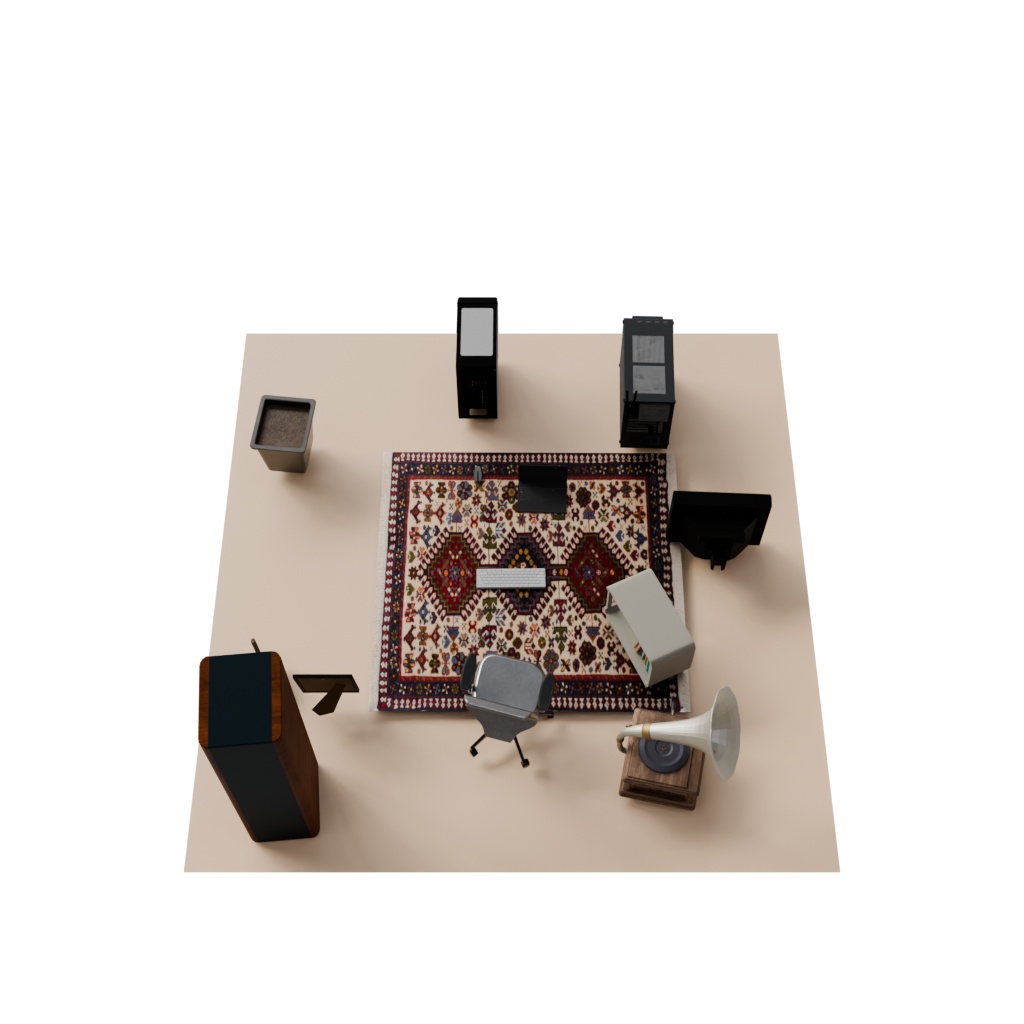}
        \caption{Computer Room}
        \label{fig:fail_computer}
    \end{subfigure}
    \hfill
    \begin{subfigure}[b]{0.45\textwidth}
        \centering
        \includegraphics[width=0.6\linewidth, trim=2cm 5cm 2cm 5cm, clip]{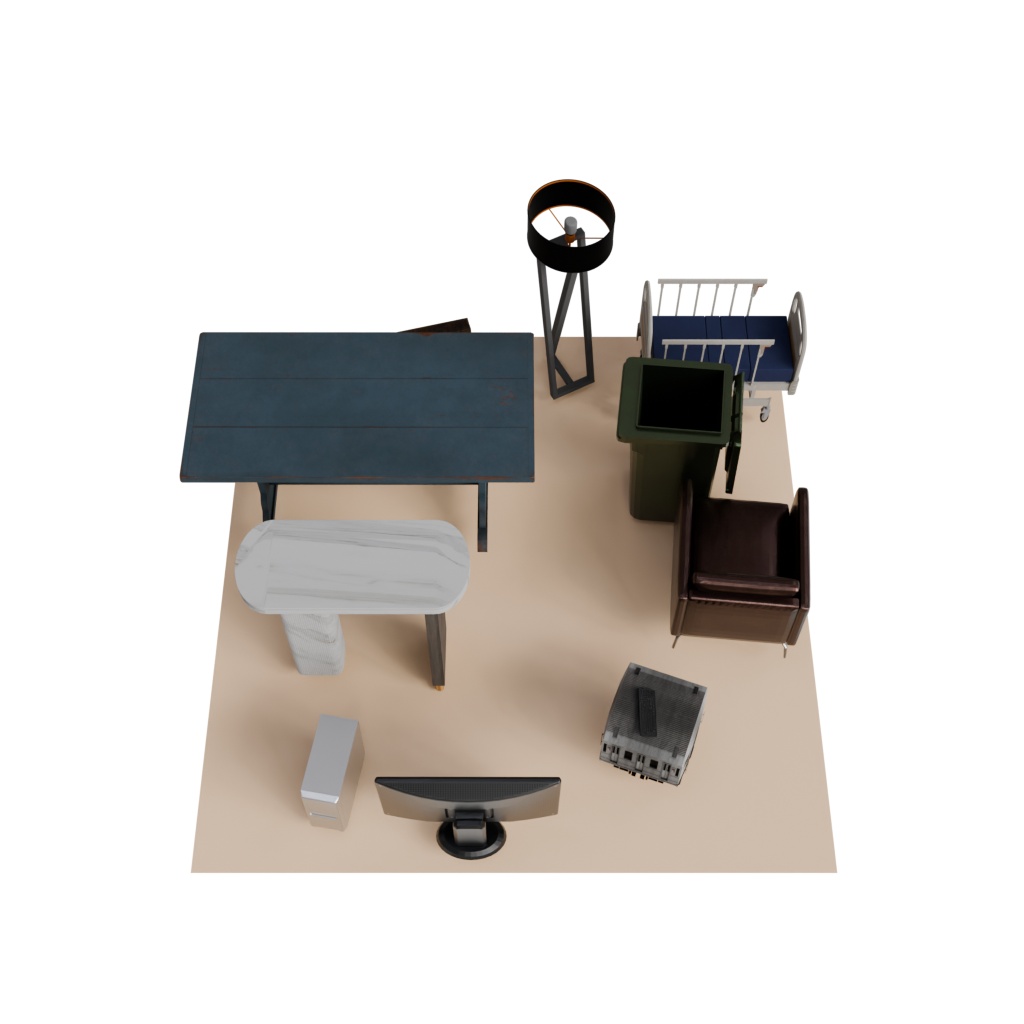}
        \caption{Clinic}
        \label{fig:fail_dental}
    \end{subfigure}
    
    \vspace{-2mm}
    
    \caption{Failure Cases of NaLA. The model finds it difficult to learn placement patterns for rare assets (e.g., the spatial relationships among computers, mice, and keyboards, or the layout of clinical equipment) from limited training examples.}
    \label{fig: failure_cases}
    \vspace{-4mm}
\end{figure*}

It should be noted that, due to limited training data and the backbone model's capability constraints, NaLA may still produce failure cases. \cref{fig: failure_cases} illustrates two such examples. When encountering assets that rarely appear in the training set—such as computers, mice, and keyboards in a computer room, or carts and stands in a clinic—the model struggles to learn their placement patterns from the scarce training data, leading to degraded placement performance.
On the other hand, due to the limitations of the LLM backbone, NaLA may also produce inference errors, such as object inflation or floating objects. We believe that these issues will be gradually mitigated as larger, high-quality layout datasets become available and foundation model capabilities continue to improve.

%% file: main.bib
@String(AAAI  = {AAAI})

@String(TOG   = {ACM Trans. Graph.})

@String(TOG   = {ACM TOG})

@article{makeithome2011,
  title={Make it home: Automatic optimization of furniture arrangement.},
  author={Yu, Lap-Fai and Yeung, Sai Kit and Tang, Chi-Keung and Terzopoulos, Demetri and Chan, Tony F and Osher, Stanley J},
  journal={ACM Trans. Graph.},
  volume={30},
  number={4},
  pages={86},
  year={2011}
}

@inproceedings{scenecraft2023,
  title={Scenecraft: automating interactive narrative scene generation in digital games with large language models},
  author={Kumaran, Vikram and Rowe, Jonathan and Mott, Bradford and Lester, James},
  booktitle={Proceedings of the AAAI Conference on Artificial Intelligence and Interactive Digital Entertainment},
  volume={19},
  number={1},
  pages={86--96},
  year={2023}
}

@misc{internscenes2025,
  title={Internscenes: A large-scale simulatable indoor scene dataset with realistic layouts},
  author={Zhong, Weipeng and Cao, Peizhou and Jin, Yichen and Li, Luo and Cai, Wenzhe and Lin, Jingli and Wang, Hanqing and Lyu, Zhaoyang and Wang, Tai and XU, Xudong and others},
  journal={Advances in Neural Information Processing Systems},
  volume={38},
  year={2026}
}

@article{rulebased2017,
  title={Authoring landscapes by combining ecosystem and terrain erosion simulation},
  author={Cordonnier, Guillaume and Galin, Eric and Gain, James and Benes, Bedrich and Gu{\'e}rin, Eric and Peytavie, Adrien and Cani, Marie-Paule},
  journal={ACM Transactions on Graphics (TOG)},
  volume={36},
  number={4},
  pages={1--12},
  year={2017},
  publisher={ACM New York, NY, USA}
}

@article{MIQP2018,
  title={Miqp-based layout design for building interiors},
  author={Wu, Wenming and Fan, Lubin and Liu, Ligang and Wonka, Peter},
  booktitle={Computer Graphics Forum},
  volume={37},
  number={2},
  pages={511--521},
  year={2018},
  organization={Wiley Online Library}
}

@article{LLMfewshot2020,
  title={Language models are few-shot learners},
  author={Brown, Tom and Mann, Benjamin and Ryder, Nick and Subbiah, Melanie and Kaplan, Jared D and Dhariwal, Prafulla and Neelakantan, Arvind and Shyam, Pranav and Sastry, Girish and Askell, Amanda and others},
  journal={Advances in neural information processing systems},
  volume={33},
  pages={1877--1901},
  year={2020}
}

@inproceedings{radford2021learning,
  title={Learning transferable visual models from natural language supervision},
  author={Radford, Alec and Kim, Jong Wook and Hallacy, Chris and Ramesh, Aditya and Goh, Gabriel and Agarwal, Sandhini and Sastry, Girish and Askell, Amanda and Mishkin, Pamela and Clark, Jack and others},
  booktitle={International conference on machine learning},
  pages={8748--8763},
  year={2021},
  organization={PmLR}
}

@misc{LayoutGPT,
  title={Layoutgpt: Compositional visual planning and generation with large language models},
  author={Feng, Weixi and Zhu, Wanrong and Fu, Tsu-jui and Jampani, Varun and Akula, Arjun and He, Xuehai and Basu, Sugato and Wang, Xin Eric and Wang, William Yang},
  journal={Advances in Neural Information Processing Systems},
  volume={36},
  pages={18225--18250},
  year={2023}
}

@misc{holodeck,
  title={Holodeck: Language guided generation of 3d embodied ai environments},
  author={Yang, Yue and Sun, Fan-Yun and Weihs, Luca and VanderBilt, Eli and Herrasti, Alvaro and Han, Winson and Wu, Jiajun and Haber, Nick and Krishna, Ranjay and Liu, Lingjie and others},
  booktitle={Proceedings of the IEEE/CVF Conference on Computer Vision and Pattern Recognition},
  pages={16227--16237},
  year={2024}
}

@misc{qwen2.5,
    title = {Qwen2.5: A Party of Foundation Models},
    url = {https://qwenlm.github.io/blog/qwen2.5/},
    author = {Qwen Team},
    month = {September},
    year = {2024}
}

@article{gemini_team2024gemini,
  title={Gemini: a family of highly capable multimodal models},
  author={Team, Gemini and Anil, Rohan and Borgeaud, Sebastian and Alayrac, Jean-Baptiste and Yu, Jiahui and Soricut, Radu and Schalkwyk, Johan and Dai, Andrew M and Hauth, Anja and Millican, Katie and others},
  journal={arXiv preprint arXiv:2312.11805},
  year={2023}
}

@article{efron1979bootstrap,
  title={Bootstrap methods: another look at the jackknife},
  author={Efron, Bradley},
  booktitle={Breakthroughs in statistics: Methodology and distribution},
  pages={569--593},
  year={1992},
  publisher={Springer}
}

@article{likert1932technique,
  title={A technique for the measurement of attitudes.},
  author={Likert, Rensis},
  journal={Archives of psychology},
  year={1932}
}

@misc{sun2025layoutvlmdifferentiableoptimization3d,
  title={Layoutvlm: Differentiable optimization of 3d layout via vision-language models},
  author={Sun, Fan-Yun and Liu, Weiyu and Gu, Siyi and Lim, Dylan and Bhat, Goutam and Tombari, Federico and Li, Manling and Haber, Nick and Wu, Jiajun},
  booktitle={Proceedings of the Computer Vision and Pattern Recognition Conference},
  pages={29469--29478},
  year={2025}
}

@misc{llplace,
  title={Llplace: The 3d indoor scene layout generation and editing via large language model},
  author={Yang, Yixuan and Lu, Junru and Zhao, Zixiang and Luo, Zhen and Yu, James JQ and Sanchez, Victor and Zheng, Feng},
  journal={arXiv preprint arXiv:2406.03866},
  year={2024}
}

@article{mildenhall2021nerf,
  title={Nerf: Representing scenes as neural radiance fields for view synthesis},
  author={Mildenhall, Ben and Srinivasan, Pratul P and Tancik, Matthew and Barron, Jonathan T and Ramamoorthi, Ravi and Ng, Ren},
  journal={Communications of the ACM},
  volume={65},
  number={1},
  pages={99--106},
  year={2021},
  publisher={ACM New York, NY, USA}
}

@article{kerbl20233dgaussian,
  title={3d gaussian splatting for real-time radiance field rendering.},
  author={Kerbl, Bernhard and Kopanas, Georgios and Leimk{\"u}hler, Thomas and Drettakis, George and others},
  journal={ACM Trans. Graph.},
  volume={42},
  number={4},
  pages={139--1},
  year={2023}
}

@inproceedings{yu2021pointbert,
  title={Point-bert: Pre-training 3d point cloud transformers with masked point modeling},
  author={Yu, Xumin and Tang, Lulu and Rao, Yongming and Huang, Tiejun and Zhou, Jie and Lu, Jiwen},
  booktitle={Proceedings of the IEEE/CVF conference on computer vision and pattern recognition},
  pages={19313--19322},
  year={2022}
}

@inproceedings{fu20213d,
  title={3d-front: 3d furnished rooms with layouts and semantics},
  author={Fu, Huan and Cai, Bowen and Gao, Lin and Zhang, Ling-Xiao and Wang, Jiaming and Li, Cao and Zeng, Qixun and Sun, Chengyue and Jia, Rongfei and Zhao, Binqiang and others},
  booktitle={Proceedings of the IEEE/CVF International Conference on Computer Vision},
  pages={10933--10942},
  year={2021}
}

@article{zhu2025imaginarium,
  title={Imaginarium: Vision-guided High-Quality 3D Scene Layout Generation},
  author={Zhu, Xiaoming and Huang, Xu and Xie, Qinghongbing and Deng, Zhi and Yu, Junsheng and Guan, Yirui and Liu, Zhongyuan and Zhu, Lin and Zhao, Qijun and Liu, Ligang and others},
  journal={ACM Transactions on Graphics (TOG)},
  volume={44},
  number={6},
  pages={1--24},
  year={2025},
  publisher={ACM New York, NY, USA}
}

@inproceedings{SceneTeller2024,
  title={Sceneteller: Language-to-3d scene generation},
  author={{\"O}cal, Ba{\c{s}}ak Melis and Tatarchenko, Maxim and Karao{\u{g}}lu, Sezer and Gevers, Theo},
  booktitle={European Conference on Computer Vision},
  pages={362--378},
  year={2024},
  organization={Springer}
}

@inproceedings{lai2024lisa,
  title={Lisa: Reasoning segmentation via large language model},
  author={Lai, Xin and Tian, Zhuotao and Chen, Yukang and Li, Yanwei and Yuan, Yuhui and Liu, Shu and Jia, Jiaya},
  booktitle={Proceedings of the IEEE/CVF conference on computer vision and pattern recognition},
  pages={9579--9589},
  year={2024}
}

@inproceedings{zhou2025scenex,
  title={Scenex: Procedural controllable large-scale scene generation},
  author={Zhou, Mengqi and Wang, Yuxi and Hou, Jun and Zhang, Shougao and Li, Yiwei and Luo, Chuanchen and Peng, Junran and Zhang, Zhaoxiang},
  booktitle={Proceedings of the AAAI Conference on Artificial Intelligence},
  volume={39},
  number={10},
  pages={10806--10814},
  year={2025}
}

@inproceedings{tang2024diffuscene,
  title={Diffuscene: Denoising diffusion models for generative indoor scene synthesis},
  author={Tang, Jiapeng and Nie, Yinyu and Markhasin, Lev and Dai, Angela and Thies, Justus and Nie{\ss}ner, Matthias},
  booktitle={Proceedings of the IEEE/CVF conference on computer vision and pattern recognition},
  pages={20507--20518},
  year={2024}
}

@article{ling2025scenethesis,
  title={Scenethesis: A language and vision agentic framework for 3d scene generation},
  author={Ling, Lu and Lin, Chen-Hsuan and Lin, Tsung-Yi and Ding, Yifan and Zeng, Yu and Sheng, Yichen and Ge, Yunhao and Liu, Ming-Yu and Bera, Aniket and Li, Zhaoshuo},
  journal={arXiv preprint arXiv:2505.02836},
  year={2025}
}

@inproceedings{li2022blip,
  title={Blip: Bootstrapping language-image pre-training for unified vision-language understanding and generation},
  author={Li, Junnan and Li, Dongxu and Xiong, Caiming and Hoi, Steven},
  booktitle={International conference on machine learning},
  pages={12888--12900},
  year={2022},
  organization={PMLR}
}

@inproceedings{gao2024graphdreamer,
  title={Graphdreamer: Compositional 3d scene synthesis from scene graphs},
  author={Gao, Gege and Liu, Weiyang and Chen, Anpei and Geiger, Andreas and Sch{\"o}lkopf, Bernhard},
  booktitle={Proceedings of the IEEE/CVF Conference on Computer Vision and Pattern Recognition},
  pages={21295--21304},
  year={2024}
}

@inproceedings{DIscene2024,
  title={Discene: Object decoupling and interaction modeling for complex scene generation},
  author={Li, Xiao-Lei and Li, Haodong and Chen, Hao-Xiang and Mu, Tai-Jiang and Hu, Shi-Min},
  booktitle={SIGGRAPH Asia 2024 Conference Papers},
  pages={1--12},
  year={2024}
}

@article{wen20253d,
  title={3d scene generation: A survey},
  author={Wen, Beichen and Xie, Haozhe and Chen, Zhaoxi and Hong, Fangzhou and Liu, Ziwei},
  journal={arXiv preprint arXiv:2505.05474},
  year={2025}
}

@InProceedings{WonderJourney2024Yu,
  title={Wonderjourney: Going from anywhere to everywhere},
  author={Yu, Hong-Xing and Duan, Haoyi and Hur, Junhwa and Sargent, Kyle and Rubinstein, Michael and Freeman, William T and Cole, Forrester and Sun, Deqing and Snavely, Noah and Wu, Jiajun and others},
  booktitle={Proceedings of the IEEE/CVF Conference on Computer Vision and Pattern Recognition},
  pages={6658--6667},
  year={2024}
}

@inproceedings{4real2024Yu,
  title={4real: Towards photorealistic 4d scene generation via video diffusion models},
  author={Yu, Heng and Wang, Chaoyang and Zhuang, Peiye and Menapace, Willi and Siarohin, Aliaksandr and Cao, Junli and Jeni, L{\'a}szl{\'o} and Tulyakov, Sergey and Lee, Hsin-Ying},
  journal={Advances in Neural Information Processing Systems},
  volume={37},
  pages={45256--45280},
  year={2024}
}

@misc{li2024director3drealworldcameratrajectory,
  title={Director3d: Real-world camera trajectory and 3d scene generation from text},
  author={Li, Xinyang and Lai, Zhangyu and Xu, Linning and Qu, Yansong and Cao, Liujuan and Zhang, Shengchuan and Dai, Bo and Ji, Rongrong},
  journal={Advances in neural information processing systems},
  volume={37},
  pages={75125--75151},
  year={2024}
}

@InProceedings{NeuralField-LDMKim_2023_CVPR,
  title={Neuralfield-ldm: Scene generation with hierarchical latent diffusion models},
  author={Kim, Seung Wook and Brown, Bradley and Yin, Kangxue and Kreis, Karsten and Schwarz, Katja and Li, Daiqing and Rombach, Robin and Torralba, Antonio and Fidler, Sanja},
  booktitle={Proceedings of the IEEE/CVF conference on computer vision and pattern recognition},
  pages={8496--8506},
  year={2023}
}

@inbook{I-Design_elen_2025,
  title={I-design: Personalized llm interior designer},
  author={{\c{C}}elen, Ata and Han, Guo and Schindler, Konrad and Van Gool, Luc and Armeni, Iro and Obukhov, Anton and Wang, Xi},
  booktitle={European Conference on Computer Vision},
  pages={217--234},
  year={2024},
  organization={Springer}
  }

@article{SceneGraphLI2024127052,
  title={Scene graph generation: A comprehensive survey},
  author={Li, Hongsheng and Zhu, Guangming and Zhang, Liang and Jiang, Youliang and Dang, Yixuan and Hou, Haoran and Shen, Peiyi and Zhao, Xia and Shah, Syed Afaq Ali and Bennamoun, Mohammed},
  journal={Neurocomputing},
  volume={566},
  pages={127052},
  year={2024},
  publisher={Elsevier}
}

@inproceedings{zausinger2025regress,
  title={Regress, Don't Guess--A Regression-like Loss on Number Tokens for Language Models},
  author={Zausinger, Jonas and Pennig, Lars and Kozina, Anamarija and Sdahl, Sean and Sikora, Julian and Dendorfer, Adrian and Kuznetsov, Timofey and Hagog, Mohamad and Wiedemann, Nina and Chlodny, Kacper and others},
  journal={arXiv preprint arXiv:2411.02083},
  year={2024}
}

@InProceedings{LLaVA-3DZhu_2025_ICCV,
  title={Llava-3d: A simple yet effective pathway to empowering lmms with 3d capabilities},
  author={Zhu, Chenming and Wang, Tai and Zhang, Wenwei and Pang, Jiangmiao and Liu, Xihui},
  booktitle={Proceedings of the IEEE/CVF International Conference on Computer Vision},
  pages={4295--4305},
  year={2025}
}

@inproceedings{spfprmer2022,
  title={Superpoint transformer for 3d scene instance segmentation},
  author={Sun, Jiahao and Qing, Chunmei and Tan, Junpeng and Xu, Xiangmin},
  booktitle={Proceedings of the AAAI Conference on Artificial Intelligence},
  volume={37},
  number={2},
  pages={2393--2401},
  year={2023}
}

@misc{zhang2026m3dlayoutmultisourcedataset3d,
  title={M3DLayout: A multi-source dataset of 3D indoor layouts and structured descriptions for 3D generation},
  author={Zhang, Yiheng and Cai, Zhuojiang and Wang, Mingdao and Guo, Meitong and Li, Tianxiao and Lin, Li and Wang, Yuwang},
  booktitle={Proceedings of the IEEE/CVF Conference on Computer Vision and Pattern Recognition},
  pages={34217--34226},
  year={2026}
}

@inproceedings{SpatialLM,
    title     = {SpatialLM: Training Large Language Models for Structured Indoor Modeling},
    author    = {Mao, Yongsen and Zhong, Junhao and Fang, Chuan and Zheng, Jia and Tang, Rui and Zhu, Hao and Tan, Ping and Zhou, Zihan},
    booktitle = {Advances in Neural Information Processing Systems},
    year      = {2025}
}

@article{ran2026direct,
  title={Direct numerical layout generation for 3d indoor scene synthesis via spatial reasoning},
  author={Ran, Xingjian and Li, Yixuan and Xu, Linning and Yu, Mulin and Dai, Bo},
  journal={Advances in Neural Information Processing Systems},
  volume={38},
  pages={125055--125081},
  year={2026}
}
